\documentclass[11pt]{article}

\usepackage[numbers]{natbib}
\usepackage{csquotes}
\usepackage{graphicx}
\usepackage{color}
\usepackage{bbm}
\usepackage{latexsym}
\usepackage{amsmath}               % ACTIVATION EFFECTS LABELS  
\usepackage{amssymb}               % ACTIVATION EFFECTS LABELS  
\usepackage{amsthm}               % already in jmlr2e.sty
\usepackage{hyperref}
\usepackage{psfrag} 
\usepackage{wasysym}
\usepackage{xparse}

\usepackage[ruled,vlined]{algorithm2e}

\usepackage{cleveref}

\definecolor{white}{rgb}{1.0,1.0,1.0}
\definecolor{brightred}{rgb}{1.0,0.1,0.1}
\definecolor{brightblue}{rgb}{0.0,0.0,0.8}
\definecolor{darkblue}{rgb}{0.0,0.0,0.5}
\definecolor{darkgreen}{rgb}{0.0,0.3,0.0}
\definecolor{brightgreen}{rgb}{0.0,0.8,0.0}
\definecolor{darkblack}{rgb}{0.0,0.0,0.0}
\definecolor{grey}{rgb}{0.3,0.3,0.3}

\graphicspath{{./}}
\theoremstyle{definition}
\newtheorem{definition}{Definition}% added for enumeration of Definitions

\theoremstyle{definition}
\newtheorem{Example}{Example}

\newtheorem{lemma}{Lemma}
\newtheorem{mytheorem}{Theorem}

\definecolor{white}{rgb}{1.0,1.0,1.0}
\definecolor{brightred}{rgb}{1.0,0.1,0.1}
\definecolor{brightblue}{rgb}{0.0,0.0,0.8}
\definecolor{darkblue}{rgb}{0.0,0.0,0.5}
\definecolor{darkgreen}{rgb}{0.0,0.3,0.0}
\definecolor{brightgreen}{rgb}{0.0,0.8,0.0}
\definecolor{darkblack}{rgb}{0.0,0.0,0.0}
\definecolor{grey}{rgb}{0.3,0.3,0.3}

\long\def\comment#1{}  % comment out long passages

\newcommand{\disT}{\textstyle}
\newcommand{\disS}{\displaystyle}

\newcommand{\undersets}[2]{\underset{\hbox to 0pt{\hss{\scriptsize #1}\hss}}{#2}}
\newcommand{\oversets}[2]{\overset{\hbox to 0pt{\hss{\scriptsize #1}\hss}}{#2}}

\newcommand{\FF}{{\mathcal F}}
\newcommand{\FFt}{\tilde{\FF}}

\newcommand{\HH}{{\mathcal H}}
\newcommand{\HHt}{\tilde{\HH}}

\newcommand{\LL}{{\mathcal L}}
\newcommand{\LLt}{\tilde{\LL}}

\newcommand{\NN}{{\mathcal N}}

\newcommand{\RRR}{\mathbbm{R}}
\newcommand{\Scal}{{\mathcal S}}  % \SS is already defined

\newcommand{\Jnew}{\frac{\partial\etaVec(\zVec;\ThetaVec)}{\partial\thetaVec^{\mathrm{\,T}}}}
\newcommand{\JTnew}{\frac{\partial\etaVec^{\mathrm{\,T}}(\zVec;\ThetaVec)}{\partial\thetaVec}}
\newcommand{\Inew}{\frac{\partial\zetaVec(\PsiVec)}{\partial\PsiVec^{\mathrm{\,T}}}}
\newcommand{\ITnew}{\frac{\partial\zetaVec^{\mathrm{\,T}}(\PsiVec)}{\partial\PsiVec}}

\newcommand{\sit}{\tilde{\sigma}}

\newcommand{\EEE}[2]{\mathbbm{E}_{#1}\big\{#2\big\}}

\def\Vhrulefill{\leavevmode\leaders\hrule height 0.7ex depth \dimexpr0.4pt-0.7ex\hfill\kern0pt}

\newcommand{\qn}{q^{(n)}}

\newcommand{\qBar}{\overline{q}}
\newcommand{\pt}{\tilde{p}}

\newcommand{\classk}{k}

\newcommand{\subc}{c}

\newcommand{\xVec}{\vec{x}}
\newcommand{\zVec}{\vec{z}}
\newcommand{\xVecN}{\vec{x}^{\hspace{0.5pt}(n)}}

\newcommand{\fVec}{\vec{f}}
\newcommand{\gVec}{\vec{g}}

\newcommand{\wVec}{\vec{w}}

\newcommand{\PsiVec}{\vec{\Psi}}
\newcommand{\PsiVecT}{\PsiVec^{\mathrm{\,T}}}
\newcommand{\pT}{p_{\footnotesize{\ThetaVec}}}
\newcommand{\pTheta}{p_{\footnotesize{\ThetaVec}}}
\newcommand{\pPsi}{p_{\footnotesize{\PsiVec}}}

\newcommand{\qPhiN}{q_{\Phi}^{(n)}}

\newcommand{\pEtaVec}{p_{\etaVec}}

\newcommand{\zetaVec}{\vec{\zeta}}
\newcommand{\zetaVecT}{\vec{\zeta}^{\mathrm{\,T}}}
\newcommand{\pZetaVec}{p_{\zetaVec\,}}

\NewDocumentCommand{\y}{O{}O{}}{y_{#2}^{\,#1}{}}
\NewDocumentCommand{\yVec}{O{}}{\vec{y}_{\vphantom{c}}^{\,#1}{}}

\NewDocumentCommand{\Wgen}{O{}O{}}{\mathcal{W}_{#1#2}}

\NewDocumentCommand{\Rgen}{O{}O{}}{\mathcal{R}_{#1#2}}

\NewDocumentCommand{\W}{O{}O{}}{W_{#1#2}}

\NewDocumentCommand{\R}{O{}O{}}{R_{#1#2}}

\NewDocumentCommand{\Sc}{O{\subc}O{}}{s_{#1}^{#2}}

\NewDocumentCommand{\sVec}{O{}}{\vec{s}_{\vphantom{\subc}}^{\,#1}}

\NewDocumentCommand{\Igenc}{O{\subc}O{}}{\mathcal{I}_{#1}^{#2}}

\NewDocumentCommand{\Ic}{O{\subc}O{}}{I_{#1}^{#2}}

\NewDocumentCommand{\tk}{O{\classk}O{}}{t_{#1}^{#2}}

\NewDocumentCommand{\tVec}{O{}}{\vec{t}_{\vphantom{\classk}}^{\,#1}}

\NewDocumentCommand{\eps}{O{}}{\epsilon_{\textnormal{\tiny $#1$}}}

\NewDocumentCommand{\epst}{O{}}{\tilde{\epsilon}_{\textnormal{\tiny $#1$}}}

\newcommand{\Bern}{\textrm{Bern}}

\newcommand{\etaVec}{\vec{\eta}}
\newcommand{\etaVecT}{\vec{\eta}^{\mathrm{T}}}
\newcommand{\TVec}{\vec{T}}

\newcommand{\AVec}{\vec{A}}

\newcommand{\ThetaVec}{\vec{\Theta}}

\newcommand{\thetaVec}{\vec{\theta}}
\newcommand{\thetaVecT}{\vec{\theta}^{\mathrm{\,T}}}

\newcommand{\alphaVec}{\vec{\alpha}}
\newcommand{\betaVec}{\vec{\beta}}

\newcommand{\qbar}{\bar{q}}

\DeclareMathOperator{\im}{im}

\newcommand{\delll}[2]{\frac{\partial{}#1}{\partial{}#2}}
\newcommand{\del}[1]{\frac{\partial}{\partial{}#1}}
\newcommand{\dell}[1]{\disT{\frac{\partial}{\partial{}#1}}\disS}

\begin{document}

%\title[Equality of the ELBO to Entropy Sums]{On the Equality of the ELBO to a Sum of Entropies at\\ Stationary Points of Learning}
\title{On the Equality of the ELBO to a Sum of Entropies at\\ Stationary Points of Learning}

\author{%\ \\[1ex]
J\"org L\"ucke\\
\footnotesize                   Artificial Intelligence Lab\\[-0.5ex]
\footnotesize                      Department of Computer Science\\[-0.5ex]
\footnotesize                   Faculty of Mathematics, Computer Science and Physics\\[-0.5ex]
\footnotesize                    University of Innsbruck, Austria\\[0ex]
\and
Jan Warnken\\
\footnotesize                   Machine Learning Lab\\[-0.5ex]
\footnotesize                    Faculties V and VI\\[-0.5ex]
\footnotesize                   University of Oldenburg, Germany\\[0ex]%\\[-0.5ex]
%\footnotesize                    26111 Oldenburg, Germany\\[0ex]
}
\date{}
\maketitle

\begin{abstract}
\noindent{}The variational lower bound (a.k.a.\ ELBO or free energy) is the central objective for many established as well as for many novel algorithms for unsupervised learning. 
Such algorithms usually increase the bound until parameters have converged to values close to a stationary point of the learning dynamics.
Here we show that (for a very large class of generative models) the variational lower bound is at all stationary points of learning equal to a sum of entropies.
Concretely, for standard generative models with one set of latents and one set of observed variables, the sum consists of three entropies: (A)~the (average) entropy of the variational distributions, (B)~the negative entropy of the model’s prior distribution, and (C)~the (expected) negative entropy of the observable distribution. The obtained result applies under realistic conditions including: finite numbers of data points, at any stationary point (including saddle points) and for any family of (well behaved) variational distributions. The class of generative models for which we show the equality to entropy sums contains many standard as well as novel generative models including variational autoencoders (a prominent class of deep generative models). % (a special case shown previously). % \cite{DammEtAl2023}).
The prerequisites we use to show equality to entropy sums are relatively mild. Concretely, the distributions defining a given generative model have to be of the exponential family, and the model has to satisfy a parameterization criterion (which is fulfilled for many well-known models). Proving equality of the ELBO to entropy sums at stationary points (under the stated conditions) is the main contribution of this work.
\end{abstract}

\section{Introduction}
We consider probabilistic generative models and their optimization based on the variational lower bound. The variational bound
is also known as {\em evidence lower bound} (ELBO; e.g.\,\cite{HoffmanEtAl2013}) or (variational) {\em free energy} \cite{NealHinton1998}.
For a given set of data points, a learning algorithm usually changes the parameters of a model until 
no or only negligible parameter changes are observed. After learning, parameters will therefore usually lie very close to a stationary
point of the learning objective. In the case of ELBO-based learning, parameters will usually lie close to a point where the derivatives of the ELBO objective w.r.t.\ all optimized parameters are zero.

In this theoretical study, we investigate the ELBO objective at its stationary points. As our main result, we show that for many generative
models (including very common models), the ELBO becomes equal to a sum of entropies at stationary points. The result is very concise such that it
can be stated here initially. We will consider standard generative models of the form:

\begin{eqnarray}
 \zVec \,\sim\, p_{\Theta}(\zVec)\,\ \ \mbox{\ and\ }\ \ \xVec \,\sim\, p_{\Theta}(\xVec\,|\,\zVec)\,, \label{EqnGenModelIntro}%\vspace{-3ex}
\end{eqnarray}
where $\zVec$ and $\xVec$ are latent and observed variables, respectively, and where $\Theta$ denotes the set of model parameters. The distribution $p_{\Theta}(\zVec)$
will be denoted as the {\em prior} distribution, and $p_{\Theta}(\xVec\,|\,\zVec)$ will be referred to as {\em noise model} or {\em observable distribution}.
Given $N$ data points $\xVecN$, the ELBO is given by (compare \cite{JordanEtAl1999,NealHinton1998,HoffmanEtAl2013,KingmaWelling2014}):
%we consider parameter optimization of the model based on the variational lower bound (a.k.a.\ ELBO or free energy) given by:
%
%\begin{eqnarray}
\begin{align}
 \FF(\Phi,\Theta) 
%&=  \textstyle\frac{1}{N}\sum_{n} \int \qn_{\Phi}\!(\zVec) \log\!\big( \pT(\xVecN\,|\,\zVec)\,p_{\PsiVec}(\zVec) \big)\hspace{0.5ex} \mathrm{d}\zVec   -\, \frac{1}{N}\sum_{n} \int \qn_{\Phi}\!(\zVec) \log\!\big( \qn_{\Phi}\!(\zVec) \big) \mathrm{d}\zVec \label{EqnFFStandard}\\
%
%\hspace{-1ex}&=&\disS\hspace{-1ex}  \frac{1}{N}\sum_{n} \int \qn_{\Phi}\!(\zVec) \log\!\big( p_{\Theta}(\xVecN\,|\,\zVec)\,q_{\Theta}(\zVec) \big)\hspace{0.5ex} \mathrm{d}\zVec   -\, \frac{1}{N}\sum_{n} \int \qn_{\Phi}\!(\zVec) \log\!\big( \qn_{\Phi}\!(\zVec) \big) \mathrm{d}\zVec \label{EqnFFStandardIntro}\\
%
%                 \hspace{-1ex}&=&\disS\hspace{-1ex}   \disS\frac{1}{N}\sum_{n} \int \qn_{\Phi}\!(\zVec)\, \log\!\big( p_{\Theta}(\xVecN\,|\,\zVec) \big)\hspace{0.5ex} \mathrm{d}\zVec   
%
=\frac{1}{N}\sum_{n} \int \qn_{\Phi}\!(\zVec)\, \log\!\big( p_{\Theta}(\xVecN\,|\,\zVec) \big)\hspace{0.5ex} \mathrm{d}\zVec
                 \,-\, \frac{1}{N}\sum_{n} D_{\mathrm{KL}}\big[ \qn_{\Phi}\!(\zVec),p_{\Theta}(\zVec) \big]\,, \label{EqnFFIntro}
%
%\end{eqnarray}
\end{align}
where $\qn_{\Phi}(\zVec)$ denote the variational distributions used for optimization.

The main result of this work states that under relatively mild conditions for the generative model in (\ref{EqnGenModelIntro}),
the ELBO decomposes into a sum of entropies at all stationary points of learning. % (i.e., at all points of zero derivatives).
Concretely, we will show that at all stationary points (i.e., at all points of vanishing derivatives of $\FF(\Phi,\Theta)$) applies:
\begin{align}
%
%\FF(\Phi,\Theta) \hspace{-1ex}&=&\disS\hspace{-1ex}  \frac{1}{N}\sum_{n} \HH[\qPhiN(\zVec)]  \,-\, \HH[\,p_{\Theta}(\zVec)] \,-\, \frac{1}{N} \sum_{n} \EEE{\qn_{\Phi}}{ \HH[\,p_{\Theta}(\xVec\,|\,\zVec)] }\,.\phantom{\small{}ix} 
\ \FF(\Phi,\Theta) =\frac{1}{N}\sum_{n} \HH[\qPhiN(\zVec)]  \,-\, \HH[\,p_{\Theta}(\zVec)] \,-\, \frac{1}{N} \sum_{n} \EEE{\qn_{\Phi}}{ \HH[\,p_{\Theta}(\xVec\,|\,\zVec)] }\,.\phantom{\small{}ix}
%
%\FF(\Phi,\Theta) \hspace{-1ex}&=&\disS\hspace{-1ex}  \frac{1}{N}\sum_{n} \HH[\qPhiN(\zVec)]  \,-\, \HH[\,p_{\Theta}(\zVec)] \,-\, \EEE{\qBar_{\Phi}}{ \HH[\,p_{\Theta}(\xVec\,|\,\zVec)] }\,.\phantom{\small{}ix} 
%
\label{EqnTheoremIntro}
%
%\FF(\Phi,\Theta) \hspace{-1ex}&=&\hspace{-1ex} \underbrace{\frac{1}{N}\sum_{n=1}^N \HH[\qPhi(\zVec;\xVecN)]}\phantom{ii}\ \underbrace{\phantom{\sum_{n=1}^N}\hspace{-3ex}-\ \HH[\pT(\zVec)]}\phantom{ii}  \underbrace{-\ \frac{1}{N} \sum_{n=1}^N \EEE{\qn_{\Phi}}{ \HH[\pT(\xVec\,|\,\zVec)] }}\phantom{\small{}ix} %
%\label{HypoVAE}\\
%
%\FF(\Phi,\Theta) \hspace{-1ex}&=&\hspace{-1ex} \underbrace{\frac{1}{N}\sum_{n=1}^N \HH[\qPhi(\zVec;\xVecN)]}_{\mbox{\small encoder}}\phantom{ii}\ \underbrace{\phantom{\sum_{n=1}^N}\hspace{-3ex}-\ \HH[\pT(\zVec)]}_{\mbox{\small prior}}\phantom{ii}  \underbrace{-\ \frac{1}{N} \sum_{n=1}^N \EEE{\qn_{\Phi}}{ \HH[\pT(\xVec\,|\,\zVec)] }}_{\mbox{\small decoder}}. %
%\nonumber
%\label{HypoVAE}\\
%
%
%\FF(\Phi,\Theta) \hspace{-1ex}&=&\hspace{-1ex} \overbrace{\frac{1}{N}\sum_{n=1}^N \HH[\qPhi(\zVec;\xVecN)]}_{\mbox{\small encoder}}\phantom{ii}\ \overbrace{\phantom{\sum_{n=1}^N}\hspace{-3ex}-\ \HH[\pT(\zVec)]}_{\mbox{\small prior}}\phantom{ii}  \overbrace{-\ \frac{1}{N} \sum_{n=1}^N \EEE{\qn_{\Phi}}{ \HH[\pT(\xVec\,|\,\zVec)] }}_{\mbox{\small decoder}}. \label{HypoVAE}
%
%
\end{align}
Notably, the result applies at any stationary point (including saddle points), for finitely many data points, and for essentially any variational distribution (very weak conditions on $\qn_{\Phi}(\zVec)$).

As ELBO-based learning can be expected to finally result in parameters close to a stationary point, we can expect the value of the ELBO objective (\ref{EqnFFIntro}) to be approximately given by the entropy sum (\ref{EqnTheoremIntro}) after learning. In this sense, we will sometimes make statements such as {\em the ELBO converges to an entropy sum}. But we stress that all formal results in this work (propositions/theorems) will be about the stationary points of the ELBO themselves, and not about the learning dynamcis. In practice, ELBO-based learning (and learning using other objectives) never fully arrives at stationary points (at which the here derived results hold exactly). However, it was empirically observed for specific models that ELBOs are very closely matched by their corresponding entropy sums after learning \cite{DammEtAl2023,VelychkoEtAl2024}.
%For a given set of data points, a learning algorithm usually changes the parameters of a model until 
%no or only negligible parameter changes are observed. After learning, parameters will therefore usually lie very close to a stationary
%point of the learning objective. In the case of ELBO-based learning, parameters will usually lie close to a point where the derivatives of the ELBO objective w.r.t.\ all optimized %parameters are zero.}
%
%Once no notable changes occur, it is (in Machine Learning) common to state that learning has {\em converged} (e.g., \cite{Bishop2006,ShalevBen2014}).
%In the following (and outside of formal propositions/theorems) we sometimes reason about parameters that {\em have converged} or parameters , with which we mean that the optimized parameter values represent a stationary point of the ELBO objective (altough in practice learning usually never fully reaches stationary points).
%

The equality of the ELBO to entropy sums at stationary points can be useful from theoretical as well as practical perspectives. The entropy decomposition (\ref{EqnTheoremIntro}) can, for instance, be closed-form also in cases when the original ELBO objective is not (cf.\,\cite{DammEtAl2023}). The computation of variational bounds and/or likelihoods can also become easier for other reasons, e.g., for variational autoencoders with linear decoder (an approach closely related to PCA \cite{DaiEtAl2018,LucasEtAl2019}) the likelihood can at stationary points be computed from the model parameters alone without knowledge of the data (cf.\,\cite{DammEtAl2023}). % (we discuss in Sec.\,\ref{SecDiscussion}).
A number of example applications of Eqn.\,\ref{EqnTheoremIntro} have been discussed for specific generative models in previous work \cite{DammEtAl2023,VelychkoEtAl2024}, and they range from efficient ELBO estimation and model selection to the analysis of optimization landscapes and posterior collapse in standard variational autoencoders. % \citep[][]{DammEtAl2023}.
Entropy sum reformulations of the ELBO can also be used for learning itself, which has recently been demonstrated for probabilistic sparse coding \cite{VelychkoEtAl2024}. Results for sparse coding can, in turn, be relevant for compressive sensing (cf.\,\cite{Donoho2006}). Learning based on entropy sums such as (\ref{EqnTheoremIntro}) is made possible because (for many generative models) it is sufficient that only derivatives w.r.t.\ a subset of the parameters $\Theta$ vanish. 
Learning using entropy sums is then advantages for practical reasons, e.g., because (A)~an exclusively ELBO-based objective suggests more principled forms of annealing \cite{VelychkoEtAl2024}, or (B)~derivatives of entropies can take on especially concise forms. From a theoretical perspective, entropies have themselves deep roots in mathematical statistics and information theory (e.g.\,\cite{Efron1978,Amari2016}), and their well studied properties can potentially exploited if learning objectives are re-expressed in terms of entropies.
%
%, and their derivatives provide a direct connection to information geometry
%(e.g.\,\cite{Efron1978,Amari2016}). % (we will discuss in Sec.\,\ref{SecDiscussion}). % (we will discuss).
Information-theoretic considerations arise, similarly, in rate-distortion theory, where the representation of random variables under reconstruction constraints is characterized in terms of information quantities~\cite{RieglerEtAl2023}. A related perspective emerges here from the entropy-based characterization of the ELBO.

\ \\[-2ex]
\noindent{\bf Own contributions.}
% Related work.
In contrast to previous work on entropy sums (see \cite{DammEtAl2023,VelychkoEtAl2024}), we will not focus on a specific model or on specific applications of the result of Eqn.\,\ref{EqnTheoremIntro}.
Instead, the focus here will be the derivation of Eqn.\,\ref{EqnTheoremIntro} itself for an as broad class of generative models as possible.
%
%the derivation of Eq.\,\ref{EqnTheoremIntro} itself, and to provide and under as general conditions as possible.
%
%Instead, the focus here will be to derive the result of Eq.\,\ref{EqnTheoremIntro} for an as broad class of generative models as possible.
%

More concretely, we aim at formalizing as general as possible conditions that have to apply for a generative model in order for
its ELBO to converge to an entropy sum. For a given generative model, it will then be sufficient to verify a set of relatively
mild and concise conditions that make its ELBO equal to a sum of entropies.
%
%in order to show that its ELBO is equal to entropy sums at stationary points.
%
%The focus of this contribution will be the derivation of the result itself, however, including the general conditions that a generative model has
%to satisfy in order for the result to apply. More concretely and in contrast to all previous contributions, we will be interested in defining
%a broad class of generative models which all show equality of the ELBO to entropy sums at stationary points.
%
%We will aim at defining the class as broadly as possible, i.e., we will aim at formalizing the required conditions as generally as possible.
%
%In principle, the result can also be used for training (because for many generative models it is sufficient that only derivatives w.r.t.\ a subset of %the parameters $\Theta$ vanish). In this contribution, we will focus on the derivation of the result itself, however, and on the conditions that have to %be satisfied for a generative model in order for the result to apply. 
%
%UNTIL HERE
One condition that we will require is that the distributions $p_{\Theta}(\zVec)$ and
$p_{\Theta}(\xVec\,|\,\zVec)$ are within the exponential family of distributions.
Furthermore, we will require conditions for the parameterization of a model in particular for the link between latents and observables.

%
%probabilistic sparse coding \citep[e.g.][]{OlshausenField1996}. 
%
%Also other hierarchical/deep models
%and other extensions of the basic graphical model structure (\ref{EqnGenModelIntro}) can be treated by using the result derived here
%or by using the here introduced treatment as basis for further generalizations.
%
% and (most) mixture models \citep[e.g.][]{McLachlanPeel2004}. Furthermore, extensions to hierarchical/deep and other extensions of the basic graphical model (\ref{EqnGenModelIntro}) are relatively straight-forward considering the proof of our main result. 

%For deep generative
%models with deterministic mapping between latents and observables 
%as well as (in generalized forms) also for hierarchical/deep and other extensions of the basic graphical model (\ref{EqnGenModelIntro}).
%
%
%can be computed from the parameter values alone (we discuss in Sec.\,\ref{SecDiscussion}).
%

\ \\[-2ex]
\noindent{\bf Relation to other contributions.}
Given an exponential family of distributions, the relation of its maximum likelihood parameter estimation to the (negative) entropy of the distribution family is well-known in mathematical statistics (e.g.\,\cite{BickelDoksum2015,Efron2023}). Some of these well-established results have been used for generative models whose joint distributions take the form of an exponential family of distributions (e.g.\,\cite{WainwrightJordan2008}). The decomposition of variational lower bounds (i.e., ELBOs) into sums of entropies has not been studied to a comparable extent. % (to the knowledge of the author).
But at least for idealized conditions
%\footnote{if the model distribution equals the data distribution, for $N\rightarrow{}\infty$, at the global optimum, and for variational distributions equal to posteriors (the ELBO is then equal to the log-likelihood)}
it is relatively straightforward to show that at stationary points the ELBO can be written as a sum of entropies (we will discuss such idealized conditions below).
% when referring to \cite{LuckeHenniges2012}).
%
%even without assuming distributions to be in the exponential family \citep[e.g.][]{LuckeHenniges2012}. However, 
%
At the same time, there is a long-standing interest in stationary points of variational learning that are reached under realistic conditions.
%
%for many generative models. %variational lower bounds.
%
For instance, likelihoods, lower bounds, and their stationary points have been studied in detail for (restricted) Boltzmann machines (e.g.\,\cite{WellingTeh2003}), elementary generative models (e.g.\,\cite{Frey1999,TippingBishop1999}) as well as for general graphical models \cite{FreemanWeiss2000,WeissFreeman2000,YedidiaEtAl2001,OpperSaad2001}. 
%
%Entropies have also been used 
%Turner, MacKay and Sahani asked the 
%
%
%Properties of stationary points are also of interest more generally, and studies of stationary points of loss functions
%defined by neural networks \citep[see][for references]{ChoromanskaEtAl2015} are evidence for their relevance in the field of deep learning.
%
More recently, stationary points of deep generative models have been of considerable interest. 
%For deep generative networks stationary points are, consequently, likewise of high interest. 
For instance, recent work \cite{LucasEtAl2019} linked phenomena like mode collapse in variational autoencoders (VAEs) to the stability of stationary points of ELBO-based learning; and other recent work \cite{DaiEtAl2018} analyzed a series of tractable special cases of VAEs to better understand convergence points of VAE learning. Notably both contributions emphasized the close relation of standard VAEs to principal component analysis (PCA; \cite{Roweis1998,TippingBishop1999}).
Theoretical results on different versions of PCA (e.g.\,\cite{TippingBishop1999,XuEtAl2010,HoltzmanEtAl2020}, for probabilistic, sparse, and robust PCA) can consequently be of direct relevance for VAEs and vice versa. Furthermore, different theoretical guarantees, e.g.\ for VAEs, have been derived for reconstruction errors \cite{MbackeEtAl2024, CheriefEtAl2019}, and, notably in our context, for convergence rates of VAEs (e.g., \cite{SurendranEtAl2024} and references therein).
%\textcolor{green}{Similarly, differentiability points of the limiting free energy are of particular theoretical interest in optimal Bayesian inference, as at such points the overlap can be shown to concentrate around the gradient of the limiting free energy. In particular,~\cite{} establishes this result for finite-rank tensor-product inference models with an additional linear observation channel.}
%Similarly, differentiability is important for characterizing stationary points of variational objectives. This makes differentiable points of such objectives of particular theoretical interest. In optimal Bayesian inference, for instance, the overlap can be shown to concentrate at differentiable points of the free energy. In particular,~\cite{} establishes this result for finite-rank tensor-product inference models with an additional linear observation channel.
Similarly, differentiability is central to the characterization of stationary points of variational objectives through their gradients. Consequently, differentiable points themselves are of particular theoretical interest. In optimal Bayesian inference, the overlap (see~\cite{Barbier2021}) can be shown to concentrate around the gradient of the (limiting) free energy. In particular,~\cite{ChenIssa2026} establishes this result for finite-rank tensor-product inference models with an additional linear observation channel.
%
%
%in terms of error bounds have been 
%
%
%We are not aware of results such as () for VAE variational bounds at stationary points, nor have there been (to the knowledge of the authors)
%any discussions speculating on the existence of closed-form results for standard VAEs. Presumably closed-form results have not been
%considered feasible because of the general non-linearities represented by the used DNNs. 

Standard VAEs and standard generative models such as probabilistic PCA (p-PCA) or probabilistic sparse coding (e.g.\,\cite{OlshausenField1996}) all
assume Gaussian distributed observables (and standard VAEs and p-PCA also assume Gaussian latents). %In such Gaussian cases,  
%
%Our derivations for VAEs will show that variational bounds are at convergence directly related
%to the entropies of those distributions that define VAEs. 
%
%the relation between the integrals of the variational lower bound and entropies has been of interest previously \citep{LuckeHenniges2012,DammEtAl2023}.
%
Likewise, another previous contribution on VAEs \cite{DammEtAl2023} assumes Gaussians for all distributions. And all previous contributions \cite{LuckeHenniges2012,DammEtAl2023,VelychkoEtAl2024} with results similar to Eqn.\,\ref{EqnTheoremIntro} derive them for models where at least the
observables are Gaussian distributed. % \citet[][]{VelychkoEtAl2024}.
% also use Gaussian observables but consider a sparse prior in the form of a Laplace distribution.
%\citep[also compare][]{LuckeHenniges2012}.
%
The relation of the ELBO to entropy sums in a not as recent contribution \cite{LuckeHenniges2012} was more general regarding the model distributions but
required conditions that are unrealistic in practice. Concretely, equality to entropy sums was shown when the following idealized conditions apply: (A)~when the limit of infinitely many data points is considered, (B)~when the model distribution can exactly match the data distribution, (C)~if the variational distributions match the posteriors, and (D)~when a global optimum is reached. The authors (of \cite{LuckeHenniges2012}) also discussed relaxations of these unrealistic assumptions (their Sec.\,6) but those relaxations relied on properties of specific models with Gaussian observables (p-PCA and sparse coding) when using expectation maximization (EM; \cite{DempsterEtAl1977}).
The study by Damm et al.\,\cite{DammEtAl2023} reported a result requiring Gaussian distributions but allowed for very general dependencies of the observables on the latent variables. More concretely, they focused on standard (Gaussian) VAEs, i.e., one specific (but very popular) class of generative models. For standard VAEs, their work shows equality to entropy sums also under realistic conditions (i.e., local optima, imperfect match of model assumptions and data, finite data sets etc.). 

In contrast to all such previous work, we in this work do not assume any properties specific to any generative model, i.e., given a generative model, none of its distributions will be considered to have a specific form such as being Gaussian. Instead, we will be interested in how generally the result of Eqn.\,\ref{EqnTheoremIntro} applies.

%\mycomment{Change here:}
We will first specify the required conditions for a generative model in Sec.\,\ref{SecGenModels}. Based on the conditions, we will then prove two theorems that state the equality of the ELBO and entropy sums at stationary points.
The theorems can be used to show, for a given concrete generative model, that its ELBO converges to an entropy sum. In a contribution applying the theorems proven in this work (see \cite{WarnkenEtAl2024}), it can thus be shown that the ELBOs of many very well-known generative models converge to entropy sums including: sigmoid belief networks (SBNs;\cite{Neal1992,SaulEtAl1996}), probabilistic PCA (\cite{Roweis1998,TippingBishop1999}), factor analysis (e.g.\,\cite{Thurstone1947}) and Gaussian mixture models (e.g.\,\cite{McLachlanPeel2004}). For standard (i.e., Gaussian) VAEs, convergence to entropy sums was shown previously \cite{LuckeEtAl2021} (with the final version of that work \cite{DammEtAl2023} already making use of the results presented in this work).%, and the here presented results show the deeper theoretical reason

%

%As examples of generative models for which the result of Eqn.\,\ref{EqnTheoremIntro}
%applies, we explicitly discuss sigmoid belief networks (SBNs;\cite{Neal1992,SaulEtAl1996}), probabilistic PCA (\cite{Roweis1998,TippingBishop1999}) and mixture models (e.g.\,\cite{McLachlanPeel2004}).
%
%, Factor Analysis \citep[e.g.][]{Everitt1984,BartholomewKnott1987} and mixture models \citep[e.g.][]{McLachlanPeel2004}.
% , sparse coding \citep[e.g.][]{OlshausenField1996}
%

%For the example of Gaussian VAEs we refer to \citet{DammEtAl2023}.
% who shows that convergence to entropy sums is
%a property that can apply for potentially very intricate non-linear relations between latents and observables.
%
%For standard Gaussian VAEs one implication of the equality to entropy sums is that closed-form expressions to compute the ELBO can be available at stationary points also in cases when the original ELBO is not closed-form \citep[see][]{DammEtAl2023}. Further such examples will be discussed later in Sec.\,\ref{SecExamples}.
%
%values can be computed at stationary points while the original ELBO
%is typically not closed-form \citep[see][]{LuckeEtAl2021}.
%
%
%
%
%
%
%
%However, the main object of our interest in this paper will be the main result and its proof.
%
%
%\ \\[-1ex]
%
%
%
%
%
%
%

\section{The Class of Considered Generative Models}

\label{SecGenModels}
We first define the class of generative models that we consider, i.e., the models and the properties that we require in order to show
convergence to entropy sums. We will also discuss two concrete examples and a counter-example. We keep focusing on generative models of the kind introduced above
(Eqn.\,\ref{EqnGenModelIntro}). Such models will be sufficient to communicate the concepts the proof relies on but generalizations will be discussed
in Sec.~\ref{SecDiscussion}. First, we will require a more specific notation for the parameterization of the generative model, however. \vspace{1ex}%\\[0ex]
%
%
%We will focus on generative models
%with one set of latent variables and one set of observed variables. It will be sufficient to consider such elementary graphical models
%to communicate the basic concepts the proof relies on. Later (in Sec. XXX), we will also point to extensions to more general graphical.
%
%We first consider generative models in their standard parameterization.\\[0ex]
%
%
%
%
%
%\noindent{\bf Definition A} (Generative Model)\\[1ex]
\begin{definition}[Generative Model]
    \label{def:Gen_Model}
%
%We consider generative models with one set of latent variables $\zVec$ and one set of observed variables $\xVec$. 
%Data is generated by first drawing a latent vector $\zVec$ from a prior distribution $\pPsi(\zVec)$. Given the latent vector
%$\zVec$ an observed variable $\xVec$ is then drawn from the conditional distribution $\pT(\xVec\,|\,\zVec)$ (which will be
%referred to as {\em noise distribution} or {\em noise model}). 
The generative models we consider have prior distribution $\pPsi$ with parameters $\PsiVec$, i.e.,
all parameters are arranged into one column vector with scalar entries. Analogously, the noise distribution $\pT$ is parameterized
by $\ThetaVec$ such that:
%
%and noise distribution $\pT$ are
%parameterized by two different sets of parameters $\PsiVec$ and $\ThetaVec$, respectively. We assume all the parameters for distributions to be arranged into one column vector each.  
%
\begin{eqnarray}
    \zVec &\sim& \pPsi(\zVec)\,, \label{EqnIntroPrior}\\ 
    \xVec &\sim& \pT(\xVec\,|\,\zVec)\,. \label{EqnIntroNoise}\vspace{-3ex}
\end{eqnarray}
\end{definition}
%\ \\
\noindent{}Arranging prior parameters and noise model parameters into column vectors will be important for further definitions and for the derivation of the main result. For our purposes, the definition of a generative model in terms of prior and noise model distributions will also be notationally more convenient than using the model's joint probability.  
%For mixture models we note that the latents are usually denoted (instead of by $\zVec$) by $c$ where $c\in\{1,\ldots,C\}$.

In general, we will adopt a somewhat less abstract mathematical notation than could have been used. For instance, we will not use the notation of Lebesgue integrals if
such a notation is not necessary. The motivation is that then the main derivations can be communicated using conventional integrals (which are more common
in the broader Machine Learning literature). Knowledge, e.g., on Lebesgue integrals, Legendre transformations and duality principles are not essential for the main
derivations. However, if we later (in Sec.\,\ref{SecGenEF}) cover the most general form of our main result, Lebesgue integrals and general measures will be used and will be required for our derivation.
%
%here made
%definitions and the derivations of results. We will, however, briefly discuss more abstract notations in Sec.\,\ref{SecDiscussion}.  
%

%Likewise, the rather
%classical description in terms of prior and noise model distribution will serve our notational purposes. Furthermore, the notation is 
%presumably more wide-spread than defining a graphical model via joint probabilisties with factorization according to the underlying
%graphical model. 

Many standard generative models are of the form as described in Definition \ref{def:Gen_Model}. Examples are those named in the introduction.
%
%probabilistic PCA, standard probabilistic sparse coding \citep[with Laplace prior][]{OlshausenField1996} or sigmoid belief networks \citep[SBNs][]{SBNCITE}. 
%
For our purposes, we will require that prior and noise distribution are of the exponential family
of distributions, and we will (at first) require constant base measures of the exponential family distributions.
For most standard generative models in Machine Learning of the form stated in Definition \ref{def:Gen_Model},
these conditions are satisfied. But there are exceptions that are not covered. Examples are, e.g., probabilistic
sparse coding models with mixture of Gaussians prior \cite{OlshausenMillman2000} or student-t prior \cite{BerkesEtAl2008},
i.e., models with a prior not in the exponential family.
%
%, which is the case for most standard generative models of the form of Definition \ref{def:Gen_Model}. But there are exceptions that are
%not (at least not directly) covered, for instance, sparse coding with mixture of Gaussians prior \citep[][]{OlshausenMillman2000}. 
%
%\mycomment{but point to generalization?}
%
The subset of generative models defined with the required exponential family distributions will be referred to as {\em exponential family generative models} or {\em EF generative models}.
%
%The class of all generative models with of the from given by Definition \ref{def:Gen_Model}
%defined using exponential family distributions (with constant base measure) will be
%referred to as {\em exponential family generative models} (or {\em EF generative models}).
%
%REMOVE?: We will denote the class of considered generative models as {\em exponential family generative models}.\vspace{1ex}
%
%For most such models (and all just named models), prior and model distributions are of the exponential family
%of distributions CITE. But not all generative models necessary use distributions of the exponential family. Examples for models that do not are sparse coding with mixture of Gaussians prior or student-t prior \citep[][]{TurnerEtAlXXXX} among others. For our purposes, we will require
%
%
%
%\ \\
%\ \\
%\ \\
%\noindent{\bf Definition B} (EF Generative Models)\\[1ex]
\begin{definition}[EF Generative Models]
    \label{def:EF_Gen_Model}
Consider a generative model as given by Definition \ref{def:Gen_Model}. We say the model is an {\em exponential family model (EF model)} if
there exist exponential family distributions $\pZetaVec(\zVec)$ (for the latents) and $\pEtaVec(\xVec)$ (for the observables)
such that the generative model can be reparameterized to take the following form:\vspace{-2ex}
\begin{eqnarray}
%
% \zVec &\sim& p_{\zetaVec(\Psi,\Theta)}(\zVec) \\ 
 \zVec &\sim& p_{\zetaVec(\PsiVec)}(\zVec)\,, \\ 
 \xVec &\sim& p_{\etaVec(\zVec;\,\ThetaVec)}(\xVec)\,,
\end{eqnarray}
where $\zetaVec(\PsiVec)$ maps the prior parameters $\PsiVec$ to the natural parameters of distribution $\pZetaVec(\zVec)$, and where $\etaVec(\zVec;\ThetaVec)$ maps $\zVec$ and the noise model parameters $\ThetaVec$ to the natural parameters of distribution $\pEtaVec(\xVec)$.
Furthermore, we demand that the exponential family distributions $\pZetaVec(\zVec)$ and $\pEtaVec(\xVec)$ have a constant base measure.%\vspace{1ex}
\end{definition}
\noindent{}The natural parameters of the prior can in principle also depend on $\ThetaVec$ because reparameterizations are conceivable that also involve part of the noise model parameters for the reparameterized prior. However, we will use the more intuitive but slightly more restricted definition above.

If a generative model defined by (\ref{EqnIntroPrior}) and (\ref{EqnIntroNoise}) is an EF generative model, it can be written as follows:\vspace{-1ex}
\begin{align}
p_{\zetaVec(\PsiVec)} (\zVec)  &= h(\zVec)\,  \exp\Big(  \zetaVecT_{(\PsiVec)} \vec{T}(\zVec) \,-\, A\big(\zetaVec(\PsiVec) \big) \Big)\,, \label{EqnExpFamA}\\
p_{\etaVec(\zVec;\,\ThetaVec)} (\xVec)  &=  h(\xVec)\,  \exp\Big(   \etaVecT_{(\zVec;\,\ThetaVec)} \vec{T}(\xVec) \,-\, A\big(\etaVec(\zVec;\,\ThetaVec)\big) \Big)\,, \label{EqnExpFamB}
\end{align}
where we additionally know that $h(\xVec)$ and $h(\zVec)$ are constant. %equal to two (postive) constants in $\RRR$.
% = h_o\in\RRR$ and $h(\xVec) = h_o\in\RRR$
The base measures $h(\cdot)$ as well as the log-partition functions $A(\cdot)$ and sufficient statistics $\TVec(\cdot)$ of latent and observable distributions are in general different, of course. To simplify notation, we do not use different symbols for these functions, however, because they will in the following be distinguishable from context (simply by their arguments).
%
%
%The dependence of the natural parameters of the prior also on $\Theta$ may at first not be intuitive, and will often not be required (for instance in the first example below, $\zetaVec(\Psi,\Theta)$ will only depend on $\Psi$). However, there will be prominent examples later, the reparameterize the original model using part of the noise model parameters for the prior distribution. 

For our derivation of the main result, it will not be sufficient for a generative model to be an EF generative model. An additional
property for the mappings to natural parameters (which includes how the latents link to observables) is required.
Before we define the property, we introduce further notations that we will require: For an EF generative model as given in Definition~\ref{def:EF_Gen_Model} with mappings to natural parameters given by $\zetaVec(\PsiVec)$ and $\etaVec(\zVec;\ThetaVec)$
we define the here used notation for Jacobian matrices. 
We denote by $\delll{\zetaVec(\PsiVec)}{\PsiVecT}$ the Jacobian matrix of the mapping~$\zetaVec(\PsiVec)$:
\begin{eqnarray}
%
%\IIT_{(\PsiVec)} &:=&\disT \Big(  \del{\PsiVec} \zeta_1(\PsiVec), \ldots, \del{\PsiVec} \zeta_K(\PsiVec) \Big)\,.
\delll{\zetaVec(\PsiVec)}{\PsiVecT}:=\Biggl(\delll{\zetaVec(\PsiVec)}{\Psi_1},\dots,\delll{\zetaVec(\PsiVec)}{\Psi_R}\Biggr),
\label{EqnIIMatrix}
\end{eqnarray}
where $R$ denotes the number of prior parameters $\PsiVec$.
Furthermore, we denote by $\delll{\etaVec(\zVec;\ThetaVec)}{\thetaVecT}$ the Jacobian matrix of the mapping $\etaVec(\zVec;\ThetaVec)$:
\begin{eqnarray}
\delll{\etaVec(\zVec;\ThetaVec)}{\thetaVecT}:=\Biggl(\delll{\etaVec(\zVec;\ThetaVec)}{\theta_1},\dots,\delll{\etaVec(\zVec;\ThetaVec)}{\theta_S}\Biggr),
\label{EqnJJMatrix}
\end{eqnarray}
for which we allow the Jacobian (\ref{EqnJJMatrix}) to be constructed using just a subset $\thetaVec$ of the parameters~$\ThetaVec$. The dimension of this subset is referred to as $S$.

\noindent We can now define the property we additionally require for EF generative models.
%
%\mycomment{REMOVE: Relations of the property to concepts such as minimal exponential
%family representations \citep[see, e.g.,][]{WainwrightJordan2008} and linear subspaces will briefly be discussed in Sec.\,\ref{SecDiscussion}.} \vspace{3ex}

%The property is related the concept of a minimal exponential family
%distribution \citep[][]{REFMINIMALEF} (we will elaborate further below). However, due to different relations potentially defined to relate
%prior to noise model in different models, the required property does not derive directly from minimality. We consequently state the required
%property here explicitly. It is formulated in the following definition and examples will show that it is relatively
%easily fulfilled.
%\ \\
%
%\noindent{\bf Definition C} (Parameterization Criterion)\\[1ex]
\begin{definition}[Parameterization Criterion]
    \label{def:Param_Crit}
     % (new version by Jan
%\vspace{-1ex}
%
Consider an EF generative model as given in Definition~\ref{def:EF_Gen_Model} with mappings to natural parameters given by $\zetaVec(\PsiVec)$ and $\etaVec(\zVec;\ThetaVec)$. 
%If $\thetaVec$
%is a proper subset of $\ThetaVec$ the full notation would be $\JJT_{(\zVec;\,\thetaVec,\ThetaVec)}$ but, for brevity, we take the last argument as
%implicit and use $\JJT_{(\zVec;\,\thetaVec)}$ throughout the paper.
We then say that the EF generative model fulfills the {\em parameterization criterion} if the following two properties hold:%\vspace{0ex}
\begin{itemize}
    \item[(A)] There exists a vectorial function $\alphaVec(\cdot)$ from the set of parameters $\PsiVec$ to $\RRR^R$ such that for all $\PsiVec$ holds:
%There is a ($R$-dim) vector $\vVec(\PsiVec)$ that may depend on the parameters $\PsiVec$ and fulfills the equation
%
%($K$-dimensional codomain) $\fVec(\Phi,\PsiVec)$ it holds: 
%
\begin{eqnarray}\label{EqnCondZeta}
%
%\IIT_{(\PsiVec)}\ \fVec(\Phi,\PsiVec) \ =\ \vec{0} &\Rightarrow&
%
%\int \JJT(\zVec;\Theta^*)\ \gVec(\zVec;\,\Phi^*,\Psi^*,\Theta^*) \,\mathrm{d}\zVec\ =\ 0 &\Rightarrow&
%
%\zetaVec^{\mathrm{T}}_{(\PsiVec)}\ \fVec(\Phi,\PsiVec)\ =\ 0\,.\label{EqnParaCondPrior} %\\[1ex]
%
%\int \etaVec^{\mathrm{T}}(\zVec;\Theta^*)\ \gVec(\zVec;\,\Phi^*,\Psi^*,\Theta^*) \,\mathrm{d}\zVec\ =\ 0\,.
%
\zetaVec(\PsiVec) &=& \delll{\zetaVec(\PsiVec)}{\PsiVec^\mathrm{T}} \ \alphaVec(\PsiVec).
\end{eqnarray}
\item[(B)] %If for any vectorial function $\gVec$ from parameter sets $\Phi$ and $\ThetaVec$ and from the latent space $\Omega_{\zVec}$ to the ($L$-dim) space of the natural parameters of the noise model, there exists a subset $\thetaVec$ of $\ThetaVec$ such that:
There exists an ($S$-dim) subset $\thetaVec$ of $\ThetaVec$ and a vectorial function $\betaVec(\cdot)$ from the space of parameters of the observable distribution $\ThetaVec$ to $\RRR^S$ such that for all $\zVec$ and $\ThetaVec$ holds:
%
%. If there exists a subset $\thetaVec$ of the parameters $\ThetaVec$ such that for
%any function $\gVec$ (at any parameter values $\Phi$ and $\ThetaVec$) holds:
%
%Furthermore, let $\gVec$ be a function from the latent space $\Omega_{\zVec}$ to the same domain as the natural parameters $\etaVec$ of the noise model.
%
%
\begin{eqnarray}\label{EqnCondEta}
%
%\int \JJT_{(\zVec;\,\thetaVec\,)}\ \gVec(\zVec;\Phi,\ThetaVec) \,\mathrm{d}\zVec\ =\ \vec{0} &\Rightarrow&
%
%\int \JJT(\zVec;\Theta^*)\ \gVec(\zVec;\,\Phi^*,\Psi^*,\Theta^*) \,\mathrm{d}\zVec\ =\ 0 &\Rightarrow&
%
%\int \etaVec^{\mathrm{T}}_{(\zVec;\ThetaVec)}\ \gVec(\zVec;\Phi,\ThetaVec) \,\mathrm{d}\zVec\ =\ 0\,.
%
%\int \etaVec^{\mathrm{T}}(\zVec;\Theta^*)\ \gVec(\zVec;\,\Phi^*,\Psi^*,\Theta^*) \,\mathrm{d}\zVec\ =\ 0\,.
%
\etaVec{(\zVec;\ThetaVec)} &=& \delll{\etaVec(\zVec;\ThetaVec)}{\thetaVecT} \ \betaVec(\ThetaVec).
\label{EqnParaCondNoise}
\end{eqnarray}
The subset $\thetaVec$ has to be non-empty ($S\geq{}1$) in order for the Jacobian to be defined. Furthermore, notice that $\betaVec(\ThetaVec)$
only depends on the parameters $\ThetaVec$ and not on the latent variable~$\zVec$.
\end{itemize}
\end{definition}
\noindent{}Part~A of the parameterization criterion is usually trivially fulfilled but is required for completeness.
For instance, the usual mappings from standard parameters to natural parameters of standard exponential family distributions (Bernoulli, Gaussain, Beta, Gamma etc.) can be shown
to fulfill part~A of the criterion.
%
%if the mapping $\zetaVec$ is one-to-one, the Jacobian is invertible such that $\fVec(\PsiVec)=\vec{0}$ (and Part~A is fulfilled).
%
Part~B can be more intricate because of the additional dependence of the natural parameters $\etaVec$ on the latent variable $\zVec$.%such that explicit derivations are required for a generative model to show that part~B of the criterion holds.
% are required of a given generative model. 
%
%Both parts of the criterion are related to minimal exponential family distributions but especially Part~(B) prevents
%a straight-forward relation. However, we will later discuss examples of properties of the function $\etaVec$ that
%allows for showing that the parameterization criterion is fulfilled.

Below, let us first discuss some examples of generative models. The examples aim at illustrating the
parameterization criterion. \vspace{-2ex}
%
%
%, which will turn out to be relatively easily satisfiable.
%\ \\
\ \\
%\noindent{\bf Example 1} (Simple SBN)\\[1ex]
\begin{Example}[Simple SBN]
    \label{ex:Simple_SBN}
Let us consider a first example, which will be a simple form of a sigmoid belief net (SBN; \cite{SaulEtAl1996}).\vspace{-1ex}
\begin{align}
 z     &\sim&\hspace{-3ex} \pPsi(z) &= \Bern(z;\,\pi), \ \mbox{with}\ 0<\pi<1, \label{EqnSSBNA}\\ 
 \xVec   &\sim&\hspace{-2ex} \pT(\xVec\,|\,z) &=  \Bern\big(x_1;\,\Scal(v\,z)\big)\ \Bern\big(x_2;\,\Scal(w\,z)\big),\ \textrm{where}\  \Scal(a)=\frac{1}{1+e^{-a}}\ ,\label{EqnSSBNB}
\end{align}
and where $\Bern(z;\,\pi) = \pi^z (1-\pi)^{1-z}$ is the standard parameterization of the Bernoulli distribution ($\pi$ is the probability of $z$
being one, $p(z=1)=\pi$). The same parameterization is used for the two Bernoulli distributions of the noise model (with the sigmoidal function setting the probability).
The prior parameters $\PsiVec$ just consist of the parameter $\pi\in(0,1)$. The noise model parameters consist of $v\in\RRR$ and $w\in\RRR$.
%, i.e., $\ThetaVec=(v,w)^\mathrm{T}$. 
%

The generative model (\ref{EqnSSBNA}) and (\ref{EqnSSBNB}) is an EF generative model because it can be reparameterized as follows:
\begin{align}
 z &\sim p_{\zeta(\pi)}(z)\,,                          &&\ \mbox{where}\ \  \zeta(\pi) = \log\big( \frac{\pi}{1-\pi} \big)\,, \label{EqnSSBNC}\\ 
 \xVec &\sim p_{\etaVec(z;\,\ThetaVec)}(\xVec)\,,  &&\ \mbox{where}\ \ThetaVec=\left(\begin{array}{c} v\\ w \end{array}\right)\ \mbox{and\ }\ \etaVec(z;\,\ThetaVec) = \left(\begin{array}{c} v\,z\\ w\,z \end{array}\right).\label{EqnSSBND}
\end{align}
$p_{\zeta}(z)$ is now the Bernoulli distribution in its exponential family form with natural parameters~$\zeta$. Likewise, $p_{\etaVec}(\xVec)$ is the Bernoulli distribution (for two observables) with natural parameters $\etaVec$. The standard sigmoidal function (\ref{EqnSSBNB}) is the reason for the function to natural parameters, $\etaVec(z;\ThetaVec)$, to take on a particularly concise form. % (see Appendix ... ).

For the generative model, the parameterization criterion of Definition~\ref{def:Param_Crit} is fulfilled:
From the parameterization as EF model, it follows for part~A that the Jacobian $\Inew$ of (\ref{EqnIIMatrix}) is the scalar $\del{\pi}\zeta(\pi)=\frac{1}{\pi(1-\pi)}>0$. 
Therefore, Eqn.~\ref{EqnCondZeta} is satisfied by $\alpha(\pi)=\pi(1-\pi)\zeta(\pi)$.
%
%The left-hand-side of (\ref{EqnParaCondPrior}) is, therefore, equivalent to $f(\Phi,\pi)=0$ such that the %right-hand-side of (\ref{EqnParaCondPrior}) is trivially satisfied.
%
%As the scalar is uneqal zero, we conclude that $f(\Phi,\pi)=0$ such that (\ref{EqnParaCondPrior}) is satisfied. 
For part~B of the criterion, we use all parameters $\ThetaVec$ to construct the Jacobian, i.e. $\thetaVec=\ThetaVec=(v,w)^{\mathrm{T}}$. % $\dell{\thetaVec}=\left(\begin{array}{c} \dell{v}\\[0.5ex] \dell{w} \end{array}\right)$
The Jacobian is
then a ($2\times{}2$)-matrix given by
\begin{align}
  \displaystyle\delll{\etaVec(\zVec;\ThetaVec)}{\thetaVecT}\,=\,\left(\begin{array}{cc} z & 0 \\ 0 & z \end{array}\right).
\end{align}
Hence, the vectorial function $\betaVec(\ThetaVec)=\left(\begin{array}{cc} v \\ w \end{array}\right)$
fulfills Eqn.~\ref{EqnCondEta},
%\begin{align}
%
%\sum_z \JJT_{(z;\thetaVec)}\ \gVec(z;\Phi,\ThetaVec) \ =\ \vec{0} \ \mbox{\ that for all $d$ applies}\ \ \sum_z z\ g_d(z;\Phi,\ThetaVec) &= 0\,, \label{EqnSSBNInterim}
%
%\end{align}
%
%so essentially $g_d(z;\Phi,\ThetaVec)=0$ (which we will not need explicitly, though). From (\ref{EqnSSBNInterim}) we conclude
%
%\begin{align}
%
%\sum_z \etaVecT_{(z;\ThetaVec)}\ \gVec(z;\Phi,\ThetaVec) \ =\ v \sum_z z\ g_1(z;\Phi,\ThetaVec) \,+\, w \sum_z z\ g_2(z;\Phi,\ThetaVec)\ =\ 0\,, \label{EqnSSBNInterimTwo}
%
%\end{align}
%
which shows part B of the criterion.
%
%%\noindent{}\TODO{Ich würde den Teil mit Part B anders schreiben:} For part B of the criterion, we use all parameters $\ThetaVec$ to compute the Jacobian. Hence, the Jacobian is the ($2\times{}2$)-matrix given by
%%$$\delll{\etaVec(\zVec;\ThetaVec)}{\thetaVecT}\,=\,\left(\begin{array}{cc} z & 0\\ 0 & z \end{array}\right)$$
%%and the vectorial function $\betaVec(\ThetaVec)=(v,w)^{\mathrm{T}}$ satiesfies equation~(\ref{EqnCondEta}), which shows part B of the criterion.
%
\end{Example}
%\ \\[0ex]
%\ \\
%\noindent{}
%
%\ \\
%\ \\
%\noindent{\bf Example 2} (Simple Factor Analysis)\\[1ex]
\begin{Example}[Simple Factor Analysis]
    \label{ex:SFA}
Let us consider as second example a simple form of factor analysis (FA; \cite{Everitt1984,Bishop2006}).
\begin{align}
 z &\sim &\pPsi(z) &= \NN\big(z;\,0, \tilde{\tau}  \big),\ &&\textrm{where}\  \tilde{\tau}>0\,,\label{EqnSFAA}\\
 \xVec &\sim &\pT(\xVec\,|\,z) &= \NN\big(\xVec;\,\wVec{}z, \Sigma  \big),\ &&\textrm{where}\  \Sigma = \left(\begin{array}{cc} \sit_1 & 0 \\ 0 & \sit_2 \end{array}\right)\,, \sit_1,\sit_2>0\,,\label{EqnSFAB} \\
 &&&&&\textrm{and\ where}\ \wVec\in\RRR^2\  \textrm{with}\ || \wVec || = 1\,. \nonumber
\end{align}
The parameter vectors are $\PsiVec = \Psi = \tilde{\tau}$ and $\ThetaVec = \left(\begin{array}{c} \sit_1 \\ \sit_2 \\ w_1 \\ w_2\end{array}\right)$, \vspace{-1.5ex} for which we abbreviated the variances $\tau^2$
by $\tilde{\tau}$ and $\sigma_d^2$ by $\sit_d$.

Although we use standard parameters (mean and variance) for the Gaussian distributions (and not natural parameters), the parameterization of the whole model is still somewhat unusual
for factor analysis because the prior also has a parameter. However, the vector $\wVec$ is of unit (here Euclidean) norm such that the model parameterizes the
same set of distributions as a more conventionally defined FA model. The reasons for the parameterization will
become clearer later on. 
%
%The vector $\wVec^o$ we assume fixed, i.e., to not be part of the parameter vector $\ThetaVec$ which we optimize. This will at first
%not be intuitive because $\wVec^o$ is, in general, important for the optimization of FA models. However, we will see that our result applies independently of the value of $\wVec^o$ such that it can be excluded from the parameters in the first place.

The model (\ref{EqnSFAA}) and (\ref{EqnSFAB}) can be recognized as an EF generative model (Definition~\ref{def:Gen_Model}) using the following natural parameters: \vspace{-3ex}
%
%Given the model we get with:
%
\begin{align}
%
%  \PsiVec = \Psi = \alt,\ \ \ThetaVec = \left(\begin{array}{c} \sit_1 \\ \sit_2 \end{array}\right),\ \  
%  
  \zetaVec(\tilde{\tau}) = \left(\begin{array}{c} 0 \\ -\frac{1}{2\tilde{\tau}} \end{array}\right)\ \ \mbox{and}\ \
  \etaVec(z; \ThetaVec) = \left(\begin{array}{c}  \frac{w_1 z}{\sit_1} \\[1mm]  \frac{w_2 z}{\sit_2} \\[1mm]  -\frac{1}{2\sit_1}  \\[1mm]  -\frac{1}{2\sit_2}\end{array}\right)\,.
\end{align}
By computing the Jacobian of the prior natural parameter mapping
\begin{align}
  \displaystyle\Inew=\delll{\zetaVec(\tilde{\tau})}{\tilde{\tau}}=\left(\begin{array}{cc}0 \\ \frac{1}{2\Tilde{\tau}^2}\end{array}\right)
\end{align}
and choosing $\alpha(\tilde{\tau})=-\Tilde{\tau}$, we can see directly that part~A of the parameterization criterion is fulfilled.
%
%\ \\[-5ex]
%For part A of the parameterization criterion we get
%
%\begin{align}
%
%0\,=\,\IIT_{(\PsiVec)}\ \fVec(\Phi,\PsiVec)\,=\,\big(0,  \frac{1}{2\alt^2} \big)\ \left(\begin{array}{c} f_1(\Phi,\alt) \\ f_2(\Phi,\alt) \end{array}\right)
%
%\ \ \Rightarrow\ \ \frac{1}{2\alt^2}\,f_2(\Phi,\alt)\,=\,0\ \ \Rightarrow\ \ f_2(\Phi,\alt)\,=\,0\,,
%
%\zetaVec^{\mathrm{T}}_{(\PsiVec)}\ \fVec(\PsiVec)\ =\ 0\,.\label{EqnParaCondPrior} %\\[1ex]
%
%\int \etaVec^{\mathrm{T}}(\zVec;\Theta^*)\ \gVec(\zVec;\,\Phi^*,\Psi^*,\Theta^*) \,\mathrm{d}\zVec\ =\ 0\,.
%
%
%\end{align}
%
%where the last implication step is due to $\alt>0$ and $\alt$ being finite. We can therefore conclude:
%
%\begin{align}
%
%   \zetaVecT_{(\PsiVec)}\ \fVec(\Phi,\PsiVec)\ =\ \zetaVecT_{(\alt)}\ \fVec(\Phi,\alt)\ =\ -\frac{1}{2\alt}\ f_2(\Phi,\alt) \ =\ 0\, \ \ \mbox{because $f_2(\Phi,\alt)$ is zero.}
%
%\end{align}
%
%because $f_2(\alt)$ is zero.
%
For part~B of the criterion, we choose as subset of $\ThetaVec$ the vector $\thetaVec=\left(\begin{array}{c}  \sit_1 \\ \sit_2 \end{array}\right)$, i.e.\,$\dell{\thetaVecT} = \big(\dell{\sit_1}, \dell{\sit_2}\big)$. The Jacobian is then the $(4\times 2)$-matrix
\begin{align}
\delll{\etaVec(\zVec;\ThetaVec)}{\thetaVecT}=
  \left(\begin{array}{cc}
    -\frac{w_1 z}{\sit_1^2} & 0\\
    0 & -\frac{w_2 z}{\sit_2^2}\\
    \frac{1}{2\sit_1^2}  & 0\\
    0 & -\frac{w_2 z}{\sit_2^2}
  \end{array}\right)
\end{align}
and $\betaVec(\ThetaVec)=\left(\begin{array}{c}-\sit_1 \\ -\sit_2\end{array}\right)$ satisfies Eqn.~\ref{EqnCondEta} such that the parameterization criterion is fulfilled.
\end{Example}
%\newpage
%\ \\
\ \\
%\noindent{}
%
%\noindent{\bf Example 3} (Counter-Example: Rigid SBN)\\[1ex]
\begin{Example}[Counter-Example: Rigid SBN]
    \label{ex:Rigid_SBN}
As an example of a model for which it can not be shown that the parameterization criterion is fulfilled, we go back to the simple SBN model (Example~\ref{ex:Simple_SBN}).
However, the natural parameters of the noise model now use only one parameter $v\in\RRR$ as follows:  
\begin{align}
 z &\sim &\pPsi(z) &= \Bern(z;\,\pi)\,, \label{EqnRSBNA}\\ 
 \xVec &\sim &\pT(\xVec\,|\,z) &=  \Bern\big(x_1;\,\Scal(v\,z)\big)\ \Bern\big(x_2;\,\Scal((v+1)\,z)\big),\label{EqnRSBNB}%\ \mathrm{where}\  \Scal(a)=\frac{1}{1+e^{-a}}\ ,\label{EqnRSBNB}
\end{align}
where $\Scal(a)=\frac{1}{1+e^{-a}}$, so $\ThetaVec=v$ (now a scalar) and $\etaVec(z;\,\ThetaVec) = \left(\begin{array}{c} v\,z\\ (v+1)\,z \end{array}\right)$.\\[1ex]
Part A of the parameterization criterion remains satisfied. For part B of the criterion, \mbox{$\thetaVec=\ThetaVec=v$} is the only non-empty subset of $\ThetaVec$. The corresponding Jacobian is the $(2\times{}1)$-matrix
\begin{align}
  \displaystyle\delll{\etaVec(\zVec;\ThetaVec)}{\thetaVecT}=\left(\begin{array}{cc} z \\ z \end{array}\right).
\end{align}
Consequently, there is no function $\beta(v)$ satisfying Eqn.~\ref{EqnCondEta} in Part~B of the parameterization criterion since $v\neq 1+v$ for all $v\in \RRR$.
%It consequently follows from
%
%\begin{align}
%
%\disT\sum_z \JJT_{(z;\thetaVec)}\ \gVec(z;\Phi,\ThetaVec) \ =\ \vec{0} \ \mbox{\ \ that}\ \ \sum_z \big( z\ g_1(z;\Phi,\ThetaVec)\,+\,z\ g_2(z;\Phi,\ThetaVec) \big) &= 0\,. \label{EqnSSBNCounterInt}
%
%\end{align}
%
%We therefore obtain:
%
%\begin{align}
%
%\disT\sum_z \etaVecT_{(z;\ThetaVec)}\ \gVec(z;\Phi,\ThetaVec) \,=\, v \sum_z \big( z\ g_1(z;\Phi,\ThetaVec) \,+\,z\ g_2(z;\Phi,\ThetaVec) \big)  \,+\, \sum_z z\ g_2(z;\Phi,\ThetaVec)
%                                                            \,=\, g_2(z;\Phi,\ThetaVec)\,.\nonumber% \label{EqnSSBNCounterTwo}
%
%\end{align}
%
%But $g_2(z;\Phi,\ThetaVec)$ is unequal to zero for arbitrary $\gVec(z;\Phi,\ThetaVec)$. So, in this case, we can not conclude that $\sum_z \etaVecT_{(z;\Phi,\ThetaVec)}\ \gVec(z;\Phi,\ThetaVec)=0$.\\
%
\end{Example}
%\ \\
\noindent{}The counter-example shows that the parameterization criterion is not necessarily fulfilled for a generative model. Other counter-examples would be simple FA (Example~\ref{ex:SFA}) for which one sigma is fixed (e.g.\ $\tilde{\sigma}_2$). Also generative models with the Poisson distribution would be counter-examples but for another reason: the Poisson distribution does not have a constant base measure (but see Sec.\,\ref{SecGenEF} for non-constant base measures). In general, however, one can observe the parameterization criterion to be satisfied for 
many (maybe most) conventional generative models. A series of concrete models and model classes is systematically addressed elsewhere \cite{WarnkenEtAl2024} and using the results we here derive. The contribution also includes the convential general forms of SBNs and Factor Analysis. 

Our general treatment of generative models in this and in the following sections makes assumptions on prior and noise distributions and their interrelation. In this context, let us note that we do {\em not} require the model's joint distribution to be of the exponential family (which would be a further restriction). 
%
%
%
 %We do note that 
%
%show that the parameterization criterion is fulfilled for the generative models mentioned in the introduction.
%We can now, n the next section we will 
%
%But first we can now, in the following section, prove equality to entropy sums for all EF generative models that satisfy the parameterization criterion. %, i.e., we state and prove our main result.
%:
%
%\begin{itemize}
%\item probabilistic PCA
%\item factor analysis
%\item most mixture models
%\item general shallow SBNs
%\item probabilistic sparse coding
%\item standard variational autoencoders
%\end{itemize}
\ \\
%
%
%this is just a comment
%
%
%
%\ \\

\section{Equality to Entropy Sums at Stationary Points}
\label{SecEntropySums}
Based on the preparations in Sec.\,\ref{SecGenModels}, we will now prove equality to entropy sums for all EF generative models that satisfy the parameterization criterion.
Let us, first, consider the optimization of a generative model of Definition~\ref{def:Gen_Model} given a set of $N$ data points~$\xVecN$. When
considering the ELBO objective (\ref{EqnFFIntro}) we, as usual, assume the variational parameters $\Phi$ to represent a set of parameters different from $\PsiVec$ and $\ThetaVec$. The variational distributions $\qn_{\Phi}(\zVec)$ are usually defined to approximate posterior distributions of the generative model well but here we will not require any conditions on the distributions (and we will not use a vector notation for $\Phi$). Given distributions $\qn_{\Phi}(\zVec)$, the ELBO is given by:
\begin{align}
%
% \FF(\Phi,\PsiVec,\ThetaVec) \nonumber\\
\FF(\Phi,\PsiVec,\ThetaVec) &=  \textstyle\frac{1}{N}\sum_{n} \int \qn_{\Phi}\!(\zVec) \log\!\big( \pT(\xVecN\,|\,\zVec)\,p_{\PsiVec}(\zVec) \big)\hspace{0.5ex} \mathrm{d}\zVec   \\
&\hspace{15mm}\textstyle-\, \frac{1}{N}\sum_{n} \int \qn_{\Phi}\!(\zVec) \log\!\big( \qn_{\Phi}\!(\zVec) \big) \mathrm{d}\zVec \nonumber\\
                 &=  \textstyle\frac{1}{N}\sum_{n} \int \qn_{\Phi}\!(\zVec) \log\!\big( \pT(\xVecN\,|\,\zVec) \big)\hspace{0.5ex} \mathrm{d}\zVec\label{EqnFFStandardKL} \\
                 &\hspace{15mm}\textstyle-\, \frac{1}{N}\sum_{n} D_{\mathrm{KL}}\big[ \qn_{\Phi}\!(\zVec),p_{\PsiVec}(\zVec) \big]\,, \nonumber
\end{align}
where the first and second line represent the most common standard forms. The only unusual part of our notation is the use of two separate symbols for prior and noise model parameters ($\PsiVec$ and $\ThetaVec$) and a notation that denotes them as vectors. For our purposes, and for generative models as defined in Definition~\ref{def:Gen_Model}, we
first decompose the ELBO into three summands as follows: % (which is also common):
%
%the evidence lower bound \citep[ELBO;][]{} a.k.a.\ variational free energy CITES. For our purposes, we decompose the ELBO into three summands as follows:
%
\begin{align}
\FF(\Phi,\PsiVec,\ThetaVec) &=  \FF_1(\Phi) - \FF_2(\Phi,\PsiVec) - \FF_3(\Phi,\ThetaVec),\ \textrm{\ \ where}\label{EqnFFMain}\\
\FF_1(\Phi)\phantom{i}\      &=   \textstyle-\, \frac{1}{N}\sum_{n} \int \qn_{\Phi}\!(\zVec) \log\!\big( \qn_{\Phi}\!(\zVec) \big) \mathrm{d}\zVec,\label{EqnFFOne}\\
\FF_2(\Phi,\PsiVec)\ &= -\, \textstyle\frac{1}{N}\sum_{n} \int \qn_{\Phi}\!(\zVec) \log\!\big( p_{\PsiVec}(\zVec) \big)\hspace{0.5ex} \mathrm{d}\zVec,\label{EqnFFTwo}\\
\FF_3(\Phi,\ThetaVec)\ &= -\, \textstyle\frac{1}{N}\sum_{n} \int \qn_{\Phi}\!(\zVec) \log\!\big( \pT(\xVecN\,|\,\zVec) \big) \mathrm{d}\zVec.\label{EqnFFThree}
\end{align}
To finalize our notational preliminaries, we will use the {\em aggregated posterior} (e.g.\,\cite{MakhzaniEtAl2015,TomczakWelling2018,AnejaEtAl2021}) as an abbreviation, i.e., we define (noting that the entity is an {\em approximation} to the posterior average):\vspace{-2ex}
\begin{eqnarray}
\qBar_{\Phi}(\zVec) &=& \frac{1}{N} \sum_{n=1}^N \qn_{\Phi}(\zVec)\,.
\label{EqnAgPost}
\end{eqnarray}
%
% add a comment
%
For the proof of our main result, we will require the parameterization criterion given in Definition~\ref{def:Param_Crit}. As a preparation,
we will show a property that applies for the (transposed) Jacobians $\delll{\zetaVecT(\PsiVec)}{\PsiVec}$ and $\delll{\etaVecT(\zVec;\ThetaVec)}{\thetaVec}$ if the parameterization criterion in Definition~\ref{def:Param_Crit} is fulfilled.
The transposed Jacobians are defined in a notation consistent with (\ref{EqnIIMatrix}) and (\ref{EqnJJMatrix}). Concretely, the transposed Jacobian of $\zetaVec(\PsiVec)$ is defined as:\vspace{0ex}
\begin{align}
    \label{lem:ITmatrix}
    \delll{\zetaVecT(\PsiVec)}{\PsiVec} := \Big(\frac{\partial}{\partial \PsiVec}\zeta_1(\PsiVec),\ldots,\frac{\partial}{\partial \PsiVec}\zeta_K(\PsiVec)\Big)\,.\vspace{0ex}
\end{align}
The transposed Jacobian of $\etaVec(\zVec;\ThetaVec)$ constructed w.r.t.\ $\thetaVec$ (as a subset of $\ThetaVec$) is defined by:\vspace{0ex}
%
%For the next Lemma we will use transposed Jacobians of the functions ... and ... . In a notation consistent with (\ref{kkkkkk}), the transposed Jacobian of $\etaVecT(\zVec;\ThetaVec)$ is defined by:
%
\begin{align}
  \label{lem:JTmatrix}
  \delll{\etaVecT(\zVec;\ThetaVec)}{\thetaVec}:=\Big(\frac{\partial}{\partial\thetaVec\,}\eta_1(\zVec;\ThetaVec),\ldots,\frac{\partial}{\partial\thetaVec\,}\eta_L(\zVec;\ThetaVec)\Big).\vspace{0ex}
\end{align}
%
%\mycomment{Text überarbeiten}
%
%In the proof of our main result we will not directly use the parameterization criterion given in Definition~\ref{def:Param_Crit}. Instead, we will use a criterion that is more difficult to check in practical applications. However, this criterion can be easily inferred from Definition~\ref{def:Param_Crit}.
%
The following lemma will now derive a property from the parameterization criterion of Definition~\ref{def:EF_Gen_Model} that we will need for the proof of our main result. 
%
%\ \\[-2ex]
%\ \\
%\ \\
\begin{lemma}
  \label{lem:ParamCrit}
Consider an EF generative model as given by Definition~\ref{def:EF_Gen_Model}, and let the dimensionalities of the natural parameter vectors $\zetaVec$ and $\etaVec$ be $K$ and $L$, respectively. Let further $\delll{\zetaVecT(\PsiVec)}{\PsiVec}$ and $\delll{\etaVecT(z;\ThetaVec)}{\thetaVec}$ denote the transposed Jacobians in (\ref{lem:ITmatrix}) and (\ref{lem:JTmatrix}), respectively.%\pagebreak

If the EF generative model fulfills the parameterization criterion (Definition~\ref{def:Param_Crit}), then the following two statements apply:
%
%then the following two statements are true:
%
%Then, under the assumption that the parameterization criterion (Definition~\ref{def:Param_Crit}) is fulfilled, the following two statements apply:
%
\begin{itemize}
  \item[(A)] For any vectorial function $\fVec(\Phi,\PsiVec)$ from the sets of parameters $\Phi$ and $\PsiVec$ to the ($K$-dim) space of natural parameters of the prior it holds:
    \begin{align}
      \label{EqnLemmaParamCritA}
      \delll{\zetaVecT(\PsiVec)}{\PsiVec}\fVec(\Phi,\PsiVec)\ =\ \vec{0}\ \Rightarrow\ \zetaVecT_{(\PsiVec)}\fVec(\Phi,\PsiVec)\ =\ 0.
    \end{align}
  \item[(B)]For any variational parameter $\Phi$ there is a non-empty subset $\thetaVec$ of $\ThetaVec$ such that the following applies for all vectorial functions $\gVec_1(\zVec;\ThetaVec),\ldots,\gVec_N(\zVec;\ThetaVec)$ from the parameter set $\ThetaVec$ and from the latent space $\Omega_{\zVec}$ to the ($L$-dim) space of natural parameters of the noise model:
  \begin{align}
    \sum_{n=1}^{N} \int \qn_{\Phi}(\zVec)\,\delll{\etaVecT(z;\ThetaVec)}{\thetaVec}\,\gVec_n(\zVec;\ThetaVec)\,\mathrm{d}\zVec &= \vec{0}\label{EqnLemmaParamCritB1}\\
    \ \Rightarrow\ \sum_{n=1}^{N}\int \qn_{\Phi}(\zVec)\,\etaVecT_{(\zVec;\ThetaVec)}\,\gVec_n(\zVec;\ThetaVec)\,\mathrm{d}\zVec &= 0.\phantom{iiiii}\nonumber %\label{EqnLemmaParamCritB1}
  \end{align}
\end{itemize}
In the case when $\zVec$ is a discrete latent variable, the integrals in (\ref{EqnLemmaParamCritB1}) become
a sum over all discrete states of $\zVec$.\vspace{1ex}
%
%%For an EF generative model as in definition B, the following two statements are equivalent:
%
%%\begin{itemize}
%%  \item[(A)] There is a vectorial function $\vVec(\PsiVec)$ form the set of parameters $\PsiVec$ to the ($R$-dim) space of $\PsiVec$ that satisfies the equation
%%  \begin{eqnarray}
%%    \zetaVec(\PsiVec) = \delll{\zetaVec(\PsiVec)}{\PsiVecT}\vVec(\PsiVec).
%%  \end{eqnarray}
%%  \item[(B)] For any vectorial function $\fVec$ from the set of parameters $\PsiVec$ to the ($K$-dim) space of natural parameters of the prior it holds:
%%  \begin{eqnarray}
%%    \delll{\zetaVecT(\PsiVec)}{\PsiVec}\fVec(\PsiVec)\ =\ \vec{0}&\Rightarrow& \zetaVecT(\PsiVec)\fVec(\PsiVec)\ =\ 0.
%%  \end{eqnarray}
%%\end{itemize}
\end{lemma}
\begin{proof}
To verify the first part of the lemma, we choose a function $\alphaVec(\PsiVec)$ satisfying Eqn.~\ref{EqnCondZeta}. This is possible since we assume that the parameterization criterion (Definition~\ref{def:Param_Crit}) is fulfilled. Moreover, consider an arbitrary vectorial function $\fVec(\Phi,\PsiVec)$ that satiesfies
\begin{align}
  \delll{\zetaVecT(\PsiVec)}{\PsiVec}\fVec(\Phi,\PsiVec) = \vec{0}.
\end{align}
%
%\ \\[-4ex]
Then we can conclude:
\begin{align}
  \zetaVecT_{(\PsiVec)}\fVec(\Phi,\PsiVec) = \alphaVec^{\mathrm{T}}(\PsiVec) \delll{\zetaVecT(\PsiVec)}{\PsiVec}\fVec(\Phi,\PsiVec) = 0.
\end{align}
Consequently, the first part of the lemma is verified.

For the proof of the second part, let $\Phi$ be any variational parameter. We consider a subset $\thetaVec$ of $\ThetaVec$ and a function $\betaVec(\ThetaVec)$ satisfying Eqn.~\ref{EqnCondEta}. Then all vectorial functions $\gVec_1(\zVec;\ThetaVec),\ldots,\gVec_N(\zVec;\ThetaVec)$ fulfilling the equation\vspace{-1.5ex}
\begin{align}
  \sum_{n=1}^{N} \int \qn_{\Phi}(\zVec)\delll{\etaVecT(\zVec;\ThetaVec)}{\thetaVec}\gVec_n(\zVec;\ThetaVec)\,\mathrm{d}\zVec = \vec{0}
\end{align}
%
%\ \\[-3ex]
also fulfill the equation\vspace{-0.5ex}
\begin{align}
  \sum_{n=1}^{N}\int \qn_{\Phi}(\zVec) \etaVecT_{(\zVec;\ThetaVec)}\gVec_n(\zVec;\ThetaVec)\,\mathrm{d}\zVec =  \betaVec^{\mathrm{T}}(\ThetaVec)\sum_{n=1}^{N} \int \qn_{\Phi}(\zVec)\delll{\etaVecT(\zVec;\ThetaVec)}{\thetaVec}\gVec_n(\zVec;\ThetaVec)\,\mathrm{d}\zVec = 0\,,
\end{align}
which proofs the second part of the lemma. Furthermore, it is easy to see that the proof proceeds analogously in the discrete case, i.e., if the integrals are replaced by sums.
\end{proof}
\noindent{}As we will only use Lemma~\ref{lem:ParamCrit} and not directly the parameterization criterion (Definition~\ref{def:Param_Crit}) to show convergence of the ELBO to entropy sums, the question may arise whether the two statements in Lemma~\ref{lem:ParamCrit} are actually equivalent to the parameterization criterion given in Definition~\ref{def:Param_Crit}. 
\noindent{}For the proof of our main result, we will require Eqns.\,\ref{EqnLemmaParamCritA} and \ref{EqnLemmaParamCritB1} which follows (according to Lemma~\ref{lem:ParamCrit}) from Definition~\ref{def:Param_Crit}.
Therefore, instead of verifying for a given generative models that it fulfills the parameterization condition (Definition~\ref{def:Param_Crit}), we could also verify
Eqns.\,\ref{EqnLemmaParamCritA} and \ref{EqnLemmaParamCritB1} directly. This was done in previous work \cite{DammEtAl2023} but is much more intricate (see, e.g., their
Appendix~B), which makes Definition~\ref{def:Param_Crit} preferable. A disadvantage of Definition~\ref{def:Param_Crit} could be that it is more restrictive, i.e., that there
are generative model that do fulfill Eqns.\,\ref{EqnLemmaParamCritA} and \ref{EqnLemmaParamCritB1} but not Definition~\ref{def:Param_Crit}.
However, given mild assumptions, the conditions of Eqns.\,\ref{EqnLemmaParamCritA} and \ref{EqnLemmaParamCritB1} can, indeed, be shown to be equivalent to the conditions
of Definition~\ref{def:Param_Crit}. We provide the proof in Appendix~\ref{app:ParaCrit}. 
%
%Hence, in practice, Definition~\ref{def:Param_Crit} can be used to verify if a generative model fulfills the required condition, and Definition~\ref{def:Param_Crit} is much easier to verify than the more intricate properties represented by (\ref{EqnLemmaParamCritA}) and (\ref{EqnLemmaParamCritB1}).
%
In this context, we do point out that also another criterion can be used as a sufficient condition for (\ref{EqnLemmaParamCritA}) and (\ref{EqnLemmaParamCritB1}). On the one hand,
that criterion (as provided by \cite{VelychkoEtAl2024}, their Appendix~A) is not equivalent to Eqns.\,\ref{EqnLemmaParamCritA} and \ref{EqnLemmaParamCritB1} (it is more restrictive). But, on the other hand, it has the benefit of being more constructive.
% \citet[see Appendix \mycomment{XXX} of][for a discussion]{VelychkoEtAl2024}. \mycomment{Appendix mal groß mal klein}
%
%
%
% and (\ref{EqnLemmaParamCritB2}). By virtue of Lemma~\ref{lem:ParamCrit} and Appendix ..., no .
%
%
%
%Thus, using Definition~~\ref{def:Param_Crit} 
%
 %virtue of Lemma~\ref{lem:ParamCrit} and Appendix ..., verifying if a generative model fulfills the condition for 
%
%the parameterization condition is at the same time 
%
\\[0.5ex]

%we will show that under a few mild additional assumptions there is indeed an equivalence. So, we do not lose any models by using the parameterization criterion (Definition~\ref{def:Param_Crit}) anstead of the two statements from Lemma~\ref{lem:ParamCrit}.}\\
%
%

\noindent{}We can now state and prove the main result. %To formulate the result concisely, we will use the {\em aggregates posterior} \citep[][]{} as an abbreviation, i.e., we define:
\ \\[-1ex]
%
% \noindent{\bf Theorem} (Sum of Entropies)\\[1ex]
\begin{mytheorem}[Equality to Entropy Sums]
    \label{th:Sum_of_Entr}
If $\pPsi(\zVec)$ and $\pT(\xVec\,|\,\zVec)$ is an EF generative model (Definition~\ref{def:EF_Gen_Model}) that
fulfills the parameterization criterion of Definition~\ref{def:Param_Crit} then at all stationary points of the ELBO (\ref{EqnFFStandardKL})
it applies that\vspace{-2ex}
%variational lower-bound $\FF(\Phi,\PsiVec,\ThetaVec)$ applies:
%
% $\Theta=\Theta^*$: 
%
\begin{eqnarray}
%
%\FF(\Phi,\PsiVec,\ThetaVec) \hspace{-1ex}&=&\hspace{-1ex}  \frac{1}{N}\sum_{n=1}^N \HH[\qPhiN(\zVec)]  \phantom{ii} \phantom{\sum_{n=1}^N}\hspace{-3ex}-\ \HH[\,\pPsi(\zVec)] \phantom{ii}   -\ \frac{1}{N} \sum_{n=1}^N \EEE{\qn_{\Phi}}{ \HH[\,\pT(\xVec\,|\,\zVec)] }\,. \phantom{\small{}ix}  \\
%
\hspace{5mm}\FF(\Phi,\PsiVec,\ThetaVec) &=&  \frac{1}{N}\sum_{n=1}^N \HH[\qPhiN(\zVec)]  \phantom{ii} \phantom{\sum_{n=1}^N}\hspace{-3ex}-\ \HH[\,\pPsi(\zVec)] \phantom{ii}   -\ \EEE{\;\qBar_{\Phi}}{ \HH[\,\pT(\xVec\,|\,\zVec)] }\,. \phantom{\small{}ix} 
\label{EqnTheoremSoE}
%
%\FF(\Phi,\Theta) \hspace{-1ex}&=&\hspace{-1ex} \underbrace{\frac{1}{N}\sum_{n=1}^N \HH[\qPhi(\zVec;\xVecN)]}\phantom{ii}\ \underbrace{\phantom{\sum_{n=1}^N}\hspace{-3ex}-\ \HH[\pT(\zVec)]}\phantom{ii}  \underbrace{-\ \frac{1}{N} \sum_{n=1}^N \EEE{\qn_{\Phi}}{ \HH[\pT(\xVec\,|\,\zVec)] }}\phantom{\small{}ix} %
%\label{HypoVAE}\\
%
%\FF(\Phi,\Theta) \hspace{-1ex}&=&\hspace{-1ex} \underbrace{\frac{1}{N}\sum_{n=1}^N \HH[\qPhi(\zVec;\xVecN)]}_{\mbox{\small encoder}}\phantom{ii}\ \underbrace{\phantom{\sum_{n=1}^N}\hspace{-3ex}-\ \HH[\pT(\zVec)]}_{\mbox{\small prior}}\phantom{ii}  \underbrace{-\ \frac{1}{N} \sum_{n=1}^N \EEE{\qn_{\Phi}}{ \HH[\pT(\xVec\,|\,\zVec)] }}_{\mbox{\small decoder}}. %
%\nonumber
%\label{HypoVAE}\\
%
%
%\FF(\Phi,\Theta) \hspace{-1ex}&=&\hspace{-1ex} \overbrace{\frac{1}{N}\sum_{n=1}^N \HH[\qPhi(\zVec;\xVecN)]}_{\mbox{\small encoder}}\phantom{ii}\ \overbrace{\phantom{\sum_{n=1}^N}\hspace{-3ex}-\ \HH[\pT(\zVec)]}_{\mbox{\small prior}}\phantom{ii}  \overbrace{-\ \frac{1}{N} \sum_{n=1}^N \EEE{\qn_{\Phi}}{ \HH[\pT(\xVec\,|\,\zVec)] }}_{\mbox{\small decoder}}. \label{HypoVAE}
%
%
\end{eqnarray}
\end{mytheorem}
%
%Equation \ref{EqnTheoremSoE} 
%\ \\[2ex]
%\ \\
\begin{proof}
The first summand of the ELBO in Eqn.\,\ref{EqnFFMain} is already given in the form of an (average) entropy ($\FF_1(\Phi) = \frac{1}{N}\sum_{n} \HH[\qPhiN(\zVec)]$).
To prove the claim of Theorem~\ref{th:Sum_of_Entr}, we will (at stationary points) show equality to entropies for the second and the third summand of Eqn.\,\ref{EqnFFMain}; and we will show equality to the corresponding entropies separately for the two terms.

%\ \\
\ \\
%\newpage
\noindent\underline{$\FF_2(\Phi,\PsiVec)$ at stationary points}\\[1ex]
%
% so we start by considering the second summand (Eqn.\,\ref{EqnFFTwo}). 
\noindent{}We have assumed that the generative model is an EF generative model. Therefore we can rewrite $\FF_2(\Phi,\PsiVec)$ of Eqn.\,\ref{EqnFFTwo} using the reparameterization of $p_{\PsiVec}(\zVec)$ in terms of the exponential family distribution $p_{\zetaVec(\PsiVec)}(\zVec)$:
\begin{align}
\FF_2(\Phi,\PsiVec)\ &= -\, \textstyle\frac{1}{N}\sum_{n} \int \qn_{\Phi}\!(\zVec) \log\!\big( p_{\PsiVec}(\zVec) \big)\hspace{0.5ex} \mathrm{d}\zVec\label{EqnFFSecond}\\
                    &= \phantom{-}\, \textstyle\frac{1}{N}\sum_{n} \int \qn_{\Phi}\!(\zVec) \Big( - \log\!\big( p_{\zetaVec(\PsiVec)}(\zVec) \big) \Big) \hspace{0.5ex} \mathrm{d}\zVec\,.\nonumber
\end{align}
%
%As $p_{\zetaVec\,}(\zVec)$ is an exponential family distribution, 
%
Using (\ref{EqnExpFamA}), the negative logarithm of $p_{\zetaVec\,}(\zVec)$ as exponential family distribution can be written as:
\begin{align}
-\log\big( p_{\zetaVec(\PsiVec)} (\zVec) \big) &= A\big(\zetaVec(\PsiVec)\,\big) \,-\, \zetaVecT_{(\PsiVec)} \vec{T}(\zVec)  \,-\, \log\big( h(\zVec) \big)\,, \label{EqnLogJoint}
\end{align}
where $A$ is the partition function, $\vec{T}$ the sufficient statistics and $h$ is the base measure.
We will later require the derivative of the (negative) log probability w.r.t.\ $\PsiVec$. The derivative is given by:
\begin{align}
\dell{\PsiVec} \Big( -\log\big( p_{\zetaVec(\PsiVec)} (\zVec) \big) \Big) &= \ITnew \Big(  \AVec'\big(\zetaVec(\PsiVec)\big)\,-\,\vec{T}(\zVec)  \Big)\,,\label{EqnDelLog}\\
                   &\ \ \ \mbox{where}\ \ \AVec'(\zetaVec(\PsiVec)\,):=\dell{\zetaVec}\,A(\zetaVec\,)\Big|_{\zetaVec=\zetaVec(\PsiVec)}\label{EqnAPrime}
%
%A\big(\zetaVec(\PsiVec)\,\big) \,-\, \zetaVecT_{(\PsiVec)} \vec{T}(\zVec)  \,-\, \log\big( h(\zVec) \big)\,. \label{EqnLogJoint}
%
\end{align}
and where $\ITnew$ is the transposed Jacobian matrix defined in (\ref{lem:ITmatrix}).

We will show that the summand $\FF_2(\Phi,\PsiVec)$ will at stationary points be equal to the prior entropy. Therefore, we first rewrite $-\log\!\big( p_{\zetaVec(\PsiVec)}(\zVec) \big)$ using the prior entropy. By using the definition of the entropy
\begin{align}
\HH[\,p_{\zetaVec(\PsiVec)}(\zVec)] &= \int p_{\zetaVec(\PsiVec)}(\zVec) \Big( -\log\big( p_{\zetaVec(\PsiVec)} (\zVec) \big) \Big)\mathrm{d}\zVec
\end{align}
and by using (\ref{EqnLogJoint}) we obtain:
\begin{align}
\HH[\,\pPsi(\zVec)] &= \HH[\,p_{\zetaVec(\PsiVec)}(\zVec)] = A\big( \zetaVec(\PsiVec)  \big) \,-\, \zetaVecT_{(\PsiVec)} \AVec'\big(\zetaVec(\PsiVec)\big) \,-\, B(\zetaVec(\PsiVec))\, \label{EqnEntropyExpression}\\[1ex]
\Rightarrow\ A\big( \zetaVec(\PsiVec)  \big)  &= \HH[\,p_{\zetaVec(\PsiVec)}(\zVec)]  \,+\, \zetaVecT_{(\PsiVec)} \AVec'\big(\zetaVec(\PsiVec)\big) \,+\, B(\zetaVec(\PsiVec))\,, \label{EqnAasH}\\[1ex]
                    &\mbox{where}\ \ B(\zetaVec\,):= \disT\int p_{\zetaVec\,}(\zVec) \,\log\big(h(\zVec)\big)\, \mathrm{d}\zVec \label{EqnBPrime}\,. %\,=\, h_o\,. 
\end{align}
%
%The last equality for $B(\zetaVec)$ follows if we use our assumption of a constant base measure $h(\zVec)=h_o$. 
%
%We now use (\ref{EqnEntropyExpression}) to replace the partition function $A\big(\zetaVec\,\big)$ by an expression
%containing the entropy:
 %
%\begin{align}
%
%
%A\big( \zetaVec(\PsiVec)  \big)  &= \HH[\,p_{\zetaVec(\PsiVec)}(\zVec)]  \,+\, \zetaVecT_{(\PsiVec)} \AVec'\big(\zetaVec(\PsiVec)\big) \,+\, B(\zetaVec(\PsiVec)) \label{EqnAasH}
%\end{align}
%
By combining (\ref{EqnLogJoint}) and (\ref{EqnAasH}), the negative logarithm of the prior probability becomes:
\begin{align}
- \log\!\big( p_{\zetaVec(\PsiVec)}(\zVec)\big) &= \HH[\,p_{\zetaVec(\PsiVec)}(\zVec)] \,+\, \zetaVecT_{(\PsiVec)} \big( \AVec'(\zetaVec{(\PsiVec)})\,-\,\vec{T}(\zVec) \big)
\,+\, B(\zetaVec(\PsiVec))\,-\,\log\big(h(\zVec)\big)\nonumber\\
  &= \HH[\,p_{\zetaVec(\PsiVec)}(\zVec)] \,+\, \zetaVecT_{(\PsiVec)} \big( \AVec'(\zetaVec{(\PsiVec)})\,-\,\vec{T}(\zVec) \big)\,, \label{EqnNegLog}
%
%\,+\, B(\zetaVec(\PsiVec))\,-\,\log\big(h(\zVec)\big)\,, 
%
%\ \mbox{where}\ \AVec'(\zetaVec)\ \mbox{as in Eqn.\,\ref{EqnAPrime}}.
%
\end{align}
where $\AVec'(\zetaVec)$ is defined as in Eqn.\,\ref{EqnAPrime}, and where the last two terms have cancelled because we assumed a constant base measure $h(\zVec)$.
%, the last two terms cancel.
Inserting the negative logarithm (\ref{EqnNegLog}) back into (\ref{EqnFFSecond}), we therefore obtain for the second summand of the ELBO:
%
%The last two terms that depend on the base measure $h(\zVec)$ of the prior cancel because we assumed constant base measure. We therefore obtain for $\FF_2(\Psi,\Theta)$:
%
%
\begin{align}
%
%& \FF_2(\Phi,\PsiVec)\\
\FF_2(\Phi,\PsiVec) &= \textstyle\frac{1}{N}\sum_{n} \int \qn_{\Phi}\!(\zVec) \Big( \HH[\,p_{\zetaVec(\PsiVec)}(\zVec)] \,+\, \zetaVecT_{(\PsiVec)} \big( \AVec'(\zetaVec_{(\PsiVec)})\,-\,\vec{T}(\zVec) \big)  \Big) \hspace{0.5ex} \mathrm{d}\zVec\nonumber\\
                   &= \textstyle\HH[\,p_{\zetaVec(\PsiVec)}(\zVec)] \,+\, \zetaVecT_{(\PsiVec)} \frac{1}{N}\sum_{n} \int \qn_{\Phi}\!(\zVec)  \big( \AVec'(\zetaVec_{(\PsiVec)})\,-\,\vec{T}(\zVec) \big)   \mathrm{d}\zVec\\
                   &= \textstyle\HH[\,p_{\zetaVec(\PsiVec)}(\zVec)] \,+\, \zetaVecT_{(\PsiVec)}  \big( \AVec'(\zetaVec_{(\PsiVec)})\,-\,\EEE{\qbar_{\Phi}}{ \vec{T}(\zVec) } \big)\,,\label{EqnQBar} \\
%                   
   %\ \ \mbox{where}\ \  \qbar_{\Phi}(\zVec) :=   \frac{1}{N}\sum_{n} \qn_{\Phi}(\zVec)  \,,\label{EqnQBar}\\
%
                   &= \textstyle\HH[\,p_{\zetaVec(\PsiVec)}(\zVec)] \,+\, \zetaVecT_{(\PsiVec)} \ \vec{f}(\Phi,\PsiVec)\,, \label{EqnFFTwoShort}%\\[1ex]
%
                   %&\ \ \mbox{where}\ \  \vec{f}(\Phi,\PsiVec) :=  \AVec'(\zetaVec(\PsiVec))\,-\,\EEE{\qbar_{\Phi}}{ \vec{T}(\zVec) }\,. \label{EqnFunctionF}
%                   
%                   \ \ \mbox{and}\ \ \ \qbar_{\Phi}(\zVec) :=   \frac{1}{N}\sum_{n} \qn_{\Phi}(\zVec)  
%
\end{align}
where $\qbar_{\Phi}(\zVec) :=   \frac{1}{N}\sum_{n} \qn_{\Phi}(\zVec)$ and $\vec{f}(\Phi,\PsiVec) :=  \AVec'(\zetaVec(\PsiVec))\,-\,\EEE{\qbar_{\Phi}}{ \vec{T}(\zVec) }\,. \label{EqnFunctionF}$
Our aim will now be to show that the second term of (\ref{EqnFFTwoShort}) vanishes at stationary points of $\FF(\Phi,\PsiVec,\ThetaVec)$. For all stationary points of $\FF(\Phi,\PsiVec,\ThetaVec)$,
it holds that the derivatives w.r.t.\ all parameters vanish. We have assumed the parameters $\Phi$, $\PsiVec$ and $\ThetaVec$ to be separate sets of parameters. $\FF_2(\Phi,\PsiVec)$ is, therefore, the only summand of $\FF(\Phi,\PsiVec,\ThetaVec)$ which depends on $\PsiVec$. By using (\ref{EqnFFSecond}) and (\ref{EqnDelLog}) we obtain at stationary points:
% we consequently obtain:
%
\begin{align}
\vec{0} =\disT \del{\PsiVec} \FF_2(\Phi,\PsiVec)
                    &= \phantom{-}\, \textstyle\frac{1}{N}\sum_{n} \int \qn_{\Phi}\!(\zVec)\  \del{\PsiVec} \Big( - \log\!\big( p_{\zetaVec(\PsiVec)}(\zVec) \big) \Big)\, \hspace{0.5ex} \mathrm{d}\zVec\\
                    &= \phantom{-}\, \textstyle\frac{1}{N}\sum_{n} \int \qn_{\Phi}\!(\zVec)\ \ITnew \Big( \big( \AVec'(\zetaVec(\PsiVec))\,-\,\vec{T}(\zVec) \big) \Big) \hspace{0.5ex} \mathrm{d}\zVec\\
                    &= \phantom{-}\, \textstyle\ITnew\ \big( \AVec'(\zetaVec_{(\PsiVec)})\,-\,\EEE{\qbar_{\Phi}}{ \vec{T}(\zVec) } \big)\,,
%
%                    &= \phantom{-}\, \IIT_{(\PsiVec)}\ \vec{f}(\Phi,\PsiVec) \label{}
%
\end{align}
where $\qbar_{\Phi}$ is defined in (\ref{EqnQBar}).
% and 
%
Hence, at all stationary points (w.r.t.\,parameters $\PsiVec$) the following concise condition holds:
\begin{align}
 \ITnew\ \vec{f}(\Phi,\PsiVec) &= \vec{0} \label{EqnConditionPrior}\,,
\end{align}
where $\vec{f}(\Phi,\PsiVec)$ is the same function as introduced in~(\ref{EqnFFTwoShort}).
We can now apply Lemma~\ref{lem:ParamCrit} as we assumed our generative
model to fulfill the criterion given in Definition~\ref{def:Param_Crit}. 
Part~A of the lemma is formulated for arbitrary functions $\vec{f}$ with arbitrary parameters $\Phi$ and arbitrary values for $\PsiVec$.
So if the parameterization criterion is fulfilled, then (\ref{EqnLemmaParamCritA}) of Lemma~\ref{lem:ParamCrit} also applies to the specific function 
$\vec{f}$ as defined in (\ref{EqnFFTwoShort}), where $\Phi$ are the variational parameters.
Using Lemma~\ref{lem:ParamCrit} we can, therefore, conclude from (\ref{EqnConditionPrior}) that at stationary points applies:
%
%then we can (using Lemma~\ref{kkk}) conclude from  (\ref{EqnLemmaParamCritA}) t
% (\ref{EqnLemmaParamCritA}) also applies for the $\vec{f}$ as defined in (\ref{EqnFunctionF}) with $\Phi$ being the variational parameters.
%
%was formulated without the function $\vec{f}$
%depending on $\Phi$. But the variational parameters $\Phi$ are treated as arbitrary constant values throughout the
%proof and can therefore be treated as part of a fixed function (with arguments $\PsiVec$).
%
%Using Lemma~\ref{lem:ParamCrit} we thus conclude from (\ref{EqnConditionPrior}) that at stationary points applies:
%
%Throughout the proof, we variational 
%
%If for the considered generative model the parameterization criterion of Definition~C holds, we can conclude from (\ref{EqnConditionPrior}) that
%at all stationary points (w.r.t.\,parameters $\PsiVec$) holds:
%
\begin{align}
\zetaVecT_{(\PsiVec)} \ \vec{f}(\Phi,\PsiVec) &= 0\,.
\end{align}
Using Eqn.~(\ref{EqnFFTwoShort}) we consequently obtain at all stationary points for the summand $\FF_2(\Phi,\PsiVec)$:
\begin{align}
\FF_2(\Phi,\PsiVec) &= \HH[\,p_{\zetaVec(\PsiVec)}(\zVec)] \,+\, \zetaVecT_{(\PsiVec)} \ \vec{f}(\Phi,\PsiVec) = \HH[\,p_{\zetaVec(\PsiVec)}(\zVec)] = \HH[\,p_{\PsiVec}(\zVec)]\,.
\label{EqnResultFFTwo}
\end{align}
%
%\ \\
%\newpage
\noindent\underline{$\FF_3(\Phi,\ThetaVec)$ at stationary points}\\[1ex]
%
% so we start by considering the second summand (Eqn.\,\ref{EqnFFTwo}). 
\noindent{}The proof part for $\FF_3(\Phi,\ThetaVec)$ will be analog to that of $\FF_2(\Phi,\PsiVec)$ above but slightly more intricate because the noise distribution is conditional on $\zVec$. We have assumed that the generative model is an EF generative model. Therefore we can rewrite $\FF_3(\Phi,\ThetaVec)$ of Eqn.\,\ref{EqnFFThree} using the reparameterization of $p_{\ThetaVec}(\xVec\,|\,\zVec)$ in terms of the exponential family distribution $p_{\etaVec(\zVec;\,\ThetaVec)}(\xVec)$:
\begin{align}
\FF_3(\Phi,\ThetaVec)\ &= -\, \textstyle\frac{1}{N}\sum_{n} \int \qn_{\Phi}\!(\zVec) \log\!\big( p_{\ThetaVec}(\xVecN\,|\,\zVec) \big)\hspace{0.5ex} \mathrm{d}\zVec \label{EqnFFThird}\\
                    &= \phantom{-}\, \textstyle\frac{1}{N}\sum_{n} \int \qn_{\Phi}\!(\zVec) \Big( - \log\!\big( p_{\etaVec(\zVec;\,\ThetaVec)}(\xVecN) \big) \Big) \hspace{0.5ex} \mathrm{d}\zVec\,.\nonumber
\end{align}
As $p_{\etaVec} (\xVec)$ is an exponential family distribution, its negative logarithm is (see Eqn.\,\ref{EqnExpFamB}) given by:
\begin{align}
%
%-\log\big( p_{\etaVec} (\xVec) \big) &=A\big(\etaVec\,\big) \,-\, \etaVecT \vec{T}(\xVec) \,-\, \log\big( h(\xVec) \big)\,. \label{EqnLogJointThree}
%
-\log\big( p_{\etaVec(\zVec;\,\ThetaVec)} (\xVec) \big) &=A\big(\etaVec(\zVec;\,\ThetaVec)\,\big) \,-\, \etaVecT_{(\zVec;\,\ThetaVec)} \vec{T}(\xVec) \,-\, \log\big( h(\xVec) \big)\,. \label{EqnLogJointThree}
\end{align}
We have noted before that we will use the same symbols for sufficient statistics $\vec{T}(\xVec)$, partition function $A(\etaVec)$ and base measure $h(\xVec)$ as for the prior in Eqn.\,\ref{EqnLogJoint}. 
%As prior and noise model distribution are (in general) different members of the exponential family, these entities (in general) differ, of course. But as they can be distinguished from context, we avoid the introduction of further symbols.  

We will later require the derivative of the (negative) log probability w.r.t.\ $\thetaVec$, i.e., w.r.t.\ a subset of the parameters $\ThetaVec$.
The derivative is given by:
\begin{align}
\dell{\thetaVec} \Big( -\log\big( p_{\etaVec(\zVec;\,\ThetaVec)} (\xVec) \big) \Big) &= \JTnew  \big( \AVec'(\etaVec(\zVec; \ThetaVec))\,-\,\vec{T}(\xVec) \big)\,, \label{EqnDelLogThree}\\
                   &\ \ \ \mbox{where}\ \ \AVec'(\etaVec(\zVec; \ThetaVec)\,):=\dell{\etaVec}\,A(\etaVec\,)\Big|_{\etaVec=\etaVec(\zVec; \ThetaVec)}\label{EqnAPrimeThree}
%
%A\big(\zetaVec(\PsiVec)\,\big) \,-\, \zetaVecT_{(\PsiVec)} \vec{T}(\zVec)  \,-\, \log\big( h(\zVec) \big)\,. \label{EqnLogJoint}
%
\end{align}
and where $\JTnew$ is the transposed Jacobian matrix defined in (\ref{lem:JTmatrix}).

We will show that the summand $\FF_3(\Phi,\ThetaVec)$ will at stationary points be equal to the (expected) noise model entropy. Therefore, we first rewrite $-\log\!\big( p_{\etaVec(\zVec; \ThetaVec)}(\xVec) \big)$ using the noise model entropy. By again using the definition of the entropy
\begin{align}
\HH[\,p_{\etaVec(\zVec; \ThetaVec)}(\xVec)] &= \int p_{\etaVec(\zVec; \ThetaVec)}(\xVec) \Big( -\log\big( p_{\etaVec(\zVec; \ThetaVec)} (\xVec) \big) \Big)\mathrm{d}\xVec
\end{align}
and (\ref{EqnLogJointThree}) we obtain:
\begin{align}
\HH[\,\pTheta(\xVec\,|\,\zVec)] &= \HH[\,p_{\etaVec(\zVec; \ThetaVec)}(\xVec)] \label{EqnEntropyExpressionThree}
\\&= A\big( \etaVec(\zVec; \ThetaVec)  \big) \,-\, \etaVecT_{(\zVec;\,\ThetaVec)} \AVec'\big(\etaVec(\zVec; \ThetaVec)\big) \,-\, B(\etaVec(\zVec; \ThetaVec))\, \nonumber\\[1ex]
\Rightarrow\ A\big( \etaVec(\zVec; \ThetaVec)  \big)  &= \HH[\,p_{\etaVec(\zVec; \ThetaVec)}(\xVec)]  \,+\, \etaVecT_{(\zVec;\,\ThetaVec)} \AVec'\big(\etaVec(\zVec; \ThetaVec)\big) \,+\, B(\etaVec(\zVec; \ThetaVec))\,, \label{EqnAasHThree}\\[1ex]
                    &\hspace{-10mm}\mbox{where we abbreviated}\ \ B(\etaVec\,):= \disT\int p_{\etaVec}(\xVec) \,\log\big(h(\xVec)\big)\, \mathrm{d}\xVec \label{EqnBPrimeThree}\,. %\,=\, h_o\,. 
\end{align}
%
%The last equality for $B(\zetaVec)$ follows if we use our assumption of a constant base measure $h(\zVec)=h_o$. 
%
%We now use (\ref{EqnEntropyExpression}) to replace the partition function $A\big(\zetaVec\,\big)$ by an expression
%containing the entropy:
 %
%\begin{align}
%
%
%A\big( \etaVec(\zVec; \ThetaVec)  \big)  &= \HH[\,p_{\etaVec(\zVec; \ThetaVec)}(\zVec)]  \,+\, \etaVecT_{(\zVec;\,\ThetaVec)} \AVec'\big(\etaVec(\zVec; \ThetaVec)\big) \,+\, B(\etaVec(\zVec; \ThetaVec)) \label{EqnAasH}
%\end{align}
%
By combining (\ref{EqnLogJointThree}) and (\ref{EqnAasHThree}), the negative logarithm of the noise distribution becomes:
\begin{align}
- \log\!\big( p_{\etaVec(\zVec; \ThetaVec)}(\xVec)\big)&= \HH[\,p_{\etaVec(\zVec; \ThetaVec)}(\xVec)] \,+\, \etaVecT_{(\zVec;\,\ThetaVec)} \big( \AVec'(\etaVec{(\zVec; \ThetaVec)})\,-\,\vec{T}(\xVec) \big)\\
&\hspace{20mm}\,+\, B(\etaVec(\zVec; \ThetaVec))\,-\,\log\big(h(\xVec)\big)\nonumber\\
  &= \HH[\,p_{\etaVec(\zVec; \ThetaVec)}(\xVec)] \,+\, \etaVecT_{(\zVec;\,\ThetaVec)} \big( \AVec'(\etaVec{(\zVec; \ThetaVec)})\,-\,\vec{T}(\xVec) \big)\,, \label{EqnNegLogThree}
%
%\,+\, B(\etaVec(\zVec; \ThetaVec))\,-\,\log\big(h(\xVec)\big)\,, 
%
%\ \mbox{where}\ \AVec'(\zetaVec)\ \mbox{as in Eqn.\,\ref{EqnAPrime}}.
%
\end{align}
where $\AVec'\big(\etaVec(\zVec;\,\ThetaVec)\big)$ is defined as in Eqn.\,\ref{EqnAPrimeThree}, and where the last two terms cancel again because we assumed a constant base measure $h(\xVec)$ also for the noise model distribution.
%, the last two terms cancel.
Inserting the negative logarithm (\ref{EqnNegLogThree}) back into (\ref{EqnFFThird}), we therefore obtain for the third summand of the ELBO:
%
%The last two terms that depend on the base measure $h(\xVec)$ of the prior cancel because we assumed constant base measure. We therefore obtain for $\FF_2(\Psi,\Theta)$:
%
%
\begin{align}
& \FF_3(\Phi,\ThetaVec)\\
&= \frac{1}{N}\sum_{n} \int \qn_{\Phi}\!(\zVec) \Big( \HH[\,p_{\etaVec(\zVec; \ThetaVec)}(\xVec)] \,+\, \etaVecT_{(\zVec;\,\ThetaVec)} \big( \AVec'(\etaVec{(\zVec; \ThetaVec)})\,-\,\vec{T}(\xVecN) \big) \Big) \hspace{0.5ex} \mathrm{d}\zVec\\
                   &= \frac{1}{N} \sum_{n} \EEE{\qn_{\Phi}}{ \HH[\,p_{\etaVec(\zVec; \ThetaVec)}(\xVec)] } \,+\, \frac{1}{N}\sum_{n} \int \qn_{\Phi}\!(\zVec)\ \etaVecT_{(\zVec;\,\ThetaVec)}  \big( \AVec'(\etaVec{(\zVec; \ThetaVec)})\\
                   &\hspace{20mm}-\,\vec{T}(\xVecN) \big)  \mathrm{d}\zVec\nonumber\\
                   &= \frac{1}{N} \sum_{n} \EEE{\qn_{\Phi}}{ \HH[\,p_{\etaVec(\zVec; \ThetaVec)}(\xVec)] } \,+\, \frac{1}{N}\sum_{n}\int \qn_{\Phi}\!(\zVec)\ \etaVecT_{(\zVec;\,\ThetaVec)}\ \vec{g}_n(\zVec;\ThetaVec)\mathrm{d}\zVec, \label{EqnFFThreeShort}\\
                   &\ \ \mbox{where}\ \  \vec{g}_n(\zVec;\ThetaVec) := \AVec'\big(\etaVec(\zVec;\,\ThetaVec)\big)\,-\,\vec{T}(\xVecN)\,. \label{EqnFunctionG}
\end{align}
In analogy to the derivation for $\FF_2(\Phi,\PsiVec)$, our aim will now be to show that the second term of (\ref{EqnFFThreeShort}) vanishes at stationary points of $\FF(\Phi,\PsiVec,\ThetaVec)$. For all stationary points of $\FF(\Phi,\PsiVec,\ThetaVec)$, it holds that the derivatives w.r.t.\ all parameters vanish. 
We have assumed the parameters $\Phi$, $\PsiVec$ and $\ThetaVec$ to be separate sets of parameters. $\FF_3(\Phi,\ThetaVec)$ is therefore the only summand of $\FF(\Phi,\PsiVec,\ThetaVec)$ which depends on $\ThetaVec$.
At stationary points we can consequently conclude for any subset $\thetaVec$ of $\ThetaVec$ that $\del{\thetaVec}\, \FF_3(\Phi,\ThetaVec)=0$. 
By using (\ref{EqnFFThird}) and (\ref{EqnDelLogThree}) we obtain at any stationary point and for any subset $\thetaVec$:
% we consequently obtain:
%
\begin{align}
0 =\disT \del{\thetaVec}\, \FF_3(\Phi,\ThetaVec)\nonumber
                    &= \phantom{-}\, \frac{1}{N}\sum_{n} \int \qn_{\Phi}\!(\zVec)\  \del{\thetaVec} \Big( - \log\!\big( p_{\etaVec(\zVec; \ThetaVec)}(\xVec^{(n)}) \big) \Big)\, \hspace{0.5ex} \mathrm{d}\zVec\\
                    &= \phantom{-}\,\frac{1}{N}\sum_{n} \int \qn_{\Phi}\!(\zVec)\ \JTnew \Big( \AVec'(\etaVec(\zVec; \ThetaVec))\,-\,\vec{T}(\xVec^{(n)}) \Big) \hspace{0.5ex} \mathrm{d}\zVec. \label{EqnDelFFThree} %\\
%
%%s                    &= \phantom{-}\,\frac{1}{N}\sum_{n} \int  \qn_{\Phi}\!(\zVec)\ \JTnew\ \Big( \AVec'\big(\etaVec(\zVec;\,\ThetaVec)\big)\,-\,\vec{T}(\xVecN) \Big)\,\mathrm{d}\zVec. \label{EqnDelFFThree}
%
%                    &= \phantom{-}\, \IIT_{(\PsiVec)}\ \vec{f}(\Phi,\PsiVec) \label{}
%
\end{align}
%
%where $\qbar_{\Phi}$ is defined in (\ref{EqnQBarThree}).
% and 
%
In expression (\ref{EqnDelFFThree}) we recognize the functions $\vec{g}_n(\zVec; \ThetaVec)$ as defined in Eqn.\,\ref{EqnFunctionG}. 
Therefore, expression~(\ref{EqnDelFFThree}) means that at all stationary points and for all subsets $\thetaVec$ of $\ThetaVec$ the
following holds:
\begin{align}
  \frac{1}{N} \sum_{n}\int \qn_{\Phi}\!(\zVec)\ \JTnew\ \vec{g}_n(\zVec; \ThetaVec)\,\mathrm{d}\zVec &= 0 \label{EqnConditionNoise}\,.
\end{align}
%
%where $\vec{f}(\Phi,\PsiVec)$ is the same function as introduced in (\ref{EqnFunctionF}).
%
As (\ref{EqnConditionNoise}) applies for all subsets $\thetaVec$, it also applies for the specific subset $\thetaVec$ of $\ThetaVec$ 
that exists according to the parameterization criterion (Definition~\ref{def:Param_Crit}, part~B) or Lemma~\ref{lem:ParamCrit} (part~B). Using Part~B of Lemma~\ref{lem:ParamCrit}, we
can consequently conclude from (\ref{EqnConditionNoise}) that at all stationary points applies:
\begin{align}
\frac{1}{N}\sum_{n}\int \qn_{\Phi}\!(\zVec)\ \etaVec^{\mathrm{T}}_{(\zVec;\ThetaVec)}\ \gVec_n(\zVec;\ThetaVec) \,\mathrm{d}\zVec\ =\ 0\,. \label{EqnConditionNoiseInProof}
\end{align}
By going back to (\ref{EqnFFThreeShort}) and by inserting (\ref{EqnConditionNoiseInProof}), we consequently obtain at all stationary points for the summand $\FF_3(\Phi,\ThetaVec)$:
\begin{align}
 \FF_3(\Phi,\ThetaVec) &= \frac{1}{N} \sum_{n} \EEE{\qn_{\Phi}}{ \HH[\,p_{\etaVec(\zVec; \ThetaVec)}(\xVec)] } \,+\, \frac{1}{N}\sum_{n}\int \qn_{\Phi}\!(\zVec)\ \etaVecT_{(\zVec;\,\ThetaVec)}\ \vec{g}_n(\zVec; \ThetaVec)\ \mathrm{d}\zVec \nonumber\\
&= \frac{1}{N} \sum_{n} \EEE{\qn_{\Phi}}{ \HH[\,p_{\etaVec(\zVec; \ThetaVec)}(\xVec)] } = \frac{1}{N} \sum_{n} \EEE{\qn_{\Phi}}{ \HH[\,p_{\ThetaVec}(\xVec\,|\,\zVec)] }\,.\label{EqnResultFFThree}
\end{align}

\noindent\underline{Conclusion}\\[1ex]
%
%The ELBO (\ref{EqnFFMain}) 
%
We rewrote the ELBO to consist of three terms, $\FF(\Phi,\PsiVec,\ThetaVec) =  \FF_1(\Phi) - \FF_2(\Phi,\PsiVec) - \FF_3(\Phi,\ThetaVec)$,
%
%\begin{align}
%
%\FF(\Phi,\PsiVec,\ThetaVec) &=  \FF_1(\Phi) - \FF_2(\Phi,\PsiVec) - \FF_3(\Phi,\ThetaVec)
%
%\end{align}
%
with $\FF_1(\Phi)$, $\FF_2(\Phi,\PsiVec)$ and $\FF_3(\Phi,\ThetaVec)$ as defined by Eqns.~(\ref{EqnFFOne}), (\ref{EqnFFTwo}) and (\ref{EqnFFThree}), respectively.
The first term is by definition an average entropy, $\FF_1(\Phi)=\frac{1}{N}\sum_{n=1}^N \HH[\qPhiN(\zVec)]$. We have shown (see Eqn.\,\ref{EqnResultFFTwo}) that $\FF_2(\Phi,\PsiVec)$ becomes equal to the entropy of the prior distribution at stationary points. Furthermore, we have shown (see Eqn.\,\ref{EqnResultFFThree}) that $\FF_3(\Phi,\ThetaVec)$ becomes equal to an expected entropy at stationary points. By using the results (\ref{EqnResultFFTwo}) and (\ref{EqnResultFFThree}), we have shown that at all stationary points the following applies for the ELBO (\ref{EqnFFStandardKL}):
%
% is equal at all stationary points applies:
%
\begin{align}
\FF(\Phi,\PsiVec,\ThetaVec) &= \disT \frac{1}{N}\sum_{n=1}^N \HH[\qPhiN(\zVec)]   - \HH[\,\pPsi(\zVec)] - \frac{1}{N} \sum_{n=1}^N \EEE{\qn_{\Phi}}{ \HH[\,\pT(\xVec\,|\,\zVec)] },
\end{align}
from which (\ref{EqnTheoremSoE}) follows by using (\ref{EqnAgPost}) for the last summand.
%which proves the claim.\\[1ex]
\end{proof}
\noindent{}Theorem~\ref{th:Sum_of_Entr} can be regarded as the main result. Most generative models of the form given by Definition \ref{def:Gen_Model}
are usually defined using exponential family distributions, and usually those distributions have a constant base measure (Bernoulli, Categorical, Exponential, Gaussian, Gamma etc.). Still, it may be a bit unsatisfactory that not all exponential family distributions can be treated, which is why we consider a
generalization of Theorem~\ref{th:Sum_of_Entr} in the following.
%
%generative models (of the form given by Definition \ref{def:Gen_Model}) will satisfy the theorem's conditions (see Sec.\,\ref{SecExamples} for examples).
%The reason is that most distributions used to define such models are standard exponential family distributions;
%and most standard exponential family distributions (Bernoulli, Categorical, Exponential, Gaussian, Gamma etc.)
%have constant base measures.
%
%
%
%

\section{Convergence for General Exponential Families}\label{SecGenEF}
%
%Most generative models with one set of latents and one set of observables (Definition \ref{def:Gen_Model}) that
%are used in Machine Learning are EF generative models (Definition \ref{def:EF_Gen_Model}) because most distributions
%that are used are exponential family distributions of which most have constant base measure.
%
While Theorem~\ref{th:Sum_of_Entr} applies for a broad range of generative models, the condition on exponential family distributions
to have constant base measures (Definition~\ref{def:EF_Gen_Model}) excludes practically relevant distributions
such as the Poisson distribution. Also from a theoretical perspective, it may not feel satisfactory to only consider a subset of exponential family distributions.
Therefore, in this section, we will treat general exponential family distributions. The treatment will, in principle, be analogous to the previous
section. However, we will in this section explicitly use Lebesgue integrals and general (non-negative) measures, which we have avoided
for notational simplicity, so far. 

%We will begin with the log-likelihood $\LL(\PsiVec,\ThetaVec)$ of a generative model as given by Definition~\ref{def:Gen_Model}.
%By using a general measure $\mu(\zVec)$ for the latent probability space ... explicitly, we obtain:
%
%\begin{align}
%    \LL(\PsiVec,\ThetaVec) = \frac{1}{N}\sum_n \log \Big(\int p_{\ThetaVec}(\xVec^{(n)}\,|\,\zVec)\, p_{\PsiVec}(\zVec)\, \mathrm{d}\mu(\zVec)\Big).
%\end{align}
%
%
As distributions $p_{\PsiVec}(\zVec)$ and $p_{\ThetaVec}(\xVec\,|\,\zVec)$ we now demand exponential family distributions but without restrictions to constant base measure.

\begin{definition}[General EF Models]
    \label{def:GenEF}
Given a generative model as given by Definition \ref{def:Gen_Model}, we say the generative model is a {\em general} EF model if
there exist exponential family distributions $\pZetaVec(\zVec)$ (for the latents) and $\pEtaVec(\xVec)$ (for the observables)
such that the generative model can be reparameterized to take the following form:\vspace{0ex}
\begin{eqnarray}
%
% \zVec &\sim& p_{\zetaVec(\Psi,\Theta)}(\zVec) \\ 
 \zVec &\sim& p_{\zetaVec(\PsiVec)}(\zVec), \\ 
 \xVec &\sim& p_{\etaVec(\zVec;\,\ThetaVec)}(\xVec),
\end{eqnarray}
where $\zetaVec(\PsiVec)$ maps the prior parameters $\PsiVec$ to the natural parameters of distribution $\pZetaVec(\zVec)$, and where $\etaVec(\zVec;\ThetaVec)$ maps $\zVec$ and the noise model parameters $\ThetaVec$ to the natural parameters of distribution~$\pEtaVec(\xVec)$.
\end{definition}
%\ \\[0ex]
%\noindent{}Definition \ref{def:GenEF} is the same as Definition \ref{def:EF_Gen_Model} but without the restriction to constant base measures.
%
\noindent{}The main idea for the generalization of Theorem~\ref{th:Sum_of_Entr} is using the base measures of the distributions in Definition~\ref{def:GenEF} to define new measures for Lebesgue integration. The derivations for the general case can then be done in analogy to the proof of Theorem~\ref{th:Sum_of_Entr} except of using other Lebesgue integrals. Before, however, we require some preparations. Concretely, we introduce the new measures and define entropies and log-likelihood in the spaces with the new measures.
%
%Let us start with defining entropies in analogy to standard entropies but using the new measures.

%As a further preparation, we will need a modified definition of entropy motivated by non-constant base measures.
\begin{definition}[New Measures and Densities]
    \label{def:Measures}
%
%\mycomment{umschreiben:} 
We consider a probability space with measure $\mu(\zVec)$. In this space let $p_{\zetaVec}(\zVec)$ be a probability density of an exponential family with base measure $h(\zVec)$. The density $p_{\zetaVec}(\zVec)$ is normalized w.r.t.\ the measure $\mu(\zVec)$ (e.g.\,, but not necessarily, the standard Lebesgue measure of $\RRR^H$). We now define a new measure and the corresponding probability density as follows:
\begin{eqnarray}
\forall{}\zVec:\phantom{ii}\tilde{\mu}(\zVec)\,=\,h(\zVec)\hspace{0.6mm}\mu(\zVec)\phantom{iii} \mbox{and}\phantom{iii} \forall{}\zVec:\phantom{ii}\pt_{\zetaVec}(\zVec)\,=\,\frac{1}{h(\zVec)}\,\,p_{\zetaVec}(\zVec)\,.\label{EqnExpFemAGenMeasure}
\end{eqnarray}
Analogously,
%\mycomment{umschreiben:}
consider a probability space with measure $\mu(\xVec)$. In this space let $p_{\etaVec}(\xVec)$ be a probability density
of an exponential family with base measure $h(\xVec)$. The density $p_{\etaVec}(\zVec)$ is normalized w.r.t.\ the measure $\mu(\xVec)$
(this can, for instance, be the standard Lebesgue measure of $\RRR^D$). We now define: %a new measure and the corresponding probability density as follows:
\begin{eqnarray}
\forall{}\xVec:\phantom{ii}\tilde{\mu}(\xVec)\,=\,h(\xVec)\,\mu(\xVec)\phantom{iii} \mbox{and}\phantom{iii} \forall{}\xVec:\phantom{ii}\pt_{\etaVec}(\xVec)\,=\,\frac{1}{h(\xVec)}\,p_{\etaVec}(\xVec)\,.\label{EqnExpFemBGenMeasure}
\end{eqnarray}
\end{definition}
%\ \\[0ex]
%
%
\noindent{}It immediately follows that if $p_{\zetaVec}(\zVec)$ is given by (\ref{EqnExpFamA}) and if $p_{\etaVec}(\xVec)$ is given by (\ref{EqnExpFamB}) then we obtain:
\begin{align}
\tilde{p}_{\zetaVec} (\zVec) \,=\, \exp\Big(  \zetaVecT \vec{T}(\zVec) \,-\, A\big(\zetaVec \big) \Big)\phantom{iii} \mbox{and}\phantom{iii}  %\label{EqnExpFamAGen}\\
\tilde{p}_{\etaVec} (\xVec) \,=\,  \exp\Big(   \etaVecT \vec{T}(\xVec) \,-\, A\big(\etaVec\big) \Big)\,, \label{EqnExpFamGen}
\end{align}
i.e., $\tilde{p}_{\zetaVec} (\zVec)$ and $\tilde{p}_{\etaVec} (\xVec)$ have unit base measures in the new probability spaces.
So, $\tilde{p}_{\zetaVec} (\zVec)$ and $\tilde{p}_{\etaVec} (\xVec)$ simply have the form of the original distributions but
with base measures equal one (in that sense Eqns.\ (\ref{EqnExpFamGen}) are formally more rigorous than (\ref{EqnExpFemAGenMeasure}) and (\ref{EqnExpFemBGenMeasure}) because the latter use divisions by $h(\zVec)$ and $h(\xVec)$, respectively, and the base measures can be zero at some points). It is also immediately obvious that $\tilde{p}_{\zetaVec} (\zVec)$ and $\tilde{p}_{\etaVec} (\xVec)$ are normalized to one in the new spaces. For completeness, we also define the probabilities w.r.t.\ the new measures but using the original parameters $\PsiVec$ and $\ThetaVec$: 
\begin{align}
\tilde{p}_{\PsiVec} (\zVec) \,=\, \tilde{p}_{\zetaVec(\PsiVec)} (\zVec) \phantom{iii} \mbox{and}\phantom{iii} 
\tilde{p}_{\ThetaVec} (\xVec\,|\,\zVec) \,=\,  \tilde{p}_{\etaVec(\zVec;\,\ThetaVec)} (\xVec)\,, \label{EqnExpFamOrgGen}
\end{align}
where the functions $\zetaVec(\zVec)$ and $\etaVec(\zVec;\,\ThetaVec)$ remain unchanged, i.e., the change of measures has no effect on the mappings to the natural parameters.
%
%
%p_{\zetaVec(\PsiVec)} (\zVec)  &= h(\zVec)\,  \exp\Big(  \zetaVecT_{(\PsiVec)} \vec{T}(\zVec) \,-\, A\big(\zetaVec(\PsiVec) \big) \Big)\,, \label{EqnExpFamA}\\
%
%
%p_{\etaVec(\zVec;\,\ThetaVec)} (\xVec)  &=  h(\xVec)\,  \exp\Big(   \etaVecT_{(\zVec;\,\ThetaVec)} \vec{T}(\xVec) \,-\, A\big(\etaVec(\zVec;\,\ThetaVec)\big) \Big)\,, 
%which is the actual meaning of Eqn.\,
%Eqn.\,is a bit informal and simply means that 

%Division by $h(\zVec)$ or $h(\xVec)$ in ... is a bit informal as the base measures can be zero. 

%As a further preparation, we will require entropies and likelihood defined in the new probability spaces. Let us start with entropy definitions
%that we define in analogy to standard entropies.
%
As a further preparation, we will require entropies and likelihood defined in the new probability spaces. Let us start with entropy definitions. For an exponential family distribution $p_{\PsiVec}(\zVec)=p_{\zetaVec(\PsiVec)}(\zVec)$ as given in Definition~\ref{def:GenEF}, we define the \textit{pseudo entropy} of $p_{\PsiVec}(\zVec)$ in the same way as the
conventional entropy but using the new measure:
%
%\begin{definition}[Pseudo Entropies]
%    \label{def:PseudoH}
%
%Consider a general EF model of Definition~\ref{def:GenEF} with probability densities $p_{\zeta}(\zVec)$ and $p_{\etaVec}(\xVec)$ defined w.r.t.\ the measures $\mu(\zVec)$ and $\mu(\xVec)$, respectively. Now consider the new measures $\tilde{\mu}(\zVec)$ and $\tilde{\mu}(\xVec)$ introduced by Defintion~\ref{def:Measures}.
%
%If $p(\zVec)$ is a probability density defined w.r.t.\ the original measure $\mu(\zVec)$, then 
%$\tilde{p}(\zVec)=\frac{1}{h(\zVec)}\,p(\zVec)$ is a probability density w.r.t.\ the new measure $\tilde{\mu}(\zVec)$.
%We can then define the {\em pseudo entropy} of $p(\zVec)$ as follows:
%
\begin{align}
 \HHt[\,p_{\PsiVec}(\zVec)] &= -\int \pt_{\PsiVec}(\zVec) \log \pt_{\PsiVec}(\zVec)\, \mathrm{d}\tilde{\mu}(\zVec)= -\zetaVecT_{(\PsiVec)} \AVec^\prime(\zetaVec(\PsiVec))+A(\zetaVec(\PsiVec)) \,,
\label{EqnPseudoHOne}
\end{align}
where $\AVec'(\zetaVec(\PsiVec)\,):=\dell{\zetaVec}\,A(\zetaVec\,)\Big|_{\zetaVec=\zetaVec(\PsiVec)}$ is the gradient of the log-partition function with respect to the natural parameters. The pseudo entropy can be expressed concisely using natural parameters and log-partition function (see right-hand-side of Eqn.~\ref{EqnPseudoHOne} which is derived in Appendix~\ref{Suppl:DefPseudoEntropies}).
%
%That is, the pseudo entropy is defined as the conventional entropy of $\tilde{p}_{\PsiVec}(\zVec)=\tilde{p}_{\zetaVec(\PsiVec)}(\zVec)$ according to the new measure $\tilde{\mu}(\zVec)$. Since $\tilde{p}_{\PsiVec}(\zVec)$ is an exponential family distribution with unit base measure, the pseudo entropy takes on a particularly concise and convenient form (see the appendix for a detailed derivation).
%
Analogously, we define the \textit{pseudo entropy} of the observable distribution $\pt_{\ThetaVec}(\xVec\mid\zVec)=p_{\etaVec(\zVec;\,\ThetaVec)}(\xVec)$ as follows:
%Analogously, if $p(\xVec)$ is a probability density defined w.r.t.\ the original measure $\mu(\xVec)$, then 
%$\tilde{p}(\xVec)=\frac{1}{h(\xVec)}\,p(\xVec)$ is a probability density w.r.t.\ the new measure $\tilde{\mu}(\xVec)$.
%We can then define the {\em pseudo entropy} of $p(\xVec)$ as follows:
%
\begin{align}
 \HHt[\,p_{\ThetaVec}(\xVec\,|\,\zVec)] = -\int \pt_{\ThetaVec}(\xVec\,|\,\zVec) \log \pt_{\ThetaVec}(\xVec\,|\,\zVec)\, \mathrm{d}\tilde{\mu}(\xVec)= -\etaVecT_{(\zVec;\ThetaVec)} \AVec^\prime(\etaVec(\zVec;\ThetaVec))+A(\etaVec(\zVec;\ThetaVec)) \,,
\label{EqnPseudoHTwo}
\end{align}
where $\AVec'(\etaVec(\zVec; \ThetaVec)\,):=\dell{\etaVec}\,A(\etaVec\,)\Big|_{\etaVec=\etaVec(\zVec; \ThetaVec)}$ is, similarly, the gradient of the log-partition function with respect to the natural parameters (see again Appendix~\ref{Suppl:DefPseudoEntropies} for the derivation of the right-hand-side of Eqn.~\ref{EqnPseudoHTwo}).
%
%\end{definition}
%\ \\[0ex]
Although we consider pseudo entropies here only for exponential families, they can be defined for any (well-behaved) distributions (see Appendix~\ref{Suppl:DefPseudoEntropies} for details).
Specifically, the \textit{pseudo entropy} of the variational distribution $\qn_\Phi(\zVec)$ is defined as
\begin{align}
  \tilde{\HH}[\qn_\Phi(\zVec)] = - \int \tilde{q}^{(n)}_{\Phi}\!(\zVec) \log \tilde{q}^{(n)}_{\Phi}\!(\zVec)\, \mathrm{d}\tilde{\mu}(\zVec) = - \int q^{(n)}_{\Phi}\!(\zVec) \log \tilde{q}^{(n)}_{\Phi}\!(\zVec)\, \mathrm{d}\mu(\zVec),
\end{align}
where $\qn_\Phi(\zVec)$ and $\tilde{q}^{(n)}_{\Phi}\!(\zVec)$ are related by the equation $q^{(n)}_{\Phi}\!(\zVec) = h(\zVec)\,\tilde{q}^{(n)}_{\Phi}\!(\zVec)$, with $h(\zVec)$ representing the base measure of the prior distribution.
\ \\
\noindent{} We now define the log-likelihood w.r.t.\ the new measure $\tilde{\mu}(\zVec)$. In analogy to the standard log-likelihood $\LL(\PsiVec,\ThetaVec)$, we define the {\em pseudo log-likelihood} function $\LLt(\PsiVec,\ThetaVec)$ to be given by:
%
%
%\begin{defi}
%    \label{def:PseudoLL}
%    (Pseudo Log-Likelihood)
%\end{defi}
%
%\defStart{}Consider a general EF model as given by Definition \ref{def:GenEF}, where $h(\xVec)$ is the base measure of the observable distribution.
%Furthermore, given a set of $N$ data points $\xVecN$ then let $\LL(\PsiVec,\ThetaVec)$ be the standard log-likelihood function (\ref{...}).
%We then define the {\em pseudo log-likelihood} $\LLt(\PsiVec,\ThetaVec)$ to be 
%
%\begin{eqnarray}
%
% \zVec &\& p_{\zetaVec(\Psi,\Theta)}(\zVec) \\ 
% \LLt(\PsiVec,\ThetaVec) &=& \LL(\PsiVec,\ThetaVec) \,-\, \frac{1}{N}\sum_n \log\big( h(\xVecN) \big)\,.\\ 
%
%\end{eqnarray}
%
%\noindent{}$\square$\\[0ex]
%\ \\[0ex]
%\noindent{}
%
%
%%\begin{definition}[Pseudo Log-Likelihood]
%%    \label{def:PseudoLL}
%%Consider a general EF model as given by Definition \ref{def:GenEF}. Let the measures $\tilde{\mu}(\zVec)$ and $\tilde{\mu}(\xVec)$ as well as the densities $\pt_{\zetaVec}(\zVec)$ and $\pt_{\etaVec}(\xVec)$ be defined as in Definition~\ref{def:Measures}. In analogy to the standard log-likelihood $\LL(\PsiVec,\ThetaVec)$, we define the {\em pseudo log-likelihood} function $\LLt(\PsiVec,\ThetaVec)$ to be given by:
%Furthermore, given a set of $N$ data points $\xVecN$ then let $\LL(\PsiVec,\ThetaVec)$ be the standard log-likelihood function (\ref{...}).
%We then define the {\em pseudo log-likelihood} $\LLt(\PsiVec,\ThetaVec)$ to be 
%
\begin{align}
\LLt(\PsiVec,\ThetaVec) &= \frac{1}{N} \sum_n \log\Big( \int \pt_{\ThetaVec} (\xVec^{(n)}\,|\,\zVec)\, \pt_{\PsiVec} (\zVec)\,\mathrm{d}\Tilde{\mu}(\zVec) \Big)\\
 &= \frac{1}{N} \sum_n \log\Big( \int \pt_{\etaVec(\zVec; \ThetaVec)} (\xVec^{(n)})\, \pt_{\zetaVec(\PsiVec)} (\zVec)\,\mathrm{d}\Tilde{\mu}(\zVec) \Big).
\end{align}
%
%%\end{definition}
%
It is straightforward to relate the pseudo log-likelihood $\LLt(\PsiVec,\ThetaVec)$ to the standard log-likelihood. They differ only by an additive constant, which does not depend on any parameter (see Appendix~\ref{Suppl:DefPseudoEntropies} for details). Therefore, parameter optimization using the pseudo log-likelihood is equivalent to parameter optimization using the standard log-likelihood.

We can treat the pseudo log-likelihood analogously to our treatment of the conventional log-likelihood. The main technical difference will be the use of probability measures $\tilde{\mu}(\zVec)$ and $\tilde{\mu}(\xVec)$. To start, observe that it is straightforward to derive a lower bound (ELBO) of the pseudo log-likelihood. By applying the probabilistic version of Jensen's inequality (e.g.\,\cite{Athreya2006}) we obtain (see Appendix~\ref{Suppl:DefPseudoEntropies} for details):
\begin{align}
\ \label{EqnFFGen} \\[-5ex]
% \zVec &\& p_{\zetaVec(\Psi,\Theta)}(\zVec) \\ 
&\LLt(\PsiVec,\ThetaVec)= \frac{1}{N} \sum_n \log\Big( \int \pt_{\ThetaVec} (\xVec^{(n)}\,|\,\zVec)\, \pt_{\PsiVec} (\zVec)\,\mathrm{d}\Tilde{\mu}(\zVec) \Big)\nonumber\\
%
%%=&\disS \frac{1}{N}\sum_n \log \Big(\mathbb{E}_{\tilde{q}^{(n)}_\Phi}\Big\{\frac{\pt_{\ThetaVec}(\xVec^{(n)}\,|\,\zVec)\, \pt_{\PsiVec}(\zVec)}{\tilde{q}^{(n)}_\Phi(\zVec)}\Big\}\Big)
%
%%\hspace{0ex} \,\geq\, \hspace{0ex}\frac{1}{N}\sum_n \mathbb{E}_{\tilde{q}^{(n)}_\Phi}\Big\{\log \Big(\frac{\pt_{\ThetaVec}(\xVec^{(n)}\,|\,\zVec)\, \pt_{\PsiVec}(\zVec)}{\tilde{q}^{(n)}_\Phi(\zVec)}\Big)\Big\}\nonumber\\
%
%% =&\frac{1}{N}\sum_n \int \tilde{q}^{(n)}_\Phi(\zVec) \log\Big(\frac{\pt_{\ThetaVec}(\xVec^{(n)}\,|\,\zVec)\, \pt_{\PsiVec}(\zVec)}{\tilde{q}^{(n)}_\Phi(\zVec)}\Big) \mathrm{d}\tilde{\mu}(\zVec)\nonumber\\
%
&\ge  \frac{1}{N}\sum_n \int \tilde{q}^{(n)}_\Phi(\zVec) \log\big( \pt_{\ThetaVec}(\xVec^{(n)}\,|\,\zVec)\, \pt_{\PsiVec}(\zVec)\big)\, \mathrm{d}\tilde{\mu}(\zVec) 
 \,-\, \frac{1}{N}\sum_n \int \tilde{q}^{(n)}_\Phi(\zVec) \log \tilde{q}^{(n)}_\Phi(\zVec)\, \mathrm{d}\tilde{\mu}(\zVec)\nonumber\\
 &= \FFt(\Phi,\PsiVec,\ThetaVec),\nonumber
% \label{EqnFFGen} 
%
\end{align}
where the variational distributions $\tilde{q}^{(n)}_\Phi(\zVec)$ are normalized w.r.t.\ the new measure $\tilde{\mu}(\zVec)$. Similar to the log-likelihood and pseudo log-likelihood, the lower bound $\FFt(\Phi,\PsiVec,\ThetaVec)$ differs from the conventional ELBO only by an additive constant that does not depend on any model parameters (see Appendix~\ref{app:PseudoEntropies} for details).\\%=h(\zVec)\mu(\zVec)$.\\[2ex]

\noindent{}In a preparation of the main result, we proof a similar statement as in Lemma \ref{lem:ParamCrit}. Since the parameterization criterion given in Definition \ref{def:Param_Crit} does not refer to the base measures $h(\zVec)$ and $h(\xVec)$, it does not change for general EF models, and hence we assume the same criterion as in Sec.~\ref{SecEntropySums}.\\[0ex]
\begin{lemma}
  \label{lem:ParamCrit_general}
Consider a general EF model as given in Definition~\ref{def:GenEF} and let the dimensionalities of the natural parameter vectors $\zetaVec$ and $\etaVec$ be $K$ and $L$, respectively. Let further $\delll{\zetaVecT(\PsiVec)}{\PsiVec}$ and $\delll{\etaVecT(z;\ThetaVec)}{\thetaVec}$ denote the transposed Jacobians in (\ref{lem:ITmatrix}) and (\ref{lem:JTmatrix}), respectively.

If the general EF model fulfills the parameterization criterion (Definition~\ref{def:Param_Crit}), then the following two statements apply:
%
%then the following two statements are true:
%
%Then, under the assumption that the parameterization criterion (Definition~\ref{def:Param_Crit}) is fulfilled, the following two statements apply:
%
\begin{itemize}
  \item[(A)] For any vectorial function $\fVec(\Phi,\PsiVec)$ from the sets of parameters $\Phi$ and $\PsiVec$ to the ($K$-dim) space of natural parameters of the prior it holds:
    \begin{eqnarray}
      \label{EqnLemmaParamCritA_general}
      \delll{\zetaVecT(\PsiVec)}{\PsiVec}\fVec(\Phi,\PsiVec)\ =\ \vec{0}\ &\Rightarrow&\ \zetaVecT_{(\PsiVec)}\fVec(\Phi,\PsiVec)\ =\ 0.
    \end{eqnarray}
  \item[(B)]For any variational parameter $\Phi$ there is a non-empty subset $\thetaVec$ of $\ThetaVec$ such that the following applies for all vectorial functions $\gVec_1(\zVec;\ThetaVec),\ldots,\gVec_N(\zVec;\ThetaVec)$ from the parameter set $\ThetaVec$ and from the latent space $\Omega_{\zVec}$ to the ($L$-dim) space of natural parameters of the noise model:
  \begin{eqnarray}
 \phantom{i\hspace{-6ex}i}   \sum_{n=1}^{N} \int \tilde{q}^{(n)}_{\Phi}\!(\zVec)\delll{\etaVecT(z;\ThetaVec)}{\thetaVec}\gVec_n(\zVec;\ThetaVec)\,\mathrm{d}\tilde{\mu}(\zVec) &=& \vec{0}
    \label{EqnLemmaParamCritB1_general}\\
    \Rightarrow\  \sum_{n=1}^{N}\int \tilde{q}^{(n)}_{\Phi}\!(\zVec) \etaVecT_{(\zVec;\ThetaVec)}\gVec_n(\zVec;\ThetaVec)\,\mathrm{d}\tilde{\mu}(\zVec) &=& 0.%\phantom{ii}%\label{EqnLemmaParamCritB1_general}
  \end{eqnarray}
\end{itemize}
%
%
%%For an EF generative model as in definition B, the following two statements are equivalent:
%
%%\begin{itemize}
%%  \item[(A)] There is a vectorial function $\vVec(\PsiVec)$ form the set of parameters $\PsiVec$ to the ($R$-dim) space of $\PsiVec$ that satisfies the equation
%%  \begin{eqnarray}
%%    \zetaVec(\PsiVec) = \delll{\zetaVec(\PsiVec)}{\PsiVecT}\vVec(\PsiVec).
%%  \end{eqnarray}
%%  \item[(B)] For any vectorial function $\fVec$ from the set of parameters $\PsiVec$ to the ($K$-dim) space of natural parameters of the prior it holds:
%%  \begin{eqnarray}
%%    \delll{\zetaVecT(\PsiVec)}{\PsiVec}\fVec(\PsiVec)\ =\ \vec{0}&\Rightarrow& \zetaVecT(\PsiVec)\fVec(\PsiVec)\ =\ 0.
%%  \end{eqnarray}
%%\end{itemize}
\end{lemma}

\begin{proof}
The proof is essentially the same as that of Lemma \ref{lem:ParamCrit}. In fact, nothing changes for part A. For part B, we only need the linearity of the integral, which also applies in the case of a general measure. So we can infer from Eqn.~\ref{EqnLemmaParamCritB1_general} that
\begin{align}
  \sum_{n=1}^{N}\int \tilde{q}^{(n)}_{\Phi}\!(\zVec) \etaVecT_{(\zVec;\ThetaVec)}\gVec_n(\zVec;\ThetaVec)\,\mathrm{d}\tilde{\mu}(\zVec) =  \betaVec^{\mathrm{T}}(\ThetaVec)\sum_{n=1}^{N} \int \tilde{q}^{(n)}_{\Phi}\!(\zVec)\delll{\etaVecT(\zVec;\ThetaVec)}{\thetaVec}\gVec_n(\zVec;\ThetaVec)\,\mathrm{d}\tilde{\mu}(\zVec) = 0.\nonumber
\end{align}
%\phantom{e}
\ \vspace{-19pt}\\
\end{proof}
%
%\ \vspace{-3ex}\\
%
\noindent{}Just like in Sec.~\ref{SecEntropySums} we will use Lemma~\ref{lem:ParamCrit_general} instead of the parameterization criterion (Definition~\ref{def:Param_Crit}) to proof our main result.  In Appendix~\ref{app:ParaCrit} we will again show that under mild assumptions the criterions given in Lemma~\ref{lem:ParamCrit_general} are actually equivalent to the parameterization criterion.\\%\vspace{1ex}

\noindent{}We are now in the position to prove the generalization of Theorem~\ref{th:Sum_of_Entr}.
%
% \noindent{\bf Theorem} (Sum of Entropies)\\[1ex]
\begin{mytheorem}[Equality to Sums of Pseudo Entropies]
    \label{th:Sum_of_Entr_Gen}
Let $\pPsi(\zVec)$ and $\pT(\xVec\,|\,\zVec)$ be a general EF model as given by Definition~\ref{def:GenEF}, and
let the model fulfill the parameterization criterion of Definition~\ref{def:Param_Crit}. 
Then at all stationary points of the ELBO (\ref{EqnFFGen})
it applies that\vspace{-1.2ex}
%variational lower-bound $\FF(\Phi,\PsiVec,\ThetaVec)$ applies:
%
% $\Theta=\Theta^*$: 
%
\begin{eqnarray}
%
%\FF(\Phi,\PsiVec,\ThetaVec) \hspace{-1ex}&=&\hspace{-1ex}  \frac{1}{N}\sum_{n=1}^N \HH[\qPhiN(\zVec)]  \phantom{ii} \phantom{\sum_{n=1}^N}\hspace{-3ex}-\ \HH[\,\pPsi(\zVec)] \phantom{ii}   -\ \frac{1}{N} \sum_{n=1}^N \EEE{\qn_{\Phi}}{ \HH[\,\pT(\xVec\,|\,\zVec)] }\,. \phantom{\small{}ix}  \\
%
\FFt(\Phi,\PsiVec,\ThetaVec) &=&  \frac{1}{N}\sum_{n=1}^N \HHt[\qPhiN(\zVec)]\ -\  \HHt[\,p_{\PsiVec}(\zVec)]\ -\ \EEE{\;\qBar_{\Phi}}{ \HHt[\,p_{\ThetaVec}(\xVec\,|\,\zVec)] }\,, \phantom{\small{}ix} 
\label{EqnTheoremSoEGen}
\end{eqnarray}
\ \\[-2.5ex]
where $\HHt[\cdot]$ are pseudo entropies as defined in~(\ref{EqnPseudoHOne}), (\ref{EqnPseudoHTwo}) and Appendix~\ref{Suppl:DefPseudoEntropies}, and where the expectation 
can be computed w.r.t.\ the original measure, i.e., $\EEE{\;\qBar_{\Phi}}{f(\zVec)}=\int \qBar_{\Phi}(\zVec)\ f(\zVec)\, \mathrm{d}\mu(\zVec)$.
%
%is w.r.t.\ the measure $\tilde{\mu}(\zVec)$, i.e., 
%$\EEE{\;\qBar_{\Phi}}{f(\zVec)}=\int \qBar_{\Phi}(\zVec)\,f(\zVec)\, \mathrm{d}\tilde{\mu}(\zVec)$.
\end{mytheorem}

\noindent{}The proof of the theorem is provided in Appendix~\ref{app:ProofOfTheoremTwo}.

\ \\[0ex]
\noindent{}It can be verified that Theorem~\ref{th:Sum_of_Entr_Gen} is a genuine generalization of Theorem~\ref{th:Sum_of_Entr} (see Appendix~\ref{Sec:SupplGeneralizationThm}).
As Theorem~\ref{th:Sum_of_Entr_Gen} applies for all general EF models (i.e., no restrictions apply for the 
exponential family distributions), the class of generative models for which Theorem~\ref{th:Sum_of_Entr_Gen}
applies is a proper super-set of the models for which Theorem~\ref{th:Sum_of_Entr} applies. For instance,
Theorem~\ref{th:Sum_of_Entr} can not be applied to mixtures of Poisson distributions, while Theorem~\ref{th:Sum_of_Entr_Gen} is
applicable as the Poisson distribution is an exponetial family distribution.
%
%Theorem~\ref{th:Sum_of_Entr_Gen} can be applied to models with Poisson distributions, while Theorem~\ref{th:Sum_of_Entr} can not.  
%
%
%
%\ \\

\section{Discussion}
\label{SecDiscussion}
We have provided the first general proof for the ELBO to be equal to a sum of entropies at stationary points. The resulting theorems (Theorem~\ref{th:Sum_of_Entr} and \ref{th:Sum_of_Entr_Gen}) apply for a very broad class of models with only mild conditions that have to be satisfied. 
As detailed in the introduction, our investigation of ELBOs at stationary points has its roots in two earlier papers (\cite{LuckeHenniges2012,DammEtAl2023}). Both contributions exploited properties of their corresponding specific models including properties of Gaussian distributions. In contrast, the here presented results show convergence of the ELBO to entropy sums independently of specific models: 
Theorems~\ref{th:Sum_of_Entr} and \ref{th:Sum_of_Entr_Gen} show that convergence to entropy sums is a very common feature of generative models, which does not require properties specific, e.g., to the Gaussian distribution. Instead, for generative models with one set of latents and one set observable variables, it mainly suffices if the model's variables are distributed according to an exponential family distribution. Otherwise only mild assumptions have to be satisfied and, as an integral part of our contribution, we have defined general conditions that imply the ELBO of a generative model to become equal to an entropy sum (Definitions~\ref{def:Param_Crit} and~\ref{def:GenEF}). 

Simplified SBN (Example~\ref{ex:Simple_SBN}) and simplified factor analysis (Example~\ref{ex:SFA}) are examples for how the conditions can be verified for concrete models. Based on Theorems~\ref{th:Sum_of_Entr} and \ref{th:Sum_of_Entr_Gen} a series of standard generative models can now be analyzed. ELBO convergence to entropy sums can thus be shown (see \cite{WarnkenEtAl2024}) for standard SBNs, for probabilistic PCA, factor analysis, Gaussian Mixture Models (GMMs), Poisson Mixture Models and, in general, for mixtures of exponential family distributions. Also the ELBO of standard probabilistic sparse coding \cite{OlshausenField1996} was recently shown to equal an entropy sum at stationary points by applying Theorem~\ref{th:Sum_of_Entr} (see \cite{VelychkoEtAl2024}), which can be verified as special case of the general result represented by~\ref{th:Sum_of_Entr}. Furthermore, the very technical proof for Gaussian VAEs \cite{DammEtAl2023} can be much simplified using Theorem~\ref{th:Sum_of_Entr}, and the theorem provides the deeper theoretical reason why ELBOs of standard VAEs converge to entropy sums. Furthermore, VAEs with non-Gaussian observable distributions (e.g., \cite{ShekhovtsovEtAl2022}) or non-Gaussian prior distributions can now be addressed relatively easily.
% using Theorem~\ref{th:Sum_of_Entr} and \ref{th:Sum_of_Entr_gen}.

Still another line of research could investigate the use of entropy sums as learning objectives. Considering entropy sums as objectives may not be intuitive 
because entropy sums (as they appear in Theorems~\ref{th:Sum_of_Entr} and \ref{th:Sum_of_Entr_Gen}) are by themselves no learning objectives. However, considering the Theorems
%~\ref{th:Sum_of_Entr} and \ref{th:Sum_of_Entr_Gen}
and their proofs, it can be observed that only a subset of parameters has to be at stationary points. All remaining parameters can then be learned
using an entropy sum objective. For probabilistic sparse coding, it was, for instance, shown that analytic solutions can be derived for those parameters of the ELBO that need to
be at stationary points for equality to entropy sums. Using the analytic solution and Theorem~\ref{th:Sum_of_Entr}
then allows for using an objective solely based on entropies. Future work will further investigate those
purely entropy-based objectives. Optimization landscapes can, for instance, be investigated on such grounds: by using Eqns.~\ref{EqnPseudoHOne} and~\ref{EqnPseudoHTwo} further analysis could exploit the relation of derivatives of entropies to Fisher-Rao metrics in information geometry (\cite{Rao1945,Amari2016}). Entropy objectives would also represent a potentially appealing basis for generalizations of ELBO objectives. For instance, \cite{LiTurner2016a} have suggested generalized ELBO objectives using R\'enyi divergences as generalizations of the Kullback-Leibler divergence. Based on an entropy sum objective, generalizations would be possible by directly using R\'enyi entropies\footnote{Indeed, \cite{Renyi1961} first introduced R\'enyi entropies before introducing R\'enyi divergences.}.

Future work may also be able to exploit the relation between studies of convergence guarantees for ELBO objectives and Theorems~\ref{th:Sum_of_Entr} and \ref{th:Sum_of_Entr_Gen}.
One recent such contribution provides convergence rates for Gaussian VAEs \cite{SurendranEtAl2024}. The theoretically quantified convergence rates imply the same rates for the convergence of the ELBO to entropy sums (as explicitly noted in \cite{SurendranEtAl2024}). Vice versa, information about the optimization landscape in the vicinity of stationary points could be provided through entropy sum expressions, which may be useful for the study of convergence guarantees.

Finally, while the class of models that can be addressed using Theorems~\ref{th:Sum_of_Entr} and~\ref{th:Sum_of_Entr_Gen} includes many very standard generative models \cite{WarnkenEtAl2024} and with VAEs also prominent deep models, future work will aim at still further extending the class of considered models. For instance, similar results can be aimed at for generative models defined based on more intricate directed graphical models, which would include deep generative models with stochastic layers (deep SBNs, stacked VAEs, diffusion models etc). Also generative models defined based on undirected graphical models could be addressed, which would include prominent models such as deep Boltzmann machines \cite{AckleyEtAl1985,HintonEtAl2006,FischerIgel2012,Montufar2018} and other deep generative models \cite{BondTaylorEtAl2022}.

\section*{Acknowledgement}
We thank Simon Damm, Dmytro Velychko and Asja Fischer for giving very valuable feedback on earlier versions of this manuscript.
We also used ChatGPT (OpenAI) solely to assist with the wording and language of the manuscript. All scientific content, analyses, and conclusions are our own.

\section*{Funding}
This work was supported by the German Research Foundation (DFG) under project~464104047, `On the Convergence of Variational Deep Learning to Sums of Entropies', within the priority program `Theoretical Foundations of Deep Learning' (SPP 2298).

\appendix

\section*{Appendix}

\section{Parameterization Criterion}
\label{app:ParaCrit}

Here we show that the two statements of Lemma~\ref{lem:ParamCrit} (or Lemma \ref{lem:ParamCrit_general} in the case of a general EF model) % as given in Sec.~\ref{SecGenEF})
are actually equivalent to the parameterization criterion given in Definition~\ref{def:Param_Crit} if we assume few mild additional conditions.
%Consequently, we do not introduce a restriction when using Lemma~\ref{lem:ParamCrit} (or Lemma \ref{lem:ParamCrit_general}) instead of the parameterization criterion.
Consequently, we do not introduce a restriction when using the parameterization criterion instead of Lemma~\ref{lem:ParamCrit} (or Lemma \ref{lem:ParamCrit_general}) directly. %Since we work with a general measure in Sec.~\ref{SecGenEF}, the same applies for the generalization: replacing every $\mathrm{d}\mu(\zVec)$ by $\mathrm{d}\zVec$ generalizes the proof w.r.t.\ the original measures. First, we claim that part~A of Lemma~\ref{lem:ParamCrit} (or Lemma \ref{lem:ParamCrit_general}) is equivalent to part~A of the parameterization criterion. For this we do not need any additional conditions as we can see in the following lemma:
Since we treat general measures in Sec.~\ref{SecGenEF}, we adopt the same approach here. However, the following statements and proofs also apply when we use the notation from Sec.~\ref{SecEntropySums}. We just need to replace each $\mathrm{d}\mu(\zVec)$ by $\mathrm{d}\zVec$. First, we claim that part~A of Lemma~\ref{lem:ParamCrit} (or Lemma \ref{lem:ParamCrit_general}) is equivalent to part~A of the parameterization criterion. For this we do not need any additional conditions as we can see in the following lemma:
\begin{lemma}[Equivalence for the Prior]
Consider an EF generative model as given in Definition~\ref{def:EF_Gen_Model} or a general EF generative model as given in Definition~\ref{def:GenEF} with a mapping to the natural parameters of the prior given by $\zetaVec(\PsiVec)$. The dimensionalities of the vectors $\zetaVec$ and $\PsiVec$ are $K$ and $R$, respectively. Then the following two statements are equivalent:
\begin{itemize}
  \item[(A)] There exists a vectorial function $\alphaVec(\cdot)$ from the set of parameters $\PsiVec$ to $\RRR^R$ such that for all $\PsiVec$ holds:% that fulfills the equation
  
  \begin{eqnarray}
    \zetaVec(\PsiVec) = \delll{\zetaVec(\PsiVec)}{\PsiVecT}\alphaVec(\PsiVec).
  \end{eqnarray}
  \item[(B)] For any vectorial function $\fVec(\Phi,\PsiVec)$ from the sets of parameters $\Phi$ and $\PsiVec$ to the ($K$-dim) space of natural parameters of the prior it holds:
  \begin{eqnarray}
    \label{EqnAppParamCritA}
    \ITnew\fVec(\Phi,\PsiVec) = \vec{0} \ \ \Rightarrow\ \ \zetaVecT_{(\PsiVec)}\fVec(\Phi,\PsiVec) = 0.
  \end{eqnarray}
\end{itemize}
\end{lemma}
\begin{proof}
%\noindent Let us denote the dimension of the space of natural parameters by $K$ and the space of by $S.$ The 
%\noindent For a fixed $\PsiVec,$ the equation bla means, that  Hence, equation is equivalent to
We may rewrite statement~B by using linear algebra. In fact, for fixed $\Phi$ and $\PsiVec$, the first equation in (\ref{EqnAppParamCritA}) is equivalent to the statement that the vector $\fVec(\Phi,\PsiVec)$ is an element of the kernel of the transposed Jacobian matrix $\delll{\zeta^T(\PsiVec)}{\PsiVec}.$ Thus, because $\zetaVecT_{(\PsiVec)}\fVec(\Phi,\PsiVec)$ is the standard Euclidean inner product, statement~B is equivalent to%the second equation in (\ref{EqnAppParamCritA}) is equivalent to
\begin{align*}
  \zetaVec(\PsiVec) \perp \ker \delll{\zetaVecT(\PsiVec)}{\PsiVec}.
\end{align*}
In other words, $\zetaVec(\PsiVec)$ is in the orthogonal complement of the transposed Jacobian, i.e.
\begin{eqnarray}\label{EqnOrthoPrior}
  \zetaVec(\PsiVec)\in \Bigl(\ker \delll{\zetaVecT(\PsiVec)}{\PsiVec}\Bigr)^\perp.
\end{eqnarray}
Denoting the image or range of a matrix $A$ by $\im A$ and using the well known identities $\ker A = (\im A^{\mathrm{T}})^\perp$ and $((\im A)^\perp)^\perp=\im A$ for an arbitrary matrix $A,$ we can rewrite stetement~B of this lemma in the form %Eqn.~\ref{EqnOrthoPrior} and hence statement~B of this lemma in the form
\begin{align*}
  \zetaVec(\PsiVec)\in \im \delll{\zetaVec(\PsiVec)}{\PsiVecT}.
\end{align*}
%%Here we denote by $\im \delll{\zetaVec(\PsiVec)}{\PsiVecT}$ the image of the matrix $\delll{\zetaVec(\PsiVec)}{\PsiVecT}.$ So, this is equivalent to condition (A) and the Lemma is proved.\\[1ex]
By definition of $\im\Inew $, this is equivalent to statement~A and the lemma is proved.\hspace{2mm}
\end{proof}
\noindent{}The equivalence of part~B of Lemma~\ref{lem:ParamCrit} (or Lemma~\ref{lem:ParamCrit_general}) and part~B of the parameterization criterion (Definition~\ref{def:Param_Crit}) is somewhat more intricate. For the proof we interpret the left integral in (\ref{EqnLemmaParamCritB1}) (or \eqref{EqnLemmaParamCritB1_general}) as a bounded linear operator between two suitable Hilbert spaces and compute its adjoint.
More precisely, if we denote the latent space by $\Omega_{\zVec}$ and the dimensionality of the natural parameter vector $\etaVec$ by $L$, we can define for any fixed variational parameter $\Phi$ and any $n\in \{1,\dots,N\}$ the Hilbert space\footnote{Although we should actually consider equivalence classes instead of functions, we will not do this for reasons of convenience. But we keep in mind that the functions may only be defined almost everywhere.}
\begin{align*}
  L^2_{n,\Phi}(\Omega_{\zVec},\RRR^L):=\{\gVec_n\colon \Omega_{\zVec}\rightarrow\RRR^L \mid \int \qn_{\Phi}(\zVec)\, \vert \gVec_n(\zVec)\vert^2 \, \mathrm{d}\mu(\zVec)<\infty \}
\end{align*}
of square integrable functions w.r.t. the weight $\qn_{\Phi}(\zVec)$ and equip it with the inner product
\begin{align*}
  \langle \fVec_n,\gVec_n\rangle_{n,\Phi}:=\int \qn_{\Phi}(\zVec)\, \fVec_n^\mathrm{\,T}(\zVec)\gVec_n(\zVec)\, \mathrm{d}\mu(\zVec),\ \ \ \fVec_n,\gVec_n\in L^2_{n,\Phi}(\Omega_{\zVec},\RRR^L).
\end{align*}
As mentioned above, we here use a general measure $\mu(\zVec)$ on the latent space $\Omega_{\zVec}$. But if we treat the standard Lebesgue measure in the continuous case we can simply replace the $\mathrm{d}\mu(\zVec)$ by $\mathrm{d}\zVec$. Analogously, we can replace the integrals by sums, when the latent variable $\zVec$ is discrete (in this case $\mu(\zVec)$ is the counting measure). Furthermore, in Sec.~\ref{SecGenEF} the measure $\mu(\zVec)$ would play the role of the new measure $\tilde{\mu}(\zVec)$ as given in Definition~\ref{def:Measures} (not of the $\mu(\zVec)$ used there) and $\qn_\Phi(\zVec)$ of the new variational distributions $\tilde{q}_\Phi^{(n)}(\zVec)$ normalized w.r.t. the measure $\tilde{\mu}(\zVec)$. However, for convenience and to be consistent with Sec.~\ref{SecEntropySums}, we omit the tilde here.
%In the case when $\zVec$ is discrete, we have to replace the integrals by sums again. Even then, $L^2_{n,\Phi}(\Omega_{\zVec},\RRR^L)$ with the given inner product is still a Hilbert space. %Alternatively, we could deal with Lebesgue integrals and general measures to summarise the discrete and continuous case. This would also have the advantage that we could absorb the weight $\qn_{\Phi}(\zVec)$ into the measure. But for convenience and by following the standard notation used in Machine Learning, we do not work with general Lebesgue integrals.
%
Either way, the Cartesian product
\begin{align*}
  L^2_\Phi(\Omega_{\zVec},\RRR^L):=L^2_{1,\Phi}(\Omega_{\zVec},\RRR^L)\times\cdots\times L^2_{N,\Phi}(\Omega_{\zVec},\RRR^L)
\end{align*}
of the individual Hilbert spaces together with the inner product
\begin{align*}
  \langle (\fVec_1,\dots,\fVec_N),(\gVec_1,\dots,\gVec_N)\rangle_{\Phi}:=\sum_{n=1}^N \langle \fVec_n,\gVec_n\rangle_{n,\Phi}
\end{align*}
is also a Hilbert space. Since we have to treat the transposed Jacobian matrix $\JTnew$, we also introduce for any fixed variational parameter $\Phi$ and any $n\in \{1,\ldots,N\}$ the Banach space
\begin{align*}
  L^2_{n,\Phi}(\Omega_{\zVec},\RRR^{S\times L}):=\{J\colon \Omega_{\zVec}\rightarrow \RRR^{S\times L}\mid \int \qn_{\Phi}(\zVec)\,\Vert J(\zVec)\Vert^2\, \mathrm{d}\mu(\zVec)<\infty\},
\end{align*}
where $S$ is the dimension of the subset $\thetaVec$ of the noise parameters $\ThetaVec$ we use to construct the Jacobian, and $\Vert\cdot\Vert$ an arbitrary matrix norm on $\RRR^{S\times L}$.

From now on, for any fixed $\ThetaVec$, we will assume the mapping $\etaVec(\zVec;\Theta)$ to be an element of all Hilbert spaces $L^2_{n,\Phi}(\Omega_{\zVec},\RRR^L)$, and the transposed Jacobian $\JTnew$ to be an element of all Banach spaces $L^2_{n,\Phi}(\Omega_{\zVec},\RRR^{S\times L})$.
This assumptions are quite mild, because often only some expectation values w.r.t.\ the variational distributions $\qn_{\Phi}(\zVec)$ need to exist, which is usually the case in machine learning.
Especially, if the latent space $\Omega_{\zVec}$ is finite, since in this case the integrals are replaced by finite sums, which always exist.

After these preparations, we can define the linear bounded operator we want to use for rewriting statement~B of Lemma~\ref{lem:ParamCrit} (or Lemma~\ref{lem:ParamCrit_general}). As the functions $\gVec_n(\zVec;\ThetaVec)$ arising there are arbitrary, we can omit their parameter $\ThetaVec$ to get the same statement (allowing for the treatment of functions $\gVec_n(\zVec)$).
Hence, for all parameters $\Phi$ and $\ThetaVec$ we define the linear operator
\begin{align*}
  %A_{\Phi,\ThetaVec}\colon L^2_{1,\Phi}(\Omega_{\zVec},\RRR^{L})\times \dots\times L^2_{N,\Phi}(\Omega_{\zVec},\RRR^{L})\rightarrow \RRR^{S},\\
  A_{\Phi,\ThetaVec}\colon L^2_{\Phi}(\Omega_{\zVec},\RRR^{L})\rightarrow \RRR^{S},\ \ \ 
  \gVec =(\gVec_1,\dots,\gVec_N)\mapsto \sum_{n=1}^{N}\int \qn_{\Phi}(\zVec)\, \delll{\etaVecT(\zVec;\ThetaVec)}{\thetaVec} \gVec_{n}(\zVec)\, \mathrm{d} \mu(\zVec).
\end{align*}
Remember that the transposed Jacobian $\JTnew$ is a ($S\times L$)-matrix, where $L$ is the dimension of the natural parameter vector $\etaVec$ and $S$ the number of parameters $\thetaVec$ we use to construct the transposed Jacobian $\JTnew$. Using the triangle inequality and Hölder's inequality it is straightforward to see, that $A_{\Phi,\ThetaVec}$ is well-defined and bounded. The boundedness is important for the adjoint to be defined on the hole Hilbert space $\RRR^S.$
\begin{lemma}[The Adjoint of $A_{\Phi,\ThetaVec}$]
\label{lem:Adjoint}
\noindent{}The adjoint operator of $A_{\Phi,\ThetaVec}$ is given by
\begin{align*}
  A^\dagger_{\Phi,\ThetaVec}\colon \RRR^{S}\rightarrow L^2_{\Phi}(\Omega_{\zVec},\RRR^{L}),\ \ \ \betaVec\mapsto A^\dagger_{\Phi,\ThetaVec}\,\betaVec,
  %A^\dagger_{\Phi,\ThetaVec}\colon \RRR^{S}\rightarrow L^2_{1,\Phi}(\Omega_{\zVec},\RRR^{L})\times \cdots\times L^2_{N,\Phi}(\Omega_{\zVec},\RRR^{L}),\ \ \betaVec\mapsto A^\ger_{\Phi,\ThetaVec}\betaVec,
\end{align*}
in which $(A^\dagger_{\Phi,\ThetaVec}\,\betaVec)(\zVec)=\Bigl(\delll{\etaVec(\zVec;\ThetaVec)}{\thetaVecT}\betaVec,\ldots,\delll{\etaVec(\zVec;\ThetaVec)}{\thetaVecT}\betaVec\Bigr)$.
\end{lemma}
\begin{proof}
For any $\gVec=(\gVec_1,\dots,\gVec_N)\in L^2_{\Phi}(\Omega_{\zVec},\RRR^L)$ and any $\betaVec\in \RRR^S$ we have to show that the equation
\begin{align*}
  \langle A_{\Phi,\ThetaVec}\,\gVec\, ,\betaVec\rangle_{\RRR^S} = \langle \gVec\, ,A^\dagger_{\Phi,\ThetaVec}\,\betaVec\rangle_{\Phi}%{L^2_1(\RRR^H,\RRR^L)\times\dots\times L^2_N(\RRR^H,\RRR^L)}
\end{align*}
is fulfilled. We can do this by a direct computation:
\begin{align*}
  \langle A_{\Phi,\ThetaVec}\,\gVec\, , \betaVec\rangle_{\RRR^S}
  &= \langle \sum_{n=1}^N \int \qn_\Phi(\zVec)\, \delll{\etaVecT(\zVec;\ThetaVec)}{\thetaVec} \gVec_n(\zVec)\, \mathrm{d}\mu(\zVec)\, ,\betaVec\rangle_{\RRR^S}\\
  &=\sum_{n=1}^N \int \qn_\Phi(\zVec)\, \langle \delll{\etaVecT(\zVec;\ThetaVec)}{\thetaVec} \gVec_n(\zVec)\, ,\betaVec\rangle_{\RRR^S}\, \mathrm{d}\mu(\zVec)\\
  &=\sum_{n=1}^N \int \qn_\Phi(\zVec)\, \langle \gVec_n(\zVec)\, ,\delll{\etaVec(\zVec;\ThetaVec)}{\thetaVecT}\betaVec\rangle_{\RRR^L}\, \mathrm{d}\mu(\zVec)\\
  &= \langle \gVec\, ,A^\dagger_{\Phi,\ThetaVec}\, \betaVec\rangle_{\Phi}.
\end{align*}
\end{proof}

\noindent{}After these preparations we are now able to show the remaining equivalence of part~B of Lemma~\ref{lem:ParamCrit} (or Lemma~\ref{lem:ParamCrit_general}) and part~B of the parameterization criterion as given in Definition~\ref{def:Param_Crit}:
%\noindent{}To show the equivalence of part~B of Lemma~\ref{lem:ParamCrit} and part~B of the parameterization criterion as given in Definition~\ref{def:Param_Crit}, we need one more additional assumption. That this assumption is not actually necessary for the convergence of the ELBO to entropy sums is discussed below.
%
\begin{lemma}[Equivalence for the Noise Model]
For an EF generative model as given in Definition~\ref{def:EF_Gen_Model} or a general EF generative model as given in Definition~\ref{def:GenEF}, let the natural parameter vector $\etaVec(\zVec;\ThetaVec)$ be in the Hilbert space $L^2_{n,\Phi}(\Omega_{\zVec},\RRR^L)$ and its transposed Jacobian $\JTnew$ in the Banach space $L^2_{n,\Phi}(\Omega_{\zVec},\RRR^{S\times L})$ for all parameters $\Phi$ and $\ThetaVec$ and all $n\in\{1,\ldots,N\}$.
%
%Moreover, assume that the variational distributions are strictly positive, i.e. $\qn_{\Phi}(\zVec)>0$ for all $\zVec$.
%%, and that the subset $\thetaVec$ of $\ThetaVec$ can be chosen independently of $\Phi$.
%
Then the following two statements are equivalent:
\begin{itemize}
  \item[(A)]There are a non-empty subset $\thetaVec$ of $\ThetaVec$, whose dimension we denote by $S$, and a vectorial function $\betaVec(\ThetaVec)$ from the set of hole parameters $\ThetaVec$ to the ($S$-dim) space of $\thetaVec$ that fulfills for almost every $\zVec$ the equation
  \begin{eqnarray}
    \etaVec(\zVec;\ThetaVec)=\delll{\etaVec(\zVec;\ThetaVec)}{\thetaVecT} \betaVec(\ThetaVec).
  \end{eqnarray}
  Here ``for almost every $\zVec$\,'' means that the set of all $\zVec$ which do not fulfill the equation has measure zero w.r.t. the weighted measures $\mu_n(\zVec):=\qn_\Phi(\zVec)\mu(\zVec)$ for all $n$. %The subset $\thetaVec$ has to be non-empty in order for the Jacobian to be defined. Furthermore, notice that $\betaVec$ may only depend on the parameters $\ThetaVec,$ but not on the latent variable $\zVec.$
  \item[(B)] For any variational parameter $\Phi$ there is a non-empty subset $\thetaVec$ of $\ThetaVec$ such that the following applies for all vectorial functions $\gVec_1(\zVec;\ThetaVec),\ldots,\gVec_N(\zVec;\ThetaVec)$ from the parameter set $\ThetaVec$ and from the latent space $\Omega_{\zVec}$ to the ($L$-dim) space of natural parameters of the noise model that are square-integrable $g_n(\zVec;\,\ThetaVec)\in L^2_{n,\Phi}(\Omega_{\zVec},\RRR^L)$ w.r.t. the latent variable $\zVec$:
  \begin{align}
    \hspace{-3mm}\sum_{n=1}^{N} \int \qn_{\Phi}(\zVec)\delll{\etaVecT(z;\ThetaVec)}{\thetaVec}\gVec_n(\zVec;\ThetaVec)\,\mathrm{d}\mu(\zVec) &= \vec{0}\label{EqnAppParamCritB2}\\
    \Rightarrow\  \sum_{n=1}^{N}\int \qn_{\Phi}(\zVec) \etaVecT_{(\zVec;\ThetaVec)}\gVec_n(\zVec;\ThetaVec)\,\mathrm{d}\mu(\zVec) &= 0.\nonumber
  \end{align}
\end{itemize}
Here $\mu(\zVec)$ is a general measure on the latent space $\Omega_{\zVec}$. But in the case of the standard Lebesgue measure, $\mathrm{d}\mu(\zVec)$ can be replaced by $\mathrm{d}\zVec$ and in the case when $\zVec$ is a discrete latent variable, the lemma is true as well if the integral is replaced by a sum over all discrete states of $\zVec$. Furthermore, $\mu(\zVec)$ and $\qn_\Phi(\zVec)$ can be replaced in the context of Sec.~\ref{SecGenEF} by $\tilde{\mu}(\zVec)$ and $\tilde{q}^{(n)}_\Phi(\zVec)$, respectively.
\end{lemma}
\begin{proof}
Since the functions $\gVec_n(\zVec;\ThetaVec)$ are arbitrary anyway, it is straightforward to realise that we can omit the dependence of the parameters $\ThetaVec$ in (B) to get the same statement. Hence, we can treat them w.l.o.g. as functions $\gVec_n(\zVec)\in L^2_{n,\Phi}(\Omega_{\zVec},\RRR^L)$ not depending on $\ThetaVec$.  By using the operator $A_{\Phi,\ThetaVec}$ and the shorter notation $\gVec=(\gVec_1,...,\gVec_N)$, the first equation in~\eqref{EqnAppParamCritB2} can be rewritten as $\gVec\in \ker A_{\Phi,\ThetaVec}.$ Hence, because the second equation in (\ref{EqnAppParamCritB2}) is an inner product in %$\bigl(\etaVec(\zVec;\ThetaVec),\dots,\etaVec(\zVec;\ThetaVec)\bigr) \perp \gVec(\zVec)$ \TODO{Zeilenumbruch} in
the Hilbert space
\begin{align*}
  L^2_\Phi(\Omega_{\zVec},\RRR^L)=L^2_{1,\Phi}(\Omega_{\zVec},\RRR^L)\times\cdots\times L^2_{N,\Phi}(\Omega_{\zVec},\RRR^L),
\end{align*}
statement~B is equivalent to
\begin{eqnarray}\label{EqnAppKernel}
  \bigl(\etaVec(\zVec;\ThetaVec),\dots,\etaVec(\zVec;\ThetaVec)\bigr) \in \bigl(\ker A_{\Phi,\ThetaVec}\bigr)^\perp.
\end{eqnarray}
For a bounded linear operator $A$ between two Hilbert spaces with a finite-dimensional image, the two identities $\ker A = (\im A^\dagger)^\perp$ and $((\im A^\dagger)^\perp)^\perp = \im A^\dagger$ hold. As before, we denote by $\im A^\dagger$ the image or range of the adjoint operator $A^\dagger$. %, which has a finite dimension and is hence closed.
Therefore, (\ref{EqnAppKernel}) is equivalent to
\begin{align*}
  \bigl(\etaVec(\zVec;\ThetaVec),\dots,\etaVec(\zVec;\ThetaVec)\bigr) \in \im A_{\Phi,\ThetaVec}^\dagger\, .
\end{align*}
By definition and by Lemma~\ref{lem:Adjoint}, there is a vetcor $\betaVec\in \RRR^S$ such that
\begin{align}\label{EqnEquality_in_L2}
  \bigl(\etaVec(\zVec;\ThetaVec),\dots,\etaVec(\zVec;\ThetaVec)\bigr)=\Bigl(\delll{\etaVec(\zVec;\ThetaVec)}{\thetaVecT}\betaVec,\ldots,\delll{\etaVec(\zVec;\ThetaVec)}{\thetaVecT}\betaVec\Bigr).
\end{align}
Because this is an equality in $L^2_\Phi(\Omega_{\zVec},\RRR^L)$, Eqn.~\ref{EqnEquality_in_L2} is equivalent to
%\begin{align}
%  L^2_{1,\Phi}(\Omega_{\zVec},\RRR^L)\times\cdots\times L^2_{N,\Phi}(\Omega_{\zVec},\RRR^L),
%\end{align}
%and all $\qn_{\Phi}(\zVec)$ are non-zero, (\ref{EqnEquality_in_L2}) is equivalent to
%
\begin{align}
  \etaVec(\zVec;\ThetaVec) = \delll{\etaVec(\zVec;\ThetaVec)}{\thetaVecT}\betaVec,\label{Eqn:dependenceofPhi}
\end{align}
for almost every $\zVec$. Since the operator $A_{\Phi,\ThetaVec}$ and its adjoint $A^\dagger_{\Phi,\ThetaVec}$ depend on the parameters $\Phi$ and $\ThetaVec$, the vector $\betaVec$ may also depend on $\Phi$ and $\ThetaVec$. However, the dependency on $\Phi$ does not matter, as in \eqref{Eqn:dependenceofPhi}, the natural parameters $\etaVec(\zVec;\ThetaVec)$ and their Jacobian $\Jnew$ do not depend on $\Phi$.
\end{proof}
%Because the operator $A_{\Phi,\ThetaVec}$ and therefore the adjoint $A^\dagger_{\Phi,\ThetaVec}$ depend on the parameters, the vector $\betaVec$ may also depend on $\Phi$ and $\ThetaVec$. But since the natural parameter vector $\etaVec(\zVec;\ThetaVec)$ is independent of $\Phi$, we get statement~A.
%%\TODO{write that the assumption are usually fulfilled, especially in the examples of the paper. Moreover there is a trick with $\langle \zVec\rangle$}
%
%\ \\
%
%
%
%
\section{Proof of Theorem~\ref{th:Sum_of_Entr_Gen}}
\label{app:ProofOfTheoremTwo}

We here provide the proof of Theorem~\ref{th:Sum_of_Entr_Gen}.
In a preparation and in analogy to Sec.\,\ref{SecEntropySums} (see Eqns.\,\ref{EqnFFMain} to \ref{EqnFFThree}), we first
decompose the ELBO (\ref{EqnFFGen}) into three summands:
%
%\begin{eqnarray}
%
% \zVec &\& p_{\zetaVec(\Psi,\Theta)}(\zVec) \\ 
% \FF_1 &=& \HH\big( \qn(\zVec)\big) \,.\\ 
%
%\end{eqnarray}
%
\begin{align}
\tilde{\FF}(\Phi,\PsiVec,\ThetaVec) &=  \Tilde{\FF}_1(\Phi) - \tilde{\FF}_2(\Phi,\PsiVec) - \tilde{\FF}_3(\Phi,\ThetaVec),\ \text{\ \ where}\label{EqnFFTMain}\\
\tilde{\FF}_1(\Phi)\phantom{i}\      &=   \textstyle-\, \frac{1}{N}\sum_{n} \int \tilde{q}^{(n)}_{\Phi}\!(\zVec) \log\!\big( \tilde{q}^{(n)}_{\Phi}\!(\zVec) \big)\, \mathrm{d}\tilde{\mu}(\zVec)\label{EqnFFTOne},\\
\tilde{\FF}_2(\Phi,\PsiVec)\ &= -\, \textstyle\frac{1}{N}\sum_{n} \int \tilde{q}^{(n)}_{\Phi}\!(\zVec) \log\!\big( \pt_{\PsiVec}(\zVec) \big)\, \mathrm{d}\tilde{\mu}(\zVec)\label{EqnFFTTwo},\\
\tilde{\FF}_3(\Phi,\ThetaVec)\ &= -\, \textstyle\frac{1}{N}\sum_{n} \int \tilde{q}^{(n)}_{\Phi}\!(\zVec) \log\!\big( \pt_{\ThetaVec}(\xVecN\,|\,\zVec) \big)\, \mathrm{d}\tilde{\mu}(\zVec).\label{EqnFFTThree}
\end{align}

\begin{proof}[Proof of Theorem~\ref{th:Sum_of_Entr_Gen}.]
The first summand of the ELBO (\ref{EqnFFTMain}) we observe to take the form of (an average of) a pseudo entropy $\tilde{\FF}_1(\Phi) = \frac{1}{N}\sum_n\HHt\big[ \qn_\Phi(\zVec)\big]$.

It remains to generalize the proof of Theorem \ref{th:Sum_of_Entr}. For this, observe that all terms of the ELBO (see Eqn.~\ref{EqnFFTMain}) are defined analogously as in \eqref{EqnFFMain} only the base measure has changed. Second, observe that all derivation steps of the proof of Theorem~\ref{th:Sum_of_Entr} used the standard properties of integration that apply to any measure for integration. As $\pt_{\zetaVec}(\zVec)$ and $\pt_{\etaVec}(\xVec)$ are probability distributions for the measures $\tilde{\mu}(\zVec)$ and $\tilde{\mu}(\xVec)$, respectively, they represent probability distributions with constant (and unit) base measure in the probability spaces defined by these measures, respectively. The proof can consequently proceed along the same lines as for Theorem~\ref{th:Sum_of_Entr}. 

\ \\
%\newpage
\noindent\underline{$\tilde{\FF}_2(\Phi,\PsiVec)$ at stationary points}\\[1ex]
%
% so we start by considering the second summand (Eqn.\,\ref{EqnFFTwo}). 
\noindent{}As in the proof of Theorem~\ref{th:Sum_of_Entr} we want to rewrite the second summand $\tilde{\FF}_2(\Phi,\PsiVec)$ of the ELBO (see Eqn.~\ref{EqnFFTTwo}) at all stationary points. To do so, we again compute the negative logarithm of $\pt_{\PsiVec}(\zVec)$ that is now given by
\begin{align}
-\log\big(\pt_{\PsiVec}(\zVec)\big)=-\log\big( \pt_{\zetaVec(\PsiVec)} (\zVec) \big) &= A\big(\zetaVec(\PsiVec)\,\big) \,-\, \zetaVecT_{(\PsiVec)} \vec{T}(\zVec) \big)\,, \label{EqnLogPriorGEN}
\end{align}
since $\pt_{\zetaVec}(\zVec)$ has unit base measure. Furthermore, the derivative of the negative logarithm w.r.t. $\PsiVec$ is given by
\begin{align}
\dell{\PsiVec} \Big( -\log\big( \pt_{\zetaVec(\PsiVec)} (\zVec) \big) \Big) &= \ITnew \Big(  \AVec'\big(\zetaVec(\PsiVec)\big)\,-\,\vec{T}(\zVec)  \Big)\,,\label{EqnDelLogPriorGEN}\\
                   &\ \ \ \mbox{where}\ \ \AVec'(\zetaVec(\PsiVec)\,):=\dell{\zetaVec}\,A(\zetaVec\,)\Big|_{\zetaVec=\zetaVec(\PsiVec)}.\label{EqnAPrimePriorGEN}
\end{align}
%
%
%\begin{align}
%
%\FF_2(\Phi,\PsiVec)\ &= -\, \textstyle\frac{1}{N}\sum_{n} \int \qn_{\Phi}\!(\zVec) \log\!\big( p_{\PsiVec}(\zVec) \big)\hspace{0.5ex} \mathrm{d}\zVec\nonumber\\
%
%                    &= \phantom{-}\, \textstyle\frac{1}{N}\sum_{n} \int \qn_{\Phi}\!(\zVec) \Big( - \log\!%\big( p_{\zetaVec(\PsiVec)}(\zVec) \big) \Big) \hspace{0.5ex} \mathrm{d}\zVec\,.\label{EqnFFSecondGEN}
%
%\end{align}
%
%As $p_{\zetaVec\,}(\zVec)$ is an exponential family distribution, 
%
%The negative logarithm of $p_{\zetaVec\,}(\zVec)$ as exponential family distribution can be written as:
%
%\begin{align}
%
%-\log\big( p_{\zetaVec(\PsiVec)} (\zVec) \big) &= A\big(\zetaVec(\PsiVec)\,\big) \,-\, \zetaVecT_{(\PsiVec)} \vec{T}(\zVec)  \,-\, \log\big( h(\zVec) \big)\,, \label{EqnLogJointGEN}
%
%\end{align}
%
Here $A$ is again the log-partition function, $\vec{T}$ the sufficient statistics and $\ITnew$ the transposed Jacobian matrix defined in (\ref{lem:ITmatrix}). %We will show that the summand $\tilde{\FF}_2(\Phi,\PsiVec)$ will at stationary points be equal to the prior pseudo entropy. Therefore, we first rewrite $-\log\!\big( p_{\zetaVec(\PsiVec)}(\zVec) \big)$ using the prior entropy.
By using Eqn.~\ref{EqnPseudoHOne} to compute the pseudo entropy of the prior
\begin{align}
\tilde{\mathcal{H}}[p_{\PsiVec}(\zVec)] = \tilde{\mathcal{H}}[p_{\zetaVec(\PsiVec)}(\zVec)] &= -\zetaVecT_{(\PsiVec)} \AVec^\prime(\zetaVec(\PsiVec))+A(\zetaVec(\PsiVec))
\end{align}
and by using (\ref{EqnLogPriorGEN}) we obtain:
%
%
%The last equality for $B(\zetaVec)$ follows if we use our assumption of a constant base measure $h(\zVec)=h_o$. 
%
%We now use (\ref{EqnEntropyExpression}) to replace the partition function $A\big(\zetaVec\,\big)$ by an expression
%containing the entropy:
 %
%\begin{align}
%
%
%A\big( \zetaVec(\PsiVec)  \big)  &= \HH[\,p_{\zetaVec(\PsiVec)}(\zVec)]  \,+\, \zetaVecT_{(\PsiVec)} \AVec'\big(\zetaVec(\PsiVec)\big) \,+\, B(\zetaVec(\PsiVec)) \label{EqnAasH}
%\end{align}
%
%By combining (\ref{EqnLogJointGEN}) and (\ref{EqnAasHGEN}), the negative logarithm of the prior probability becomes:
%
\begin{align}
- \log\!\big( \pt_{\zetaVec(\PsiVec)}(\zVec)\big) = \HHt[\,p_{\zetaVec(\PsiVec)}(\zVec)] \,+\, \zetaVecT_{(\PsiVec)} \big( \AVec'(\zetaVec{(\PsiVec)})\,-\,\vec{T}(\zVec) \big). \label{EqnNegLogGEN}
%
%\,+\, B(\zetaVec(\PsiVec))\,-\,\log\big(h(\zVec)\big)\,, 
%
%\ \mbox{where}\ \AVec'(\zetaVec)\ \mbox{as in Eqn.\,\ref{EqnAPrime}}.
%
\end{align}
%
%, the last two terms cancel.
Inserting the negative logarithm of the prior probability density (Eqn.~\ref{EqnNegLogGEN}) into $\tilde{\mathcal{F}}_2(\Phi,\PsiVec)$ (see Eqn.~\ref{EqnFFTTwo}), we therefore obtain for the second summand of the ELBO:
%
%The last two terms that depend on the base measure $h(\zVec)$ of the prior cancel because we assumed constant base measure. We therefore obtain for $\FF_2(\Psi,\Theta)$:
%
%
\begin{align}
\tilde{\FF}_2(\Phi,\PsiVec)\ &= -\, \textstyle\frac{1}{N}\sum_{n} \int \tilde{q}^{(n)}_{\Phi}\!(\zVec) \log\!\big( \pt_{\zetaVec(\PsiVec)}(\zVec) \big)\hspace{0.5ex} \mathrm{d}\tilde{\mu}(\zVec)= \textstyle\HHt[\,p_{\zetaVec(\PsiVec)}(\zVec)] \,+\, \zetaVecT_{(\PsiVec)} \ \vec{f}(\Phi,\PsiVec)\, ,\label{EqnFFTwoShortGEN}\\[1ex]
%
%\label{EqnFFTwoShortGEN}
                   &\ \ \mbox{where}\ \  \textstyle\vec{f}(\Phi,\PsiVec) :=  \AVec'(\zetaVec(\PsiVec))\,-\,\frac{1}{N}\sum_n\mathbb{E}_{\tilde{q}^{(n)}_\Phi}  \vec{T}(\zVec)\,. \label{EqnFunctionFGEN}
%                   
%                   \ \ \mbox{and}\ \ \ \qbar_{\Phi}(\zVec) :=   \frac{1}{N}\sum_{n} \qn_{\Phi}(\zVec)  
%
\end{align}
So, it remains to show that the last term of \eqref{EqnFFTwoShortGEN} disappears at all stationary points of $\tilde{\FF}(\Phi,\PsiVec,\ThetaVec)$. Precisely in the same way as in the proof of Theroem~\ref{th:Sum_of_Entr}, the only summand of $\tilde{\FF}(\Phi,\PsiVec,\ThetaVec)$ which depends on $\PsiVec$ is $\tilde{\FF}_2(\Phi,\PsiVec)$. We therefore obtain at stationary points:
% we consequently obtain:
%
\begin{align}
\vec{0} =\disT \del{\PsiVec} \tilde{\FF}_2(\Phi,\PsiVec)
                    &= \phantom{-}\, \textstyle\frac{1}{N}\sum_{n} \int \qn_{\Phi}\!(\zVec)\  \del{\PsiVec} \Big( - \log\!\big( p_{\zetaVec(\PsiVec)}(\zVec) \big) \Big)\, \hspace{0.5ex} \mathrm{d}\tilde{\mu}(\zVec)\nonumber\\
                    &= \phantom{-}\, \textstyle\frac{1}{N}\sum_{n} \int \qn_{\Phi}\!(\zVec)\ \ITnew \Big( \big( \AVec'(\zetaVec(\PsiVec))\,-\,\vec{T}(\zVec) \big) \Big) \hspace{0.5ex} \mathrm{d}\tilde{\mu}(\zVec)\nonumber\\
                    &= \phantom{-}\, \textstyle\ITnew\ \big( \AVec'(\zetaVec_{(\PsiVec)})\,-\,\frac{1}{N}\sum_n\mathbb{E}_{\tilde{q}^{(n)}_\Phi}  \vec{T}(\zVec) \big)=\ITnew\ \vec{f}(\Phi,\PsiVec)\,,\label{EqnDerivativeF2generalEF}
%
%                    &= \phantom{-}\, \IIT_{(\PsiVec)}\ \vec{f}(\Phi,\PsiVec) \label{}
%
\end{align}
%
%where $\qbar_{\Phi}$ is defined in (\ref{EqnQBar}).
% and 
%
%Hence, at all stationary points (w.r.t.\,parameters $\PsiVec$) the following concise condition holds:
%
%\begin{align}
%
% \ITnew\ \vec{f}(\Phi,\PsiVec) &= \vec{0} \label{EqnConditionPriorGEN}\,,
%
%\end{align}
%
where $\vec{f}(\Phi,\PsiVec)$ is the same function as introduced in (\ref{EqnFunctionFGEN}). Now we can apply Lemma~\ref{lem:ParamCrit_general}, since we also assumed our general EF generative
model to fulfill the parameterization criterion given in Definition~\ref{def:Param_Crit}. 
%
%\textcolor{red}{Folgendes vielleicht löschen: Part~A of the lemma is formulated for arbitrary functions $\vec{f}$ with arbitrary parameters $\Phi$ and arbitrary values for $\ThetaVec$.
%So if the criterion is satisfied, (\ref{EqnLemmaParamCritA}) also applies for the $\vec{f}$ as defined in (\ref{EqnFunctionF}) with $\Phi$ being the variational
%parameters.
%
%was formulated without the function $\vec{f}$
%depending on $\Phi$. But the variational parameters $\Phi$ are treated as arbitrary constant values throughout the
%proof and can therefore be treated as part of a fixed function (with arguments $\PsiVec$).
%
%Using Lemma~\ref{lem:ParamCrit} we thus conclude from (\ref{EqnConditionPrior}) that at stationary points applies:} 
%
We thus conclude from \eqref{EqnDerivativeF2generalEF} that at all stationary points applies:
%
%Throughout the proof, we variational 
%
%If for the considered generative model the parameterization criterion of Definition~C holds, we can conclude from (\ref{EqnConditionPrior}) that
%at all stationary points (w.r.t.\,parameters $\PsiVec$) holds:
%
\begin{align}
\zetaVecT_{(\PsiVec)} \ \vec{f}(\Phi,\PsiVec) &= 0\,.
\end{align}
Using Eqn.~\eqref{EqnFFTwoShortGEN} we consequently obtain at all stationary points for the summand $\tilde{\FF}_2(\Phi,\PsiVec)$:
\begin{align}
\tilde{\FF}_2(\Phi,\PsiVec) &= \HHt[\,p_{\zetaVec(\PsiVec)}(\zVec)] \,+\, \zetaVecT_{(\PsiVec)} \ \vec{f}(\Phi,\PsiVec) = \HHt[\,p_{\zetaVec(\PsiVec)}(\zVec)] = \HHt[\,p_{\PsiVec}(\zVec)]\,.
\label{EqnResultFFTwoGEN}
\end{align}
\ \\ \ \\
%\newpage
\noindent\underline{$\tilde{\FF}_3(\Phi,\ThetaVec)$ at stationary points}\\[1ex]
%
% so we start by considering the second summand (Eqn.\,\ref{EqnFFTwo}).
\noindent{}To show convergence for the third summand of the ELBO \eqref{EqnFFTMain}, nothing significant changes compared to the second summand $\tilde{\mathcal{F}}_2(\Phi,\ThetaVec)$ or to the proof of Theorem~\ref{th:Sum_of_Entr}. In fact, the negative logarithm of $\pt_{\ThetaVec}(\xVec\,|\,\zVec)$ takes the form %The proof part for $\FF_3(\Phi,\ThetaVec)$ will be analog to that of $\FF_2(\Phi,\PsiVec)$ above but slightly more intricate because the noise distribution is conditional on $\zVec$. We have assumed that the generative model is an EF generative model. Therefore we can rewrite $\FF_3(\Phi,\ThetaVec)$ of Eqn.\,\ref{EqnFFThree} using the reparameterization of $p_{\ThetaVec}(\xVec\,|\,\zVec)$ in terms of the exponential family distribution $p_{\etaVec(\zVec;\,\ThetaVec)}(\xVec)$:
%
%
%\begin{align}
%
%\FF_3(\Phi,\ThetaVec)\ &= -\, \textstyle\frac{1}{N}\sum_{n} \int \qn_{\Phi}\!(\zVec) \log\!\big( p_{\ThetaVec}(\xVecN\,|\,\zVec) \big)\hspace{0.5ex} \mathrm{d}\zVec\nonumber\\
%
%                    &= \phantom{-}\, \textstyle\frac{1}{N}\sum_{n} \int \qn_{\Phi}\!(\zVec) \Big( - \log\!\big( p_{\etaVec(\zVec;\,\ThetaVec)}(\xVecN) \big) \Big) \hspace{0.5ex} \mathrm{d}\zVec\,. \label{EqnFFThirdGEN}
%
%\end{align}
%
%
%As $p_{\etaVec} (\xVec)$ is an exponential family distribution, its negative logarithm is given by:
%
%
\begin{align}
%
%-\log\big( p_{\etaVec} (\xVec) \big) &=A\big(\etaVec\,\big) \,-\, \etaVecT \vec{T}(\xVec) \,-\, \log\big( h(\xVec) \big)\,. \label{EqnLogJointThreeGEN}
%
-\log\big(\pt_{\ThetaVec}(\xVec\,|\,\zVec)\big)=-\log\big( \pt_{\etaVec(\zVec;\,\ThetaVec)} (\xVec) \big) &=A\big(\etaVec(\zVec;\,\ThetaVec)\,\big) \,-\, \etaVecT_{(\zVec;\,\ThetaVec)} \vec{T}(\xVec)\,. \label{EqnLogJointThreeGEN}
\end{align}
Here we again use the same symbols for sufficient statistics $\vec{T}(\xVec)$ and partition function $A(\etaVec)$ as for the prior in Eqn.\,\ref{EqnLogPriorGEN}. Accordingly, for the derivative of the negative logarithm applies:
%Note that we will use the same symbols for sufficient statistics $\vec{T}(\xVec)$, partition function $A(\etaVec)$ and base measure $h(\xVec)$ as for the
%prior in Eqn.\,\ref{EqnLogJoint}. As prior and noise model distribution are (in general) different members of the exponential family, these entities (in general) differ, of course. But as they can be distinguished from context, we avoid the introduction of further symbols.  
%
%We will later require the derivative of the (negative) log probability w.r.t.\ $\thetaVec$, i.e., w.r.t.\ a subset of the parameters $\ThetaVec$.
%The derivative is given by:
%
\begin{align}
\dell{\thetaVec} \Big( -\log\big( \pt_{\etaVec(\zVec;\,\ThetaVec)} (\xVec) \big) \Big) &= \JTnew \Big( \big( \AVec'(\etaVec(\zVec; \ThetaVec))\,-\,\vec{T}(\xVec) \big) \Big)\,,\label{EqnDelLogThreeGEN}\\
                   &\ \ \ \mbox{where}\ \ \AVec'(\etaVec(\zVec; \ThetaVec)\,):=\dell{\etaVec}\,A(\etaVec\,)\Big|_{\etaVec=\etaVec(\zVec; \ThetaVec)}\label{EqnAPrimeThreeGEN}
%
%A\big(\zetaVec(\PsiVec)\,\big) \,-\, \zetaVecT_{(\PsiVec)} \vec{T}(\zVec)  \,-\, \log\big( h(\zVec) \big)\,. \label{EqnLogJointGEN}
%
\end{align}
and where $\JTnew$ is the transposed Jacobian matrix defined in (\ref{lem:JTmatrix}). Furthermore, by using Eqn.~\ref{EqnPseudoHTwo} to compute the pseudo entropy of the noise model
%
%We will show that the summand $\FF_3(\Phi,\ThetaVec)$ will at stationary points be equal to the noise model entropy. Therefore, we first rewrite $-\log\!\big( p_{\etaVec(\zVec; \ThetaVec)}(\xVec) \big)$ using the noise model entropy. By again using the definition of the entropy
%
%\begin{align}
%
%\HH[\,p_{\etaVec(\zVec; \ThetaVec)}(\xVec)] &= \int p_{\etaVec(\zVec; \ThetaVec)}(\xVec) \Big( -\log\big( p_{\etaVec(\zVec; \ThetaVec)} (\xVec) \big) \Big)\mathrm{d}\xVec
%
%\end{align}
%
%and (\ref{EqnLogJointThree}) we obtain:
%
\begin{align}
\HHt[\,p_{\ThetaVec}(\xVec\,|\,\zVec)] &= \HHt[\,p_{\etaVec(\zVec; \ThetaVec)}(\xVec)] = A\big( \etaVec(\zVec; \ThetaVec)  \big) \,-\, \etaVecT_{(\zVec;\,\ThetaVec)} \AVec'\big(\etaVec(\zVec; \ThetaVec)\big)\,, \label{EqnEntropyExpressionThreeGEN}%\\[1ex]
%
%\Rightarrow\ A\big( \etaVec(\zVec; \ThetaVec)  \big)  &= \HH[\,p_{\etaVec(\zVec; \ThetaVec)}(\xVec)]  \,+\, \etaVecT_{(\zVec;\,\ThetaVec)} \AVec'\big(\etaVec(\zVec; \ThetaVec)\big) \,+\, B(\etaVec(\zVec; \ThetaVec))\,, \label{EqnAasHThreeGEN}\\[1ex]
%
%
%                    &\mbox{where we abbreviated}\ \ %B(\etaVec\,):= \disT\int p_{\etaVec}(\xVec) \,\log\big(h(\xVec)%\big)\, \mathrm{d}\xVec \label{EqnBPrimeThreeGEN}\,. %\,=\, %h_o\,. 
%
\end{align}
%
%The last equality for $B(\zetaVec)$ follows if we use our assumption of a constant base measure $h(\zVec)=h_o$. 
%
%We now use (\ref{EqnEntropyExpression}) to replace the partition function $A\big(\zetaVec\,\big)$ by an expression
%containing the entropy:
 %
%\begin{align}
%
%
%A\big( \etaVec(\zVec; \ThetaVec)  \big)  &= \HH[\,p_{\etaVec(\zVec; \ThetaVec)}(\zVec)]  \,+\, \etaVecT_{(\zVec;\,\ThetaVec)} \AVec'\big(\etaVec(\zVec; \ThetaVec)\big) \,+\, B(\etaVec(\zVec; \ThetaVec)) \label{EqnAasHGEN}
%\end{align}
%
the negative logarithm of the probability density then takes the form
%By combining (\ref{EqnLogJointThree}) and (\ref{EqnAasHThree}), the negative logarithm of the noise distribution becomes:
%
\begin{align}
- \log\!\big( \pt_{\etaVec(\zVec; \ThetaVec)}(\xVec)\big)
  &= \HHt[\,p_{\etaVec(\zVec; \ThetaVec)}(\xVec)] \,+\, \etaVecT_{(\zVec;\,\ThetaVec)} \big( \AVec'(\etaVec{(\zVec; \ThetaVec)})\,-\,\vec{T}(\xVec) \big)\,, \label{EqnNegLogThreeGEN}
\end{align}
where $\AVec'\big(\etaVec(\zVec;\,\ThetaVec)\big)$ is defined as in Eqn.\,\ref{EqnAPrimeThreeGEN}. Inserting the negative logarithm (\ref{EqnNegLogThreeGEN}) into $\tilde{\FF}_3(\Phi,\ThetaVec)$ (see Eqn.~\ref{EqnFFTThree}), we therefore obtain for the third summand of the ELBO:
%
%The last two terms that depend on the base measure $h(\xVec)$ of the prior cancel because we assumed constant base measure. We therefore obtain for $\FF_2(\Psi,\Theta)$:
%
%
\begin{align}
\tilde{\FF}_3(\Phi,\ThetaVec)&=\frac{1}{N}\sum_{n} \int \tilde{q}^{(n)}_{\Phi}\!(\zVec) \Big(-\,\log\!\big( \pt_{\etaVec(\zVec;\,\ThetaVec)}(\xVecN) \big)\Big)\, \mathrm{d}\tilde{\mu}(\zVec)\nonumber\\
&= \frac{1}{N}\sum_{n} \int \tilde{q}^{(n)}_{\Phi}\!(\zVec) \Big( \HHt[\,p_{\etaVec(\zVec; \ThetaVec)}(\xVec)] \,+\, \etaVecT_{(\zVec;\,\ThetaVec)} \big( \AVec'(\etaVec{(\zVec; \ThetaVec)})\,-\,\vec{T}(\xVecN) \big) \Big) \hspace{0.5ex} \mathrm{d}\tilde{\mu}(\zVec)\nonumber\\
&= \frac{1}{N}\sum_{n} \int q^{(n)}_{\Phi}\!(\zVec)  \HHt[\,p_{\etaVec(\zVec; \ThetaVec)}(\xVec)]\,\mathrm{d}\mu(\zVec)\nonumber\\
&\hspace{15mm}+ \frac{1}{N}\sum_{n} \int \tilde{q}^{(n)}_{\Phi}\!(\zVec)\, \etaVecT_{(\zVec;\,\ThetaVec)} \vec{g}_n(\zVec;\ThetaVec) \hspace{0.5ex} \mathrm{d}\tilde{\mu}(\zVec)\nonumber\\
                   &= \frac{1}{N} \sum_{n}\EEE{\,q^{(n)}_\Phi}{ \HHt[\,p_{\etaVec(\zVec; \ThetaVec)}(\xVec)] } \,+\, \frac{1}{N}\sum_{n}\int \tilde{q}^{(n)}_{\Phi}\!(\zVec)\ \etaVecT_{(\zVec;\,\ThetaVec)}\ \vec{g}_n(\zVec;\ThetaVec)\,\mathrm{d}\tilde{\mu}(\zVec)\,, \label{EqnFFThreeShortGEN}\\
                   &\hspace{35mm}\mbox{where}\ \  \vec{g}_n(\zVec;\ThetaVec) := \AVec'\big(\etaVec(\zVec;\,\ThetaVec)\big)\,-\,\vec{T}(\xVecN)\,. \label{EqnFunctionGGEN}
\end{align}
In analogy to the derivation for $\FFt_2(\Phi,\PsiVec)$ and to the proof of Theorem~\ref{th:Sum_of_Entr}, the only summand of the ELBO that depends on the noise model parameters $\ThetaVec$ is $\FFt_3(\Phi,\ThetaVec)$, Therefore, %at every stationary points of the ELBO it applies:% our aim will now be to show that the second term of (\ref{EqnFFThreeShort}) vanishes at stationary points of $\FF(\Phi,\PsiVec,\ThetaVec)$. For all stationary points of $\FF(\Phi,\PsiVec,\ThetaVec)$ applies that the derivatives w.r.t.\ all parameters vanish. 
%
%We have assumed the parameters $\Phi$, $\PsiVec$ and $\ThetaVec$ to be separate sets of parameters. $\FF_3(\Phi,\ThetaVec)$ is therefore the only summand of $\FF(\Phi,\PsiVec,\ThetaVec)$ which depends on $\ThetaVec$.
%
at stationary points we can conclude for any subset $\thetaVec$ of $\ThetaVec$ that $\del{\thetaVec}\, \FFt_3(\Phi,\ThetaVec)=0$. 
By using the definition of $\FFt_3(\Phi,\ThetaVec)$ (see Eqn.~\ref{EqnFFTThree}) and Eqn. \ref{EqnDelLogThreeGEN}, we obtain at any stationary point and for any subset $\thetaVec$:
% we consequently obtain:
%
\begin{align}
\vec{0} =\disT \del{\thetaVec}\, \FFt_3(\Phi,\ThetaVec)
                    &= \, \frac{1}{N}\sum_{n} \int \tilde{q}^{(n)}_{\Phi}\!(\zVec)\  \del{\thetaVec} \Big( - \log\!\big( \pt_{\etaVec(\zVec; \ThetaVec)}(\xVec^{(n)}) \big) \Big)\, \hspace{0.5ex} \mathrm{d}\tilde{\mu}(\zVec)\nonumber\\
                    &= \,\frac{1}{N}\sum_{n} \int \tilde{q}^{(n)}_{\Phi}\!(\zVec)\ \JTnew \Big( \AVec'(\etaVec(\zVec; \ThetaVec))\,-\,\vec{T}(\xVec^{(n)}) \Big) \hspace{0.5ex} \mathrm{d}\tilde{\mu}(\zVec)\nonumber\\
&=\,\frac{1}{N}\sum_{n} \int \tilde{q}^{(n)}_{\Phi}\!(\zVec)\ \JTnew\ \vec{g}_n(\zVec; \ThetaVec)\,\mathrm{d}\tilde{\mu}(\zVec)\,, \label{EqnDelFFThreeGEN}         
\end{align}
%
%where $\qbar_{\Phi}$ is defined in (\ref{EqnQBarThree}).
% and 
%
where $\vec{g}_n(\zVec; \ThetaVec)$ are the same functions as introduced in \eqref{EqnFunctionGGEN}. Again, we can use Lemma~\ref{lem:ParamCrit_general} and conclude from \eqref{EqnDelFFThreeGEN} that at all stationary points applies:
%In expression (\ref{EqnDelFFThreeGEN}) we recognize the functions $\vec{g}_n(\zVec; \ThetaVec)$ as defined in Eqn.\,\ref{EqnFunctionGGEN}.
%Therefore, expression~(\ref{EqnDelFFThreeGEN}) means that at all stationary points and for all subsets $\thetaVec$ of $\ThetaVec$ the
%following holds:
%
%\begin{align}
%
%  \frac{1}{N} \sum_{n}\int \qn_{\Phi}\!(\zVec)\ \JTnew\ \vec{g}_n(\zVec; \ThetaVec)\,\mathrm{d}\zVec &= 0 \label{EqnConditionNoiseGEN}\,.
%
%\end{align}
%
%where $\vec{f}(\Phi,\PsiVec)$ is the same function as introduced in (\ref{EqnFunctionF}).
%
%As (\ref{EqnConditionNoise}) applies for all subsets $\thetaVec$, it also applies for the specific subset $\thetaVec$ of $\ThetaVec$ 
%that exists according to the parameterization criterion (Definition~\ref{def:Param_Crit}, part~B) or Lemma~\ref{lem:ParamCrit} (part~B). Using Part~B of Lemma~\ref{lem:ParamCrit}, we
%can consequently conclude from (\ref{EqnConditionNoise}) that at all stationary points applies:
%
\begin{align}
\frac{1}{N}\sum_{n}\int \tilde{q}^{(n)}_{\Phi}\!(\zVec)\ \etaVec^{\mathrm{T}}_{(\zVec;\ThetaVec)}\ \gVec_n(\zVec;\ThetaVec) \,\mathrm{d}\tilde{\mu}(\zVec)\ =\ 0\,. \label{EqnConditionNoiseInProofGEN}
\end{align}
By going back to (\ref{EqnFFThreeShortGEN}) and by inserting (\ref{EqnConditionNoiseInProofGEN}), we consequently obtain at all stationary points for the summand $\FFt_3(\Phi,\ThetaVec)$:
\begin{align}
 \FFt_3(\Phi,\ThetaVec) &= \frac{1}{N} \sum_{n}\EEE{\,q^{(n)}_\Phi}{ \HHt[\,p_{\etaVec(\zVec; \ThetaVec)}(\xVec)] } \,+\, \frac{1}{N}\sum_{n}\int \tilde{q}^{(n)}_{\Phi}\!(\zVec)\ \etaVecT_{(\zVec;\,\ThetaVec)}\ \vec{g}_n(\zVec; \ThetaVec)\ \mathrm{d}\tilde{\mu}(\zVec) \nonumber\\
&= \frac{1}{N} \sum_{n}\EEE{\,q^{(n)}_\Phi}{ \HHt[\,p_{\etaVec(\zVec; \ThetaVec)}(\xVec)] } = \frac{1}{N} \sum_{n}\EEE{\,q^{(n)}_\Phi}{ \HHt[\,p_{\ThetaVec}(\xVec\,|\,\zVec)] }\,.\label{EqnResultFFThreeGEN}
\end{align}

\vspace{20mm}
\noindent\underline{Conclusion}\\[1ex]
%
%The ELBO (\ref{EqnFFMain}) 
%
We conclude as in the proof of Theorem~\ref{th:Sum_of_Entr}, noting that we started by rewriting the ELBO consisting of three terms. In this case: $\FFt(\Phi,\PsiVec,\ThetaVec) =  \FFt_1(\Phi) - \FFt_2(\Phi,\PsiVec) - \FFt_3(\Phi,\ThetaVec)$,
%
%\begin{align}
%
%\FF(\Phi,\PsiVec,\ThetaVec) &=  \FF_1(\Phi) - \FF_2(\Phi,\PsiVec) - \FF_3(\Phi,\ThetaVec)
%
%\end{align}
%
with $\FFt_1(\Phi)$, $\FFt_2(\Phi,\PsiVec)$ and $\FFt_3(\Phi,\ThetaVec)$ as defined by (\ref{EqnFFTOne}), (\ref{EqnFFTTwo}) and (\ref{EqnFFTThree}), respectively.
The first term is by definition an average of pseudo entropies, $\FFt_1(\Phi)=\frac{1}{N}\sum_{n=1}^N \HHt[\qPhiN(\zVec)]$. We have shown (see Eqn.\,\ref{EqnResultFFTwoGEN}) that $\FFt_2(\Phi,\PsiVec)$ becomes equal to the pseudo entropy of the prior distribution at stationary points. Furthermore, we have shown (see Eqn.\,\ref{EqnResultFFThreeGEN}) that $\FFt_3(\Phi,\ThetaVec)$ becomes equal to an expected pseudo entropy at stationary points. By inserting the results (\ref{EqnResultFFTwoGEN}) and (\ref{EqnResultFFThreeGEN}), we have shown that at all stationary points the following applies for the ELBO (\ref{EqnFFGen}):
%
% is equal at all stationary points applies:
%
\begin{align}
\FFt(\Phi,\PsiVec,\ThetaVec) &= \disT \frac{1}{N}\sum_{n=1}^N \HHt[\qPhiN(\zVec)]   - \HHt[\,\pPsi(\zVec)] - \frac{1}{N} \sum_{n=1}^N \EEE{\qn_{\Phi}}{ \HHt[\,\pT(\xVec\,|\,\zVec)] },
\end{align}
from which (\ref{EqnTheoremSoEGen}) follows by using the aggregated posterior (see Eqn. \ref{EqnAgPost}) for the last summand.
%which proves the claim.\\[1ex]
%\noindent{}$\square$%\\
%\ \\
%\noindent{}
%
%
\end{proof}
\section{Entropies, Log-Likelihood, and Free Energy in the New Probability Spaces}\label{Suppl:PseudoEntropies}
In this section, we again provide the definitions of entropies, log-likelihood and free energy with respect to the new measures introduced in Section~\ref{SecGenEF}. However, the definition of the entropy here is somewhat more general than the one from Section~\ref{SecGenEF}, as we do not restrict ourselves to exponential family distributions. This generalization is necessary, since we do not demand that the variational distribution $\qPhiN(\zVec)$ belong to an exponential family. Moreover, we discuss the connections to the conventional quantities.
\subsection{Definitions and Basic Properties}\label{Suppl:DefPseudoEntropies}
%As a further preparation, we will require entropies and likelihood defined in the new probability spaces.
We start with the definition of entropies in the new probability space that we define in analogy to standard entropies.

\begin{definition}[Pseudo Entropies]
    \label{def:APPPseudoH}
Consider a general EF model from Definition~\ref{def:GenEF} with probability densities $p_{\zeta}(\zVec)$ and $p_{\etaVec}(\xVec)$ defined w.r.t.\ the measures $\mu(\zVec)$ and $\mu(\xVec)$, respectively. Now consider the new measures $\tilde{\mu}(\zVec)$ and $\tilde{\mu}(\xVec)$ introduced by Defintion~\ref{def:Measures}.

If $p(\zVec)$ is a probability density defined w.r.t.\ the original measure $\mu(\zVec)$, then 
$\tilde{p}(\zVec)=\frac{1}{h(\zVec)}\,p(\zVec)$ is a probability density w.r.t.\ the new measure $\tilde{\mu}(\zVec)$.
We can then define the {\em pseudo entropy} of $p(\zVec)$ as follows:
\begin{eqnarray}
 \HHt[\,p(\zVec)] &=& -\int \pt(\zVec) \log \pt(\zVec)\, \mathrm{d}\tilde{\mu}(\zVec) \,.
\end{eqnarray}
Analogously, if $p(\xVec)$ is a probability density defined w.r.t.\ the original measure $\mu(\xVec)$, then 
$\tilde{p}(\xVec)=\frac{1}{h(\xVec)}\,p(\xVec)$ is a probability density w.r.t.\ the new measure $\tilde{\mu}(\xVec)$.
We can then define the {\em pseudo entropy} of $p(\xVec)$ as follows:
\begin{eqnarray}
 \HHt[\,p(\xVec)] &=& -\int \pt(\xVec) \log \pt(\xVec)\, \mathrm{d}\tilde{\mu}(\xVec) \,.
\end{eqnarray}
\end{definition}
%\ \\[0ex]

\noindent{}Pseudo entropies can, hence, be defined for any (well-behaved) distributions. For the exponential family distributions
 $p_{\zetaVec}(\zVec)$ and $p_{\etaVec}(\xVec)$, the pseudo entropies take on a particularly concise and convenient form, however. 
%
%Pseudo entropies are also defined for any other distributions in the new probability spaces. For instance, for
%a distribution $q(\zVec)$ the pseudo entropy is defined by:
%
%
%
%Since the distributions $\tilde{p}_{\zetaVec}(\zVec)$ and $\tilde{p}_{\etaVec}(\xVec)$ are densities of an exponential family, the pseudo entropies can be expressed very concisely, and similary to conventional entropies.
%
%closed form that we now compute. This leads to similar expressions as for the conventional entropies of $p_{\zetaVec}(\zVec)$ and $p_{\etaVec}(\xVec)$. %We discuss the relation between entropies and pseudoentropies in more detail in the appendix.
%
\begin{lemma}
	\label{lem:AppPseudoEntropies}
For a general EF model as given in Definition \ref{def:GenEF} consider the new measures $\tilde{\mu}(\zVec)$ and $\tilde{\mu}(\xVec)$ introduced in Definition~\ref{def:Measures}. Then it applies for the pseudo entropies of Definition~\ref{def:APPPseudoH} that
\begin{align}
	\tilde{\mathcal{H}}[p_{\PsiVec}(\zVec)] &= \tilde{\mathcal{H}}[p_{\zetaVec(\PsiVec)}(\zVec)]
	= -\zetaVecT_{(\PsiVec)} \AVec^\prime(\zetaVec(\PsiVec))+A(\zetaVec(\PsiVec))\label{EqnAppPseudoEntropyLatents}\ \mbox{and}\\
	\tilde{\mathcal{H}}[p_{\ThetaVec}(\xVec\,|\,\zVec)] &= \tilde{\mathcal{H}}[p_{\etaVec(\zVec; \ThetaVec)}(\xVec)]
	= -\etaVecT_{(\zVec;\ThetaVec)} \AVec^\prime(\etaVec(\zVec;\ThetaVec))+A(\etaVec(\zVec;\ThetaVec))\,,\label{EqnAppPseudoEntropyObservables}
\end{align}
where $\AVec'(\zetaVec(\PsiVec)\,):=\dell{\zetaVec}\,A(\zetaVec\,)\Big|_{\zetaVec=\zetaVec(\PsiVec)}$ and $\AVec'(\etaVec(\zVec; \ThetaVec)\,):=\dell{\etaVec}\,A(\etaVec\,)\Big|_{\etaVec=\etaVec(\zVec; \ThetaVec)}$ are the gradients of the log-partition functions with respect to the natural parameters.
\end{lemma}
%\ \\
\begin{proof}
%
%Because the proofs for the latent variables und the observables do not differ, we only show equation~\eqref{EqnPseudoEntropyObservables}.
%
We only show Eqn.~\ref{EqnAppPseudoEntropyObservables}, since Eqn.~\ref{EqnAppPseudoEntropyLatents} follows along the same lines. Abbreviating $\etaVec=\etaVec(\zVec; \ThetaVec)$ we obtain:
%
%is basically the same.
%
\begin{align}
   &\HHt[\,p_{\etaVec}(\xVec)] = -\int \pt_{\etaVec}(\xVec) \log \pt_{\etaVec}(\xVec) \,\mathrm{d}\tilde{\mu}(\xVec)=-\int p_{\etaVec}(\xVec) \log \pt_{\etaVec}(\xVec) \,\mathrm{d}\mu(\xVec)\nonumber\\
   &=-\int p_{\etaVec}(\xVec)\Big(\etaVecT \TVec(\xVec)+A(\etaVec)\Big)\,\mathrm{d}\mu(\xVec)=-\etaVecT \AVec^\prime(\etaVec)+A(\etaVec)\,,\nonumber
\end{align}
where we applied the standard result $\AVec'(\zetaVec) = \dell{\etaVec}\,A(\etaVec\,) = \disT\int p_{\etaVec}(\xVec)\,\TVec(\xVec)\,\mathrm{d}\mu(\xVec)$ (note that $\mu(\xVec)$ is the orginial/old measure again).
\end{proof}
%consider the modified densities $\tilde{p}_{\PsiVec}(\zVec)$ and $\tilde{p}_{\ThetaVec}(\xVec^{(n)}\,|\,\zVec)$. The the following applies to their pseudo entropies:
%
\noindent{}Considering Lemma~\ref{lem:AppPseudoEntropies}, the pseudo entropies of distributions $p_{\PsiVec}(\zVec)$ and $p_{\ThetaVec}(\xVec\,|\,\zVec)$ can simply be computed from the mappings to natural parameters and their corresponding log-partition functions.\\

\noindent{}We now define the log-likelihood w.r.t.\ the new measure $\tilde{\mu}(\zVec)$.
\begin{definition}[Pseudo Log-Likelihood]
    \label{def:PseudoLL}
Consider a general EF model as given by Definition \ref{def:GenEF}. Let the measures $\tilde{\mu}(\zVec)$ and $\tilde{\mu}(\xVec)$ as well as the densities $\pt_{\zetaVec}(\zVec)$ and $\pt_{\etaVec}(\xVec)$ be defined as in Definition~\ref{def:Measures}. In analogy to the standard log-likelihood $\LL(\PsiVec,\ThetaVec)$, we define the pseudo log-likelihood function $\LLt(\PsiVec,\ThetaVec)$ to be given by:
\begin{align}
\LLt(\PsiVec,\ThetaVec) &= \frac{1}{N} \sum_n \log\Big( \int \pt_{\ThetaVec} (\xVec^{(n)}\,|\,\zVec)\, \pt_{\PsiVec} (\zVec)\,\mathrm{d}\Tilde{\mu}(\zVec) \Big)\\
  &= \frac{1}{N} \sum_n \log\Big( \int \pt_{\etaVec(\zVec; \ThetaVec)} (\xVec^{(n)})\, \pt_{\zetaVec(\PsiVec)} (\zVec)\,\mathrm{d}\Tilde{\mu}(\zVec) \Big).
\end{align}
\end{definition}
\noindent{}It is straightforward to relate the pseudo log-likelihood $\LLt(\PsiVec,\ThetaVec)$ to the standard log-likelihood.
\begin{lemma}
    \label{lemma:PropPseudoLL}
Consider a general EF model as given by Definition \ref{def:GenEF} and the corresponding pseudo log-likelihood as given in Definition \ref{def:PseudoLL}. Then the difference to the standard log-likelihood is a constant not depending on any parameters:
\begin{eqnarray}
	\LLt(\PsiVec,\ThetaVec) &=& \LL(\PsiVec,\ThetaVec) \,-\, \frac{1}{N}\sum_n \log\big( h(\xVecN) \big).
\end{eqnarray}
\end{lemma}
\begin{proof}
Starting from the standard log-likelihood, we obtain: %can calculate:
%
%\mycomment{Use $\mu$ and not $\tilde{\mu}$ where appropriate ...}
%
\begin{eqnarray}
 \LL(\PsiVec,\ThetaVec) &=& \frac{1}{N} \sum_{n} \log \Big(\int p_{\etaVec(\zVec;\,\ThetaVec)}(x^{(n)})\, p_{\zetaVec(\PsiVec)}(\zVec)\,\mathrm{d}\mu(\zVec)\Big)\nonumber\\
 &=& \frac{1}{N} \sum_{n} \log\Big(h(\xVec^{(n)})\int\Tilde{p}_{\etaVec(\zVec;\,\ThetaVec)}(\xVec^{(n)})\, \Tilde{p}_{\zetaVec(\PsiVec)}(\zVec)\,h(\zVec)\,\mathrm{d}\mu(\zVec)\Big)\nonumber\\
 &=& \frac{1}{N} \sum_{n} \log  h(\xVec^{(n)})+\frac{1}{N} \sum_{n} \log \int\Tilde{p}_{\etaVec(\zVec;\,\ThetaVec)}(\xVec^{(n)})\, \Tilde{p}_{\zetaVec(\PsiVec)}(\zVec)\,\mathrm{d}\Tilde{\mu}(\zVec).\nonumber
\end{eqnarray}
If we subtract the term $\frac{1}{N}\sum_n \log h(\xVec^{(n)})$ on both sides, together with the Definition of the pseudo log-likelihood (Definition \ref{def:PseudoLL}), we obtain the claim. %equation stated above.\\
\end{proof}
\noindent{}By virtue of Lemma~\ref{lemma:PropPseudoLL}, parameter optimization using the pseudo log-like\-lihood is equivalent to parameter optimization using the standard log-likelihood. % (the offset is independent of the model parameters). %y just differ by an offset 
The main technical difference will be the use of probability measures $\tilde{\mu}(\zVec)$ and $\tilde{\mu}(\xVec)$. Furthermore, we can also derive a lower bound for the pseudo log-likelihood, just as we do for the standard log-likelihood. %To start, observe that it is straightforward to derive a lower (ELBO) bound of the pseudo log-likelihood.
By applying the probabilistic version of Jensen's inequality (e.g. \cite{Athreya2006}) we obtain:
\begin{align}
%
% \zVec &\& p_{\zetaVec(\Psi,\Theta)}(\zVec) \\ 
\lefteqn{ \LLt(\PsiVec,\ThetaVec) }\nonumber\\
=\ & \frac{1}{N} \sum_n \log\Big( \int \pt_{\ThetaVec} (\xVec^{(n)}\,|\,\zVec)\, \pt_{\PsiVec} (\zVec)\,\mathrm{d}\Tilde{\mu}(\zVec) \Big)\nonumber\\
=\ &\disS \frac{1}{N}\sum_n \log \Big(\mathbb{E}_{\tilde{q}^{(n)}_\Phi}\Big\{\frac{\pt_{\ThetaVec}(\xVec^{(n)}\,|\,\zVec)\, \pt_{\PsiVec}(\zVec)}{\tilde{q}^{(n)}_\Phi(\zVec)}\Big\}\Big)
\hspace{0ex} \,\geq\, \hspace{0ex}\frac{1}{N}\sum_n \mathbb{E}_{\tilde{q}^{(n)}_\Phi}\Big\{\log \Big(\frac{\pt_{\ThetaVec}(\xVec^{(n)}\,|\,\zVec)\, \pt_{\PsiVec}(\zVec)}{\tilde{q}^{(n)}_\Phi(\zVec)}\Big)\Big\}\nonumber\\
  =\ &\frac{1}{N}\sum_n \int \tilde{q}^{(n)}_\Phi(\zVec) \log\Big(\frac{\pt_{\ThetaVec}(\xVec^{(n)}\,|\,\zVec)\, \pt_{\PsiVec}(\zVec)}{\tilde{q}^{(n)}_\Phi(\zVec)}\Big) \mathrm{d}\tilde{\mu}(\zVec)\nonumber\\
=\ &  \frac{1}{N}\sum_n \int \tilde{q}^{(n)}_\Phi(\zVec) \log\big( \pt_{\ThetaVec}(\xVec^{(n)}\,|\,\zVec)\, \pt_{\PsiVec}(\zVec)\big)\, \mathrm{d}\tilde{\mu}(\zVec) 
  \,-\, \frac{1}{N}\sum_n \int \tilde{q}^{(n)}_\Phi(\zVec) \log \tilde{q}^{(n)}_\Phi(\zVec)\, \mathrm{d}\tilde{\mu}(\zVec)\nonumber\\
  =:\ &\FFt(\Phi,\PsiVec,\ThetaVec),\label{EqnAppFFGen} 
\end{align}
where the variational distributions $\tilde{q}^{(n)}_\Phi(\zVec)$ are normalized w.r.t.\ the new measure $\tilde{\mu}(\zVec)$. %=h(\zVec)\mu(\zVec)$.\\[2ex]

\subsection{Relation to Entropy and Free Energy}
\label{app:PseudoEntropies}
%
%Now we have shown that the statements given in Lemma~\ref{lem:ParamCrit} (or in the case of a general EF generative model Lemma~\ref{lem:ParamCrit_general}) are actually equivalent to the parameterization criterion given in Definition~\ref{def:Param_Crit}.
In the following two lemmas we provide, first, the relation between the pseudo entropies of a general EF generative model from Sec.~\ref{SecGenEF} and the conventional entropies, and, second, the relation between the ELBOs $\FFt(\Phi,\PsiVec,\ThetaVec)$ and $\FF(\Phi,\PsiVec,\ThetaVec)$.
%
%\begin{lemma}
% (Connection between Entropies and Pseudo Entropies)
%\end{lemma}
%
%\defStart{}Consider a general EF generative model as given in Definition~\ref{def:GenEF} und the new measures $\tilde{\mu}(\zVec)$ and $\tilde{\mu}(\xVec)$ as given in Definition~\ref{def:Measures}. Moreover, let the new variational distributions
%
\begin{lemma}[Relation between Entropies and Pseudo Entropies]
	\label{lem:ConnectionEntropies}
Consider a general EF model as given in Definition~\ref{def:GenEF} with densities $p_{\zeta}(\zVec)$ and $p_{\etaVec}(\xVec)$ w.r.t.\ measures $\mu(\zVec)$ and $\mu(\xVec)$. Let $\tilde{\mu}(\zVec)$ and $\tilde{\mu}(\xVec)$ be the new measures as given in Definition~\ref{def:Measures}. If the variational distributions $\qn_\Phi(\zVec)$ and $\tilde{q}_\Phi^{(n)}(\zVec)$ w.r.t.\ the measures $\mu(\zVec)$ and $\tilde{\mu}(\zVec)$, respectively, are related by the equation $\qn_\Phi(\zVec)=h(\zVec)\,\tilde{q}_\Phi^{(n)}(\zVec)$, then the following applies:
\begin{itemize}
	\item[(A)] $\HHt[p_{\zetaVec}(\zVec)] = \HH[p_{\zetaVec}(\zVec)]+\int p_{\zetaVec}(\zVec) \log\big(h(\zVec)\big) \mathrm{d}\mu(\zVec)\, ,$
	\item[(B)] $\HHt[p_{\etaVec}(\xVec)] = \HH[p_{\etaVec}(\xVec)]+\int p_{\etaVec}(\xVec) \log\big(h(\xVec)\big) \mathrm{d}\mu(\xVec)\, ,$
	\item[(C)] $\HHt[\qn_{\Phi}(\zVec)] = \HH[\qn_{\Phi}(\zVec)]+\int \qn_{\Phi}(\zVec) \log\big(h(\zVec)\big) \mathrm{d}\mu(\zVec)\, .$
\end{itemize}
\end{lemma}
\begin{proof}
In order to proof the first equation we derive:
\begin{align*}
	&\HH[p_{\zetaVec}(\zVec)] = \textstyle-\int p_{\zetaVec}(\zVec) \log\big(p_{\zetaVec}(\zVec)\big) \mathrm{d}\mu(\zVec)\\
	&=\textstyle-\int p_{\zetaVec}(\zVec) \log\big(\pt_{\zetaVec}(\zVec)\big) \mathrm{d}\mu(\zVec)-\int p_{\zetaVec}(\zVec) \log\big(h(\zVec)\big) \mathrm{d}\mu(\zVec)\\
	&=\textstyle-\int \pt_{\zetaVec}(\zVec) \log\big(\pt_{\zetaVec}(\zVec)\big) \mathrm{d}\tilde{\mu}(\zVec)-\int p_{\zetaVec}(\zVec) \log\big(h(\zVec)\big) \mathrm{d}\mu(\zVec)\\
	&=\textstyle\HHt[p_{\zetaVec}(\zVec)]-\int p_{\zetaVec}(\zVec) \log\big(h(\zVec)\big) \mathrm{d}\mu(\zVec)\, .
\end{align*}
By adding the second term of the last equation on both sides we get statement~A. The other two statements can be shown analogously. We only have to replace the density $p_{\zetaVec}(\zVec)$ by the densities $p_{\etaVec}(\xVec)$ and $\qn_\Phi(\zVec)$. In the case of statement~B, we must also replace the measures $\mu(\zVec)$ and $\tilde{\mu}(\zVec)$ by $\mu(\xVec)$ and $\tilde{\mu}(\xVec)$, respectively.\\[0ex]
\end{proof}

\begin{lemma}[Relation between ELBOs w.r.t.\ different measures]\label{def:SupplPseudoELBO}
Consider a general EF generative model as given in Definition~\ref{def:GenEF} with densities $p_{\zeta}(\zVec)$ and $p_{\etaVec}(\xVec)$ w.r.t. measures $\mu(\zVec)$ and $\mu(\xVec)$. Let $\tilde{\mu}(\zVec)$ and $\tilde{\mu}(\xVec)$ be new measures as given in Definition~\ref{def:Measures}. If the variational distributions $\qn_\Phi(\zVec)$ and $\tilde{q}_\Phi^{(n)}(\zVec)$ w.r.t. the measures $\mu(\zVec)$ and $\tilde{\mu}(\zVec)$, respectively, are related by the equation $\qn_\Phi(\zVec)=h(\zVec)\tilde{q}_\Phi^{(n)}(\zVec)$, then the following applies:
\begin{align*}
	\tilde{\mathcal{F}}(\Phi,\PsiVec,\ThetaVec)=\mathcal{F}(\Phi,\PsiVec,\ThetaVec)-\frac{1}{N}\sum_n\log \big(h(\xVec^{(n)})\big)\, .
\end{align*}
\end{lemma}
\begin{proof}
In Eqns. \ref{EqnFFMain} and \ref{EqnFFTMain} we have decomposed the ELBOs into three summands, respectively. We now show a relationship between each summand of $\FF(\Phi,\PsiVec,\ThetaVec)$ and $\FFt(\Phi,\PsiVec,\ThetaVec)$. For the first one, we can easily apply Lemma~\ref{lem:ConnectionEntropies}, since this is the average (pseudo) entropy of the variational distribution:
%\begin{align}
%\mathcal{F}_1(\Phi)&=-\frac{1}{N}\sum_n \int \qn_\Phi(\zVec) \log \big(\qn_\Phi(\zVec)\big) \,\mathrm{d}\mu(\zVec)\\
%&=-\frac{1}{N}\sum_n \int \qn_\Phi(\zVec) \log \big(\tilde{q}^{(n)}_\Phi(\zVec)\big) \,\mathrm{d}\mu(\zVec)-\frac{1}{N}\sum_n \int \qn_\Phi(\zVec) \log \big(h(\zVec)\big) \,\mathrm{d}\mu(\zVec)\\
%&=-\frac{1}{N}\sum_n \int \tilde{q}^{(n)}_\Phi(\zVec) \log \big(\tilde{q}^{(n)}_\Phi(\zVec)\big) \,\mathrm{d}\tilde{\mu}(\zVec)-\frac{1}{N}\sum_n \int \qn_\Phi(\zVec) \log \big(h(\zVec)\big) \,\mathrm{d}\mu(\zVec)\\
%&=\tilde{\mathcal{F}}_1(\Phi)-\frac{1}{N}\sum_n \int \qn_\Phi(\zVec) \log \big(h(\zVec)\big) \,\mathrm{d}\mu(\zVec)\,.
%\end{align}
\begin{align*}
\tilde{\mathcal{F}}_1(\Phi)&=\textstyle\frac{1}{N}\sum_n \HHt[\qn_{\Phi}(\zVec)]=\textstyle\frac{1}{N}\sum_n \HHt[\qn_{\Phi}(\zVec)]+\frac{1}{N}\sum_n \int \qn_{\Phi}(\zVec) \log\big(h(\zVec)\big) \mathrm{d}\mu(\zVec)\\
&=\textstyle\mathcal{F}_1(\Phi)+\frac{1}{N}\sum_n\int \qn_{\Phi}(\zVec) \log\big(h(\zVec)\big) \mathrm{d}\mu(\zVec)\,.
\end{align*}
For the second summand, we explicitly calculate:
\begin{align*}
\mathcal{F}_2(\Phi,\PsiVec)&=\textstyle-\frac{1}{N}\sum_n \int \qn_\Phi(\zVec) \log \big(p_{\zetaVec}\,(\zVec)\big) \,\mathrm{d}\mu(\zVec)\\
&=\textstyle-\frac{1}{N}\sum_n \int \qn_\Phi(\zVec) \log \big(\pt_{\zetaVec}\,(\zVec)\big) \,\mathrm{d}\mu(\zVec)-\frac{1}{N}\sum_n \int \qn_\Phi(\zVec) \log \big(h(\zVec)\big) \,\mathrm{d}\mu(\zVec)\\
&=\textstyle-\frac{1}{N}\sum_n \int \tilde{q}^{(n)}_\Phi(\zVec) \log \big(\pt_{\zetaVec}\,(\zVec)\big) \,\mathrm{d}\tilde{\mu}(\zVec)-\frac{1}{N}\sum_n \int \qn_\Phi(\zVec) \log \big(h(\zVec)\big) \,\mathrm{d}\mu(\zVec)\\
&=\textstyle\tilde{\mathcal{F}}_2(\Phi,\PsiVec)-\frac{1}{N}\sum_n \int \qn_\Phi(\zVec) \log \big(h(\zVec)\big) \,\mathrm{d}\mu(\zVec)\,.
\end{align*}
Consequently, the two expressions $\mathcal{F}_1(\Phi)-\mathcal{F}_2(\Phi,\PsiVec)$ and $\tilde{\mathcal{F}}_1(\Phi)-\tilde{\mathcal{F}}_2(\Phi,\PsiVec)$ coincide. So, the difference originates from the third summand of the ELBOs:
\begin{align*}
\mathcal{F}_3(\Phi,\ThetaVec)&=\textstyle-\frac{1}{N}\sum_n \int \qn_\Phi(\zVec) \log \big(p_{\etaVec}\,(\xVec^{(n)})\big) \,\mathrm{d}\mu(\zVec)\\
&=\textstyle-\frac{1}{N}\sum_n \int \qn_\Phi(\zVec) \log \big(\pt_{\etaVec}\,(\xVec^{(n)})\big) \,\mathrm{d}\mu(\zVec)\\
&\hspace{10mm}\textstyle-\frac{1}{N}\sum_n \int \qn_\Phi(\zVec) \log \big(h(\xVec^{(n)})\big) \,\mathrm{d}\mu(\zVec)\\
&=\textstyle-\frac{1}{N}\sum_n \int \tilde{q}^{(n)}_\Phi(\zVec) \log \big(\pt_{\etaVec}\,(\xVec^{(n)})\big) \,\mathrm{d}\tilde{\mu}(\zVec)-\frac{1}{N}\sum_n\log \big(h(\xVec^{(n)})\big)\\
&=\textstyle\tilde{\mathcal{F}}_3(\Phi,\ThetaVec) -\frac{1}{N}\sum_n\log \big(h(\xVec^{(n)})\big)\,.
\end{align*}
If we insert everything into the ELBOs, we finally get:
\begin{align*}
    \textstyle\tilde{\mathcal{F}}(\Phi,\PsiVec,\ThetaVec) &= \tilde{\mathcal{F}}_1(\Phi)-\tilde{\mathcal{F}}_2(\Phi,\PsiVec)-\tilde{\mathcal{F}}_3(\Phi,\ThetaVec)\\
    &=\textstyle\mathcal{F}_1(\Phi)-\mathcal{F}_2(\Phi,\PsiVec)-\mathcal{F}_3(\Phi,\ThetaVec)-\frac{1}{N}\sum_n\log \big(h(\xVec^{(n)})\big)\\
    &=\textstyle\mathcal{F}(\Phi,\PsiVec,\ThetaVec)-\frac{1}{N}\sum_n\log \big(h(\xVec^{(n)})\big)\, .
\end{align*}
\end{proof}

%
% transfer proof of theorem 2 here
%
%
%
%
%
%
%
\subsection{Theorem~\ref{th:Sum_of_Entr_Gen} as Generalization of Theorem~\ref{th:Sum_of_Entr}}\label{Sec:SupplGeneralizationThm}
We have not verified that Theorem~\ref{th:Sum_of_Entr_Gen} is a genuine generalization of Theorem~\ref{th:Sum_of_Entr}, yet.
%Finally, let us verify that Theorem~\ref{th:Sum_of_Entr_Gen} is a genuine generalization of Theorem~\ref{th:Sum_of_Entr}. 
In order to do so, let us apply Theorem~\ref{th:Sum_of_Entr_Gen} to a model that also fulfills the conditions for Theorem~\ref{th:Sum_of_Entr}. Concretely, 
we consider an EF model (Def.\,\ref{def:EF_Gen_Model}) that fulfills the parameterization condition (Def.\,\ref{def:Param_Crit}). 
An EF model has exponential family distributions $p_{\PsiVec}(\zVec)$ and $p_{\ThetaVec}(\xVec\,|\,\zVec)$ with
constant base measures. For such a model, the lower bound used by Theorem~\ref{th:Sum_of_Entr_Gen}, i.e.\ $\FFt(\Phi,\PsiVec,\ThetaVec)$, 
and the lower bound used by Theorem~\ref{th:Sum_of_Entr}, i.e. $\FF(\Phi,\PsiVec,\ThetaVec)$, are related as follows (see Lemma~\ref{def:SupplPseudoELBO}):
%
%in (\ref{...}). If $h(\xVec)=h_o$ is the base measure
%of the observable distribution, the relation is given by (see Appendix~\ref{AppTheoRelation}):
%
%Using constant base measures, it is straightforward to relate the lower bound of Theorem~\ref{th:Sum_of_Entr_Gen}, i.e.\ $\FFt(\Phi,\PsiVec,\ThetaVec)$, to the standard lower bound $\FF(\Phi,\PsiVec,\ThetaVec)$ in (\ref{...}). If $h(\xVec)=h_o$ is the base measure
%of the observable distribution, the relation is given by (see Appendix~\ref{AppTheoRelation}):
%
\begin{align}
\FFt(\Phi,\PsiVec,\ThetaVec) &= \FF(\Phi,\PsiVec,\ThetaVec) \,-\, \log\big( h_0 \big)\,.\label{EqnConsistRelFF}
\end{align}
Analogously, the pseudo entropies are equal to the corresponding standard entropies except of offsets
given by the corresponding base measures:
\begin{align}
\HHt[\qPhiN(\zVec)] &= \HH[\qPhiN(\zVec)] \,+\, \log\big( h^\mathrm{prior}_0 \big)\,,\label{EqnConsistRelA}\\
\HHt[p_{\PsiVec}(\zVec)] &= \HH[p_{\PsiVec}(\zVec)] \,+\, \log\big( h^\mathrm{prior}_0 \big)\,,\label{EqnConsistRelB}\\
\HHt[p_{\ThetaVec}(\xVec\,|\,\zVec)] &= \HH[p_{\ThetaVec}(\xVec\,|\,\zVec)] \,+\, \log\big( h_0 \big)\,,\label{EqnConsistRelC}
\end{align}
where we have used $h^\mathrm{prior}_0$ to distinguish the constant base measure of the prior distribution, $h(\zVec)=h^\mathrm{prior}_0$, from the
constant base measure of the noise distribution, $h(\xVec)=h_0$.

Now for the considered EF model Theorem~\ref{th:Sum_of_Entr_Gen} applies. If we insert the relations (\ref{EqnConsistRelFF}) to (\ref{EqnConsistRelC}) into (\ref{EqnTheoremSoEGen}) of the theorem, we obtain:
\begin{align}
%
%\FF(\Phi,\PsiVec,\ThetaVec) \hspace{-1ex}=&\hspace{-1ex}  \frac{1}{N}\sum_{n=1}^N \HH[\qPhiN(\zVec)]  \phantom{ii} \phantom{\sum_{n=1}^N}\hspace{-3ex}-\ \HH[\,\pPsi(\zVec)] \phantom{ii}   -\ \frac{1}{N} \sum_{n=1}^N \EEE{\qn_{\Phi}}{ \HH[\,\pT(\xVec\,|\,\zVec)] }\,. \phantom{\small{}ix}  \\
%
\FFt(\Phi,\PsiVec,\ThetaVec) =&  \frac{1}{N}\sum_{n=1}^N \HHt[\qPhiN(\zVec)]  \phantom{ii} \phantom{\sum_{n=1}^N}\hspace{-3ex}-\ \HHt[\,p_{\PsiVec}(\zVec)] \phantom{ii}   -\ \EEE{\;\qBar_{\Phi}}{ \HHt[\,p_{\ThetaVec}(\xVec\,|\,\zVec)] }\,, \phantom{\small{}ix} \nonumber\\
=& \frac{1}{N}\sum_{n=1}^N \HH[\qPhiN(\zVec)]  \phantom{ii} \phantom{\sum_{n=1}^N}\hspace{-3ex}-\ \HH[\,p_{\PsiVec}(\zVec)] \phantom{ii}   -\ \EEE{\;\qBar_{\Phi}}{ \HH[\,p_{\ThetaVec}(\xVec\,|\,\zVec)] }\,-\,\log\big( h_0 \big)\,\nonumber\\ % \phantom{\small{}ix},\\
\Rightarrow\  %
\FF(\Phi,\PsiVec,\ThetaVec) =&  \frac{1}{N}\sum_{n=1}^N \HH[\qPhiN(\zVec)]  \phantom{ii} \phantom{\sum_{n=1}^N}\hspace{-3ex}-\ \HH[\,p_{\PsiVec}(\zVec)] \phantom{ii}   -\ \EEE{\;\qBar_{\Phi}}{ \HH[\,p_{\ThetaVec}(\xVec\,|\,\zVec)] }\,, \phantom{\small{}ix}
\label{EqnBothTheorems}
\end{align}
i.e., applying Theorem~\ref{th:Sum_of_Entr_Gen} recovers the result of Theorem~\ref{th:Sum_of_Entr} for any model fulfilling
the prerequisits of Theorem~\ref{th:Sum_of_Entr}. 
%
%
%
%we recover Theorem~\ref{th:Sum_of_Entr} by applying Theorem~\ref{th:Sum_of_Entr_Gen} to EF models with constant base measure.
%
%

\end{document}